\documentclass[twoside,11pt]{article}
\usepackage{xcolor}
\usepackage{amsmath}
\usepackage{wrapfig}
\usepackage{subcaption}
\usepackage{fancyhdr}
\usepackage[export]{adjustbox}
\usepackage{enumitem}
\usepackage{math_defs}
\usepackage{booktabs}
\usepackage{mathtools}
\usepackage[abbrvbib,preprint]{jmlr2e}

\usepackage{lastpage}

\begin{document}

\ShortHeadings{The Blessing of Dimensionality}{Woodring, Chan, Khan, Yun, Wong, and Chen}
\firstpageno{1}

\title{The Blessing of Dimensionality: How Near-Orthogonality in High-Dimensional Spaces Explains Temporal Portability}
\author{\name Abigail Woodring \email ajwoodri@ncsu.edu \\
        \addr Department of Electrical and Computer Engineering \\
        NC State University,
        Raleigh, NC 27606, USA \AND
        \name Adrian Chan \email adrian27513@gmail.com \\
        \addr Independent Researcher  \AND
        \name Rana Muhammad Shahroz Khan \email shahroz@cs.unc.edu \\
        \addr Department of Computer Science \\
        University of North Carolina at Chapel Hill,
        Chapel Hill, NC 27599, USA \AND
        \name Sukwon Yun \email swyun@cs.unc.edu \\
        \addr Department of Computer Science \\
        University of North Carolina at Chapel Hill,
        Chapel Hill, NC 27599, USA \AND
        \name Chau-Wai Wong \email chauwai.wong@ncsu.edu \\
        \addr Department of Electrical and Computer Engineering \\
        NC State University,
        Raleigh, NC 27606, USA \AND
        \name Tianlong Chen \email tianlong@cs.unc.edu \\
        \addr Department of Computer Science \\
        University of North Carolina at Chapel Hill,
        Chapel Hill, NC 27599, USA
        }
\editor{}

\maketitle

\begin{abstract}%
Fine-tuning has been widely used to adapt large language models (LLMs) for domain-specific tasks. Parameter efficient fine-tuning (PEFT) methods such as low-rank adaptation~(LoRA) are frequently used to reduce computational costs. PortLLM is a training-free and data-free scheme used to adapt LLMs after continual pretraining. Although the initial PortLLM results show that LoRA patches exhibit short-term temporal portability, the long-term performance of PortLLM across several updates of continual pretraining remains underexplored. Furthermore, the intriguing effectiveness of PortLLM is not well understood from a theoretical standpoint. We address these two open questions by (1) performing an extensive empirical study of the long-term temporal portability of PortLLM patches across $10$ continual pretraining steps using base models Mistral, Gemma, and Qwen; and (2) offering two theoretical analyses to explain our observation that the simple PortLLM method achieves competitive performance. We find empirically that the portability persists across longer time duration, indicating that repeated fine-tuning is not required when the base model is periodically updated. We find theoretically that near-orthogonality of high-dimensional vectors is a key justification for temporal portability. Our analyses also demonstrate a geometric perspective of the loss landscape in facilitating the theoretical comparison of different adaptation options.

\end{abstract}

\begin{keywords}
  Continual pretraining, downstream fine-tuning, loss landscape, low-rank adaptation, orthogonality
\end{keywords}

\section{Introduction}\label{intro}
The remarkable success and rapid growth of large language models (LLMs) is facilitated by a two-stage process of pretraining and fine-tuning~\citep{radford2018}. A model first learns general knowledge and language understanding via unsupervised pretraining on a huge corpus of general data. The model then excels at domain-specific tasks through supervised downstream fine-tuning on a smaller, more targeted data set. 

AI companies devote massive compute budgets to pretraining large models. These pretrained models are often fine-tuned by downstream developers with more constrained compute resources. Because of the scale of LLMs, direction fine-tuning of all parameters of the base model can be infeasible for resource-constrained downstream developers. To address this problem, parameter efficient fine-tuning (PEFT) methods, i.e., methods to reduce the number of trainable parameters during the fine-tuning stage, have been proposed~\citep{han2024}. Low-rank adaptation (LoRA), a widely used PEFT method, achieves this goal by restricting the rank of the update matrices~\citep{hu2022}. These rank-restricted update matrices are collectively referred to as a LoRA patch.

A challenge for the use of a LoRA patch occurs when the base model is continually pretrained. Continual pretraining is a process by which an initial base model is updated with new information that was not provided during the initial pretraining phase~\citep{gogoulou2024,jiang2024i}. While continual pretraining is generally performed by companies with substantial compute resources, downstream fine-tuning may be performed by smaller companies with limited resources. Re-training a new LoRA patch after each step of continual pretraining can be computationally expensive in such scenarios. Retraining may also be infeasible in proprietary domains where the downstream developer has only temporary access to private data, e.g., medical domains~\citep{Smajic2023}. To address this problem,~\cite{Khan2025} propose PortLLM, a method in which a LoRA patch fine-tuned on the initial pretrained base model is applied directly to the updated base model after continual pretraining without additional fine-tuning. They show that LoRA patches exhibit impressive temporal portability. However, their analysis focuses on short-term temporal portability, including only one time step of continual pretraining for most experiments and including at most $4$ time steps. 

To address the practical scenario of repeated updates to the base model, we study the long-term performance of PortLLM. We find empirically that the intriguing temporal portability of LoRA patches observed in~\cite{Khan2025} persists throughout our observation span of $10$ time steps and is consistent across three base models and three repetitions. To better understand these surprising results, we propose two methods for theoretical analysis leveraging (1) a 1-D slice of the loss landscape, and (2) iterative relations connected by consecutive continual pretraining steps. Although we apply these methods specifically to the PortLLM scenario, they remain applicable in broader use cases such as CAR-LoRA~\citep{shahroz2026} and other LoRA variants. Focusing on the PortLLM example, our analysis considers three possible deployment strategies: [I. Baseline] \textit{no patching}, the base model with no adapter; [II. Case Under Investigation] \textit{PortLLM patching}, the base model with the PortLLM patch that was fine-tuned on the initial base model;and [III. Oracle] \textit{stepwise fine-tuning}, the base model with the stepwise fine-tuning patch that was trained on the updated base model after continual pretraining. We organize our exploration of temporal portability into three research questions (RQs), illustrated in Figure~\ref{rq_diagram}:
\vspace{2mm}
\begin{wrapfigure}{t}{0.5\textwidth}%
    \centering
    \includegraphics[trim={20mm 48mm 80mm 49mm},clip,width=0.5\textwidth]{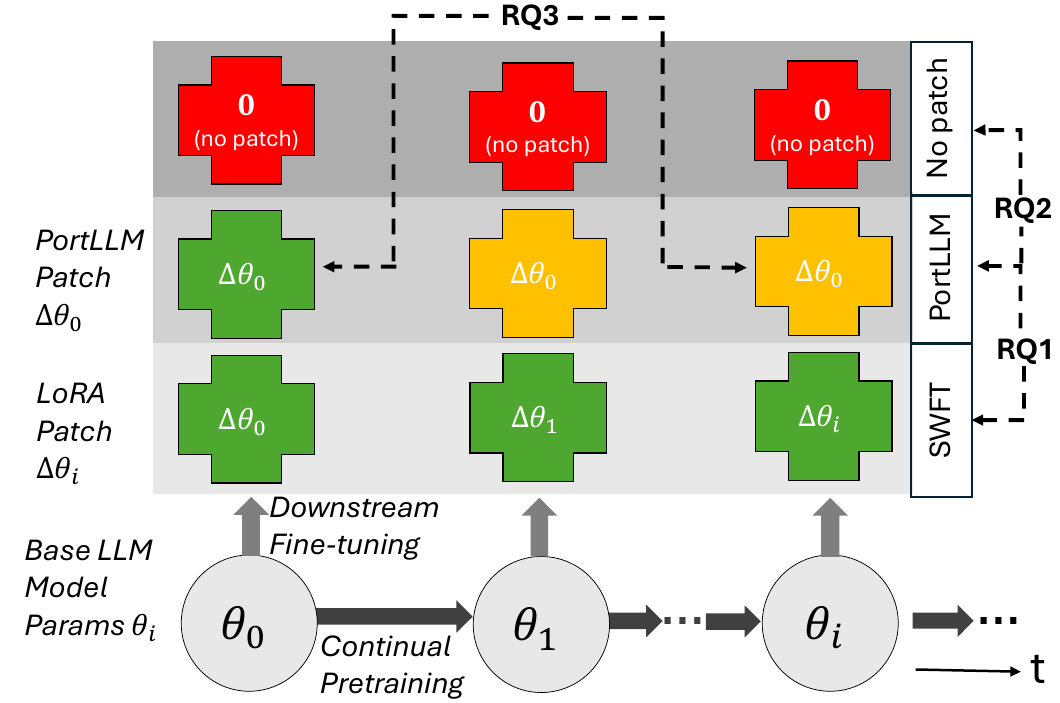}
    \caption{Comparison of three possible patching strategies represented by the three rows of patches: (bottom) stepwise fine-tuning (SWFT) fine-tunes a new patch $\expp{i}$ at each time step, (middle) PortLLM retains $\expp{0}$ trained at time step $0$, and (top) no patching uses the unadapted base model. 
}\label{rq_diagram}
\vspace{-10mm}
\end{wrapfigure}%

\noindent\textit{RQ1: Cost of PortLLM compared to stepwise fine-tuning.} 
To aid in relatively high-resource scenarios, we ask the research question: is there a substantial performance gain of using stepwise fine-tuning instead of the PortLLM method? In this scenario, the downstream developer has resources for stepwise fine-tuning. If stepwise fine-tuning will substantially outperform PortLLM, they will fine-tune a new patch at each time step. If there is no substantial difference, they will employ PortLLM. As shown in Figure~\ref{rq_diagram}, RQ1 compares stepwise fine-tuning (SWFT) in the bottom row of patches to PortLLM in the middle row.

\vspace{2mm}
\noindent\textit{RQ2: Benefit of PortLLM compared to no patching.}
To offer insight in low-resource scenarios, we ask the research question: is there a substantial performance gain of using the PortLLM method instead of no patching? In this scenario, the downstream developer cannot feasibly perform stepwise fine-tuning. If PortLLM substantially outperforms no patching performance, they will use PortLLM. Otherwise, no patching will be used. In Figure~\ref{rq_diagram}, RQ2 compares PortLLM performance to no patching.

\vspace{2mm}
\noindent\textit{RQ3: Variation in PortLLM performance compared to initial performance.}
To facilitate planning in general scenarios, we ask the research question: can the PortLLM method approximately maintain the $t=0$ performance? In this scenario, the downstream developer is satisfied with patching performance at $t=0$. If PortLLM can maintain that performance, they will employ PortLLM. If there is a substantial drop, they may choose to employ stepwise fine-tuning or to plan for a drop in performance. RQ3 compares PortLLM performance at different time steps, as illustrated in Figure~\ref{rq_diagram}.

Our contributions are as follows:
\begin{enumerate}[leftmargin=*]
\item We perform a large-scale study of long-term temporal portability across ten time steps, three base model architectures, and three repetitions. We find that LoRA patches exhibit remarkable temporal portability over repeated base model updates, different base model architectures, and multiple repetitions (Section~\ref{exp-results-main-text}). Our study offers the first substantial comparison of PortLLM and stepwise fine-tuning (RQ1).\footnote{\cite{Khan2025} preliminarily addressed this question with a limited scope of  one time step, one base model, and one repetition in Table 2 of their work.}
\item We propose a method to mathematically compare performance of two patching strategies based on a 1-D slice of the fine-tuning loss landscape (Section~\ref{1d-slice}). This method leverages geometric intuition to guide analysis and reveals that the difference in performance between two patching strategies can be compactly written as a single inner product. 
\item We propose an additional method to analyze temporal portability harnessing the iterative relationship of continual pretraining optimization steps (Section~\ref{pt-iter-analysis}). This provides a more fine-grained analysis result for RQ3, the change of PortLLM performance across time steps, by exploiting the relationship between fine-tuning and pretraining gradients.
\item We find that near-orthogonality of high-dimensional vectors is a root cause of temporal portability. This is a shared conclusion of both of our analysis methods.
\end{enumerate}

\section{Related Work}\label{related}
\noindent\textit{Continual pretraining of LLMs.}
Continual pretraining for LLMs is a process by which LLMs are updated with new data that was not learned during the initial pretraining phase. This process has been used to incorporate knowledge of different languages~\citep{gogoulou2024} and different domains~\citep{yildiz2024}. Other works have considered using continual pretraining to provide new temporal information as old information becomes outdated~\citep{jiang2024i}. Our study of long-term temporal portability across several time steps focuses on the latter case of updating temporal information, which is expected to involve repeated updates as old information continues to become outdated. The goal of such continual pretraining is to retain general information that remains relevant while updating old information and incorporating new knowledge~\citep{jang2022}.

\vspace{2mm}
\noindent\textit{Parameter efficient fine-tuning.} 
Full fine-tuning on all model parameters is prohibitively expensive in resource-constrained scenarios even for smaller-scale LLMs. Parameter efficient fine-tuning (PEFT)~\citep{han2024} refers to a class of methods used to adapt models for downstream tasks while reducing the number of trainable parameters. Some PEFT methods, e.g.,~\cite{houlsby2019}, insert light-weight modules with trainable parameters to the existing model architecture while freezing base model parameters. Other methods select a subset of base model parameters to fine-tune, e.g.,~\cite{guo2021} and~\cite{sung2021}. 
A third approach is prefix tuning~\citep{li2021}, which prepends a learnable prefix token sequence to each input sequence. 

Low-rank adaptation~\citep[LoRA, ][]{hu2022} is a widely-used PEFT method which is of particular interest to our work. The LoRA method adapts the base model with a LoRA patch comprised of a rank-restricted update for each weight matrix. Specifically, an update $\Delta W$ for a fully connected layer or projection matrix $W\in\real^{k \times d}$ is restricted to be low-rank, i.e., $\Delta W=BA$, where $B\in\real^{k\times r}$ and $A\in \real^{r\times d}$ contain learnable parameters, $d$ and $k$ are the input and output dimensions, and the rank $r\ll \min(k,d)$ is a hyperparameter.
This more restrictive parameterization of the update $\Delta W$ drastically reduces training complexity from quadratic to linear. The impact of LoRA is evidenced by its many variants. Weight-decomposed low-rank adaptation (DoRA) decomposes the LoRA patch into magnitude and direction components which are optimized separately~\citep{liu2024dora}. LoRA+~\citep{hayou2024} proposes training LoRA patches with different learning rates for $A$ and $B$ matrices. QLoRA~\citep{Dettmers2023} suggests training LoRA adapters on a quantized base model to reduce memory cost. 

\vspace{2mm}
\noindent\textit{Training-free methods and data-free methods.}
Training-free methods, i.e., methods to adapt a model for a new task or equip a model for a new capability without training, may be motivated by limited access to computational resources, monetary costs of obtaining computational resources, or logistical costs of initializing and monitoring training. Training-free methods have been proposed for various downstream applications. \cite{Zhang2025} propose a training-free method to integrate representations from non-text-based foundation models with text-based LLMs. \cite{ma2025} introduce a training-free method to steer the style of LLM-generated text without substantially changing the semantic meaning of the text output. \cite{Xiao2024} adapt LLMs trained on short text sequences to be capable of processing longer sequences without additional training. \cite{mijar2025} propose a training-free method to reuse a LoRA patch fine-tuned on an initial base model on a pruned version of the base model.

Data-free methods are often motivated by limited access to private data in domains such as medicine or finance. \cite{shukla2026} propose a data-free, training-free method to jointly use a set of LoRA patches. Focusing on the problem of private data,~\cite{Deng2025} develop a method to fine-tune a pretrained LLM based on access to a smaller model fine-tuned on the private data without access to the private data itself. \cite{das2025} seek to recover performance of a model that was degraded during quantization or datatype casting without access to the data set that was originally used to train the model.

PortLLM~\citep{Khan2025} is a training-free and data-free method for adapting a new continually pretrained model given an old LoRA patch downstream fine-tuned on the initial base model. Surprisingly, the authors show that simply applying the unchanged LoRA patch trained on the initial base model to the continually pretrained base model achieves competitive performance. Influenced by PortLLM,~\cite{shahroz2026} propose CAR-LoRA as a method for training an initial patch on the initial base model so that it will be robust to future continual pretraining and remains effective under different compression schemes. While the former discovers temporal robustness, the later emphasizes robustness to quantization.

While \cite{Khan2025} and~\cite{shahroz2026} focus on one time step of continual pretraining and include at most four time steps without statistical tests, our work explores long-term temporal portability by testing ten time steps of continual pretraining on Fineweb~\citep{penedo2024} data and eight time steps on Cosmopedia~\citep{benallal2024} data with three repetitions to offer statistical significance. Both works offer brief theoretical analysis based on distance in parameter space,\footnote{This focus on distance in parameter space uses the assumption that the loss function is Lipschitz---if the function is not Lipschitz, a small change in parameters may cause a large change in loss. This assumption is implicit in PortLLM and explicit in CAR-LoRA.} and this work is introducing a full theoretical treatment with more detailed analysis based on properties of the loss landscape in a region around the pretrained models. Even though our work focuses on the PortLLM scenario, our analysis techniques could be applied to more general cases, e.g., CAR-LoRA or other LoRA variants.

\section{Motivating Experimental Study of Long-Term Temporal Portability}\label{experimental}
We empirically study the long-term performance of PortLLM, motivating analysis in Section~\ref{theoretical}. We describe our experimental setup in Section~\ref{exp-setup} and present results in Section~\ref{exp-results-main-text}.

\subsection{Experimental Setup}\label{exp-setup}
We describe base models, pretraining data sets, and fine-tuning data sets and benchmarks. More training details and hyperparameters are outlined in Appendix~\ref{app:hyperparams}.

\vspace{2mm}
\noindent\textit{Base models.}
To empirically assess long-term (4+) temporal portability of PortLLM patches across different model architectures and scales, we perform $10$ time steps of continual pretraining on Mistral-7B-v0.1~\citep{jiang2023} and Gemma3-12B~\citep{gemma_2025}. Due to compute resource constraints, we follow the prior works~\cite{Khan2025,shahroz2026} and approximate full pretraining by training relatively high rank LoRA patches (using $r=64$ for Mistral-7B-v0.1 and $r=128$ for Gemma3-12B) and merging them to the base model. We confirm that PortLLM remains comparable to stepwise fine-tuning and outperforms no patching under full pretraining by running full pretraining on all parameters of a smaller model, namely, Qwen2-0.5B~\citep{yang2024}.\footnote{We find that, when stepwise fine-tuning performance degrades, PortLLM performance mirrors that degradation. This highlights the importance of careful pretraining so that fine-tuning remains effective.} 

\vspace{2mm}

\noindent\textit{Continual pretraining data sets.}
We aim to perform continual pretraining that continually updates outdated information given new temporal knowledge. To accomplish this, we continually pretrain on 10 consecutive temporal chunks for Fineweb~\citep{penedo2024}, a corpus of filtered English web data. 
Each chunk for Mistral-7B and Gemma3-12B consists of $50$M tokens. For Qwen2-0.5B, we use $200$M tokens per chunk. To investigate temporal portability under temporally heterogeneous pretraining data, we continually pretrain Mistral\footnote{We validate the generality of our results by considering three base models and two pretraining data sets. Due to computational constraints, we cannot test all permutations. Instead, we test all base models with Fineweb pretraining data and both pretraining data sets on one base model, namely, Mistral.} on eight sections of Cosmopedia~\citep{benallal2024}, where each section comprises a time step. Cosmopedia consists of eight splits of synthetic data generated from different seed data including web data (\texttt{web\_samples\_v1} and \texttt{web\_samples\_v2} splits), course outlines (\texttt{stanford}, \texttt{openstax}, and \texttt{khanacademy} splits), general questions (\texttt{stories} split), WikiHow article titles (\texttt{wikihow} split), and mathematical texts (\texttt{auto\_math\_text} split).\footnote{Similar to Fineweb, we restrict each time step to train on at most $50$M tokens.}

\vspace{2mm}
\noindent\textit{Downstream datasets and benchmarks.}
We test temporal portability of PortLLM patches with four downstream benchmarks targeting commonsense reasoning [WinoGrande~\citep{sakaguchi2021}], reading comprehension [BoolQ~\citep{clark2019}], and science knowledge and reasoning [ARC-Easy and ARC-Challenge~\citep{clark2018}]. We also use additional benchmarks when Mistral is used as the base model. 
These benchmarks include GSM8k~\citep{cobbe2021} for mathematical reasoning and two GLUE benchmarks~\citep{wang2019}, namely, MNLI~\citep{williams2018} on recognizing textual entailment and SST2~\citep{socher2013} on sentiment analysis. For most benchmarks, we train using the publicly available training split of the relevant data set. Patches for the GSM8k benchmark are trained using MetaMath~\citep{yu2023}. To test PortLLM and stepwise fine-tuning performance, we train PortLLM patches on the initial base model at time step $0$ and stepwise fine-tuning patches on continually pretrained base models at time steps $t>0$. To reduce computation expense, we periodically retrain stepwise fine-tuning patches at even time steps and consequently show evaluation results at even time steps. Each patch is trained with a LoRA rank of $r=8$.
\subsection{Experimental Results}\label{exp-results-main-text}
\begin{figure}%
    \subfloat{\includegraphics[width=0.165\textwidth,clip,trim={5mm 35mm 55mm 9mm}]{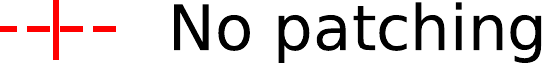}
    \includegraphics[width=0.22\textwidth,clip,trim={5mm 19mm 25mm 25mm}]{figures/experiments/legend.pdf}
    \includegraphics[width=0.25\textwidth,clip,trim={4mm 4mm 9mm 40mm}]{figures/experiments/legend.pdf}\hfill}
    \centering
    \addtocounter{subfigure}{-1}\\
    \subfloat[]{\includegraphics[width=0.25\textwidth]{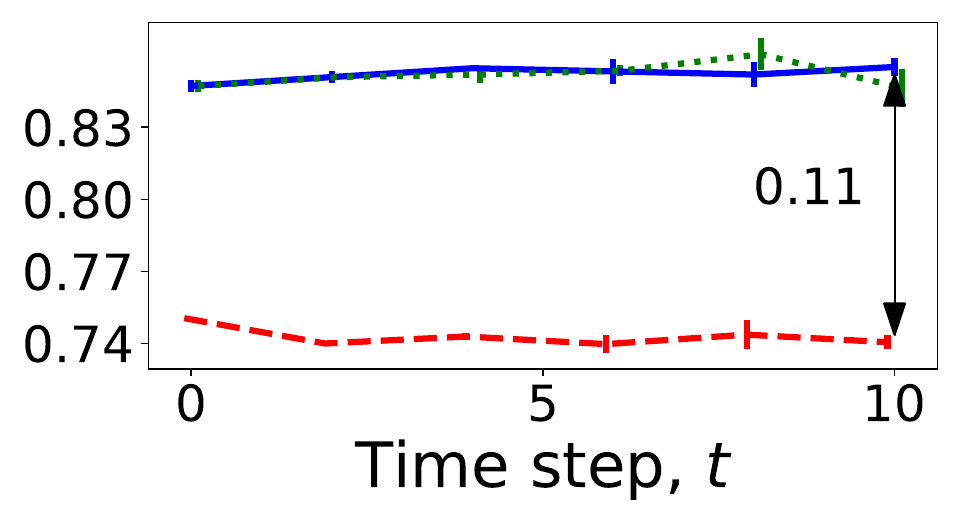}}\subfloat[]{\includegraphics[width=0.25\textwidth]{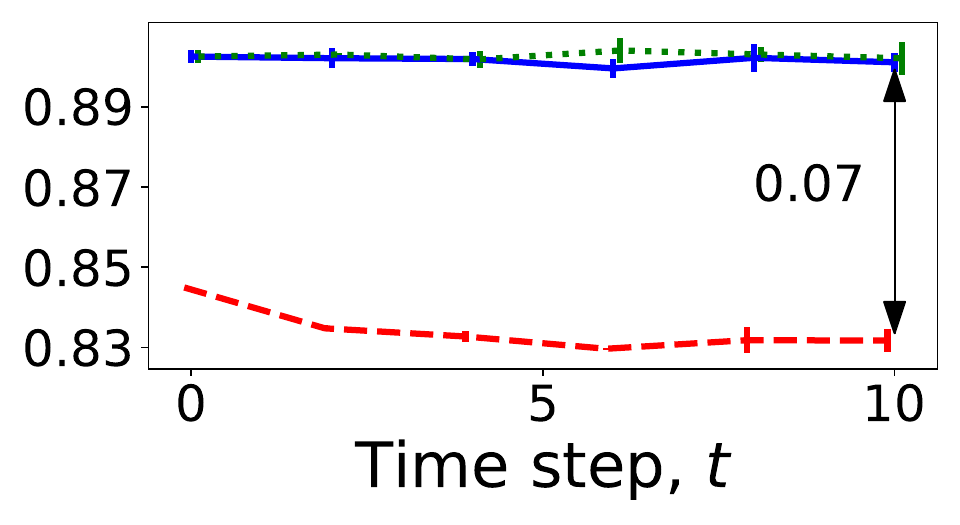}}\subfloat[]{\includegraphics[width=0.25\textwidth]{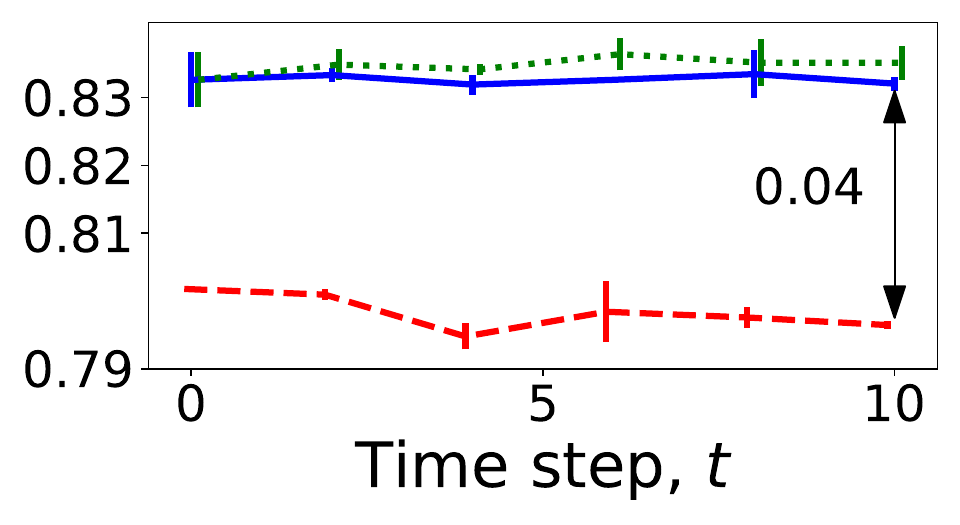}}\subfloat[]{\includegraphics[width=0.25\textwidth]{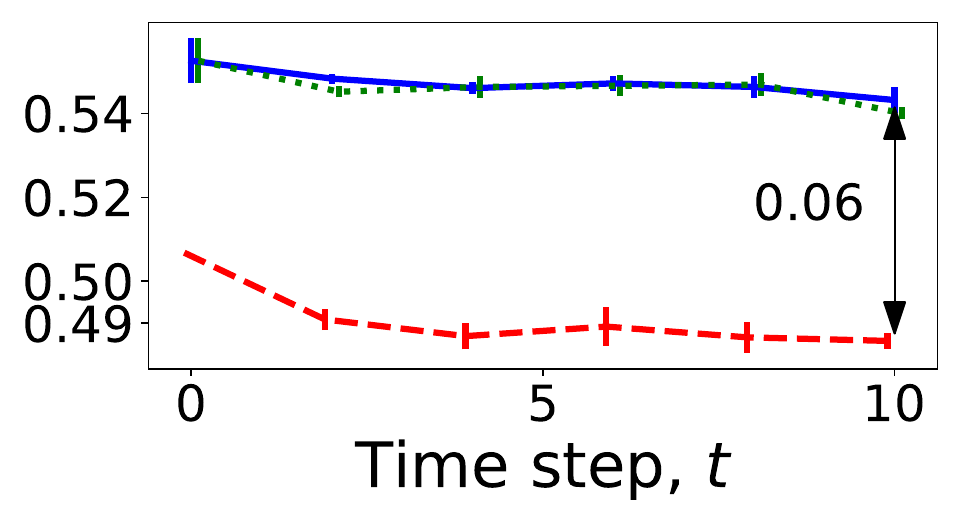}}
    \caption{Temporal portability results for continually pretraining Mistral-7B-v0.1 on Fineweb data and downstream fine-tuning on (a) WinoGrande, (b) BoolQ, (c) ARC-Easy, and (d) ARC-Challenge. Average performance across $3$ repetitions for PortLLM patching, stepwise fine-tuning, and no patching performance are shown, with the error bars indicating the 95\% confidence interval. It can be observed that (RQ1) the cost of using PortLLM compared to stepwise fine-tuning is almost negligible given the overlapping solid blue and dotted green curves, (RQ2) the benefit of using PortLLM instead of no patching is substantial as shown by the black arrows, and (RQ3) the delta in PortLLM performance across time steps is small given the almost flat solid blue curves.}
    \label{fig:fw_mistral_exp}
    \vspace{-5mm}
\end{figure}%
\begin{wraptable}{R}{0.3\textwidth}%
    \centering
    \vspace{-\intextsep}
    \begin{tabular}{p{1.9cm} p{2cm} 
    }%
    \toprule
        \textbf{Repetition} & \textbf{Slope} $\mathbf{(\times10^{-4})}$
        \\
        \hline
            \multicolumn{2}{l}{WinoGrande} \\
            \hline
            \#1& $4.85$
            \\
            \#2 & $8.12$**
            \\
            \#3 & $4.74$
            \\
            \hline
            \multicolumn{2}{l}{BoolQ} \\
            \hline
            \#1& $3.32$
            \\
            \#2 & $-2.62$ 
            \\
            \#3 & $-4.63$* 
            \\
            \hline
            \multicolumn{2}{l}{ARC-Easy} \\
            \hline
            \#1& $0.90$ 
            \\
            \#2 & $-1.14$
            \\
            \#3 & $-0.48$
            \\
            \hline
            \multicolumn{2}{l}{ARC-Challenge} \\
            \hline
            \#1& $-4.39$ 
            \\
            \#2 & $-5.85$
            \\
            \#3 & $-11.95$*
            \\
        \toprule 
        \multicolumn{2}{l}{{\tiny*$p\le0.05$, **$p\leq0.01$}}
    \end{tabular}%
    \vspace{-4mm}
    \caption{Statistical results for PortLLM performance. }
    \label{tab:mistral_fw_stats_results}
    \vspace{-15mm}
\end{wraptable}%
Temporal portability results for Mistral-7B pretrained on Fineweb are shown in Figure~\ref{fig:fw_mistral_exp}. Additional experimental results are shown in Appendix~\ref{app:exp}. First, PortLLM performance (blue solid curves) and stepwise fine-tuning performance (green dotted curves) are comparable, with a slight difference only for ARC-Easy. Second, PortLLM clearly outperforms the base model (red dashed curves), with $11$, $7$, $4$, or $6$ percentage points gains for the four benchmarks at $t=10$. Third, performance is approximately constant across time steps, with mild degradation only for ARC-Challenge. To justify the approximately constant claim, we perform a statistical test considering an intercept model as the null hypothesis and a linear model as the alternative hypothesis and report results in Table~\ref{tab:mistral_fw_stats_results}. We find statistically significant evidence ($p\leq 0.05$) for a linear trend in only $3$ out of $12$ cases, indicating that the change across time steps is not substantial. Furthermore, the estimated slope is at most on the order of $10^{-3}$. 

These results directly address our three research questions. Specifically, there is not a substantial cost of using PortLLM instead of stepwise fine-tuning, answering RQ1; there is a substantial benefit of using PortLLM instead of no patching, answering RQ2; and the variation in PortLLM performance across time steps is small, answering RQ3. The $4$-step temporal portability preliminarily observed in~\cite{Khan2025} persists across $10$ time steps. Additionally, we find that PortLLM performance is comparable to stepwise fine-tuning---a comparison that was only briefly considered in~\cite{Khan2025}. We offer theoretical analysis for these results in Section~\ref{theoretical}.

\section{Proposed Theoretical Analyses of Temporal Portability}\label{theoretical}
We propose two approaches for theoretical analysis to explain the empirical results observed in Section~\ref{experimental}. Our first method of analysis, outlined in Section~\ref{1d-slice}, uses the geometry of the loss landscape to guide comparison of three patching strategies. Specifically, we leverage a 1-D slice of the loss landscape between pairs of patches to characterize the difference in performance. Our second method of analysis, described in Section~\ref{pt-iter-analysis}, exploits the iterative relationship between continual pretraining optimization steps to characterize the difference in PortLLM performance across time steps. 

\subsection{Notation and Definitions}\label{definitions}
We introduce notation and definitions to facilitate mathematical analysis of temporal portability. Let $\base{}\in\real^N$ be a column vector of all parameters for a base model of size $N$, where $N$ is on the order of billions for LLMs. We define \textit{expanded vectorized LoRA patch parameters} trained on base model $\base{t}$ as $\expp{t}\in\real^N$. Similarly, we denote \textit{unexpanded vectorized LoRA patch parameters} trained on $\base{t}$ as $\unexpp{t}\in\real^n$, where $n\ll N$ is the number of LoRA trainable parameters on the order of millions in our work.\footnote{See Appendix~\ref{app:definitions} for a more thorough explanation of the connection between $\expp{}$ and $\unexpp{}$. In short, $\expp{}$ is obtained by vectorizing the LoRA matrices $A$ and $B$ separately, while $\unexpp{}$ is obtained by vectorizing the $BA$ matrix product.} 

Let $\lpt:\real^N\rightarrow \real$ be a pretraining testing loss for a model $\base{}$, defined as the expectation averaged over pretraining testing data $\mathcal{D}_\text{pt}$, namely,  $\lpt(\base{})=\E_{x\sim\mathcal{D}_\text{pt}}[\ce(x,\base{})]$, where $x\in\mathcal{X}$ is a random dummy testing example and $\ce:\mathcal{X}\times \real^N\rightarrow \real$ is per-example cross-entropy loss. In contrast, let $\lft: \real^N\rightarrow \real$ be a downstream testing loss defined as the expectation averaged over fine-tuning testing data $\mathcal{D}_\text{ft}$ associated with a domain-specific task, namely,  $\lft(\base{})=\E_{x\sim\mathcal{D}_\text{ft}}[\ce(x,\base{})]$. We further define $\uloss{\base{}}(\unexpp{})=\eloss{\base{}+\expp{}}$ to be the downstream testing loss function applied to unexpanded patch parameters on base model $\base{}$, where $\expp{}\in\real^N$ and $\unexpp{}\in\real^n$ are corresponding expanded and unexpanded parameter vectors for the same patch given the base model $\base{}$. This definition allows analysis in lower-dimensional space.

To analyze RQ1, we quantify the cost/penalty of employing PortLLM with reference to stepwise fine-tuning as the difference in fine-tuning loss between the PortLLM-adapted model $\base{t}+\expp{0}$ and the stepwise fine-tuning-adapted model $\base{t}+\expp{t}$:
\begin{equation}\label{cost-def}
        \cpllm(t)=\eloss{\base{t}+\expp{0}}-\eloss{\base{t}+\expp{t}} = \uloss{\base{t}}(\unexpp{0})-\uloss{\base{t}}(\unexpp{t}).
\end{equation}
To address RQ2, we define the benefit of using PortLLM instead of no patching as the difference in loss on the base model $\base{t}$ and the PortLLM-adapted model $\base{t}+\expp{0}$:
\begin{equation}\label{benefit-def}
        \bpllm(t)=\eloss{\base{t}}-\eloss{\base{t}+\expp{0}} = \uloss{\base{t}}(\mathbf{0})-\uloss{\base{t}}(\unexpp{0}).
\end{equation}
Analysis for RQ3 is aided by the following definition, which describes the variation in PortLLM performance at time step $0$, $\base{0}+\expp{0}$, and time step $t>0$, $\base{t}+\expp{0}$:
\begin{equation}\label{variation-def}
        \vpllm(t)=\eloss{\base{0}+\expp{0}}-\eloss{\base{t}+\expp{0}} = \uloss{\base{0}}(\unexpp{0})-\uloss{\base{t}}(\unexpp{0}).
\end{equation}

\subsection{Analysis via 1-D Slice of Loss Landscape}\label{1d-slice}
We analyze the difference in performance across patching strategies via a 1-D slice in the loss landscape. We outline this method by providing three examples corresponding to our three research questions. Because derivations for RQ1, RQ2, and RQ3 are similar, we present a full derivation and result for RQ1 and summarize only the results for RQ2 and RQ3 in the main text. Full derivations for RQ2 and RQ3 are provided in Appendix~\ref{app:1d-slice-analysis}. 

Before presenting analysis and results for our research questions, we present an idea that helps interpret our results. For each research question, we find that the relevant metric can be compactly written as a single inner product. Although the three research questions appear unrelated, we find that their corresponding inner products have a shared form, which we call a \textit{displaced-gradient--displacement product}. Consider a function $f:\real^P\rightarrow\real$ and a reference input $x_0\in\real^P$. Let $\Delta x\in \real^P$ be a displacement direction. We use $\gamma \Delta x$ with scale $\gamma>0$ to control the norm of the displacement and assume without loss of generality that $\|\Delta x\|_2=1$. Define the displaced-gradient--displacement product as
\begin{equation}
    F(\gamma \Delta x)=\langle \nabla f(x_0+\gamma\Delta x),\gamma \Delta x\rangle,
\end{equation}
or the inner product of the displacement $\gamma \Delta x$ and the gradient evaluated at the reference input $x_0$ displaced by $\gamma \Delta x$. In many cases for our analysis, the reference input $x_0$ is a proxy for a local minimum. Based on intuition for 3-D space, one may expect that the displaced-gradient--displacement product $F(\gamma \Delta x)$ is large because the gradient $\nabla f(x_0+\gamma \Delta x)$ points in the direction $\Delta x$ away from the optimal point $x_0$ and hence is aligned with the displacement. We find that more fine-grained analysis is needed in high-dimensional spaces. In Appendix~\ref{app:dis_grad_inner}, we derive the following Lemma.
\begin{lemma}\label{lim-dgdp}
    The displaced-gradient--displacement product $F(\gamma \Delta x)=\langle \nabla f(x_0+\gamma\Delta x),\gamma \Delta x\rangle$ for a function $f:\real^P\rightarrow\real$, where $x_0$ is a reference point, $\Delta x$ is a unit norm displacement direction, and $\gamma$ is a displacement magnitude, is upper-bounded as
    \begin{equation}
    F(\gamma \Delta x)\leq \gamma\langle\nabla f(x_0) ,\Delta x\rangle+\gamma^2\sigma_{\text{max}}(x_0+\beta'\gamma\Delta x)
    \end{equation}
    where $\sigma_\text{max}(x)=\|\nabla^2f(x)\|_2$ is the spectral norm of the Hessian and $\beta'\in[0,1]$ is fixed but unknown.
\end{lemma}
From this upper bound, we find two metrics that aid our analysis of our three research questions. First, the \textit{alignment} $\langle\nabla f(x_0) ,\Delta x\rangle$ of the gradient at the reference input $x_0$ with the displacement $\Delta x$ characterizes the first term of the upper bound. If the displacement is approximately orthogonal to the gradient at the reference input, the alignment is small. Second, the \textit{sharpness} $\sigma_{\text{max}}(x_0+\beta'\gamma\Delta x)$ of the loss landscape characterizes the second term. If the loss landscape is flat around the reference input, the sharpness is small. We discuss these two metrics for each of our three research questions in Remarks~\ref{rmk:gc_inner},~\ref{rmk:gg_inner}, and~\ref{rmk:gd_inner}. 

\vspace{2mm}
\noindent\textit{Analysis for $\cpllm$ (RQ1).}
We demonstrate the method of using a 1-D slice of the loss landscape to compare batching strategies by applying it to compare the PortLLM patch $\unexpp{0}$ to the stepwise fine-tuning patch $\unexpp{t}$. Geometric interpretations for this analysis are depicted in Figure~\ref{cost_illustration}. The key insight is to consider the two patches in $n$-dimensional space, as shown in the left side of Figure~\ref{cost_illustration}, and to focus on the 1-dimensional line segment connecting the two patches, as shown in the right side of Figure~\ref{cost_illustration}. We define the scalar function $g_c:\real \rightarrow \real$ as the downstream fine-tuning loss on a 1-D slice in $n$-dimensional space between the PortLLM patch $\unexpp{0}$ (case under investigation) and the stepwise fine-tuned patch $\unexpp{t}$ (the oracle), plotted as a solid blue arrows in Figure~\ref{cost_illustration}. Specifically, for a fixed time step $t>0$, 
\begin{equation}\label{gc-def}
    \gc{\alpha}=\uloss{\base{t}}\big(\alpha \unexpp{0}+(1-\alpha)\unexpp{t}\big)
\end{equation}%
where $\alpha\in\real$ but the interval $\alpha\in[0,1]$ between the two patches is of primary interest. With this definition, $\gc{0}=\uloss{\base{t}}(\unexpp{t})$ degenerates to the stepwise fine-tuning performance, $\gc{1}=\uloss{\base{t}}(\unexpp{0})$ degenerates to the PortLLM performance, hence~\eqref{cost-def} can be rewritten as $\cpllm(t)=\gc{1}-\gc{0}$. Taylor's theorem gives that, for some $\alpha'\in [0,1]$,
\begin{equation}\label{gc-taylor}
    \gc{1}=\gc{0}+(1-0)\gcp{\alpha'}
\end{equation}
where the derivative is taken with respect to $\alpha$.\footnote{Note that, if the function is quadratic, then $\alpha'=0.5$. If $\gc{\alpha}=a\alpha^2+b\alpha+c$, then $\gc{1}=a+b+c$ and $\gc{0}=c$ so that $\gc{1}-\gc{0}=a+b$. Setting $
\gcp{\alpha'}=2a\alpha'+b=\gc{1}-\gc{0}=a+b$, we find that $\alpha'=0.5$. 
} This gives that $\cpllm(t)=\gc{1}-\gc{0}=\gcp{\alpha'}$. This equality is demonstrated by the filled circle on the red dashed curve $\gcp{\alpha}$ corresponding to $\alpha'$ in Figure~\ref{cost_illustration}. Further insight can be achieved by considering this result in $n$-dimensional space. By the chain rule, this derivative can be written as
\begin{subequations}\label{gc-derivative}%
    \begin{align}%
        \gcp{\alpha}&=\frac{d}{d\alpha}\uloss{\base{t}}\big(\alpha \unexpp{0}+(1-\alpha)\unexpp{t}\big) 
        =\langle \nabla_{\unexpp{ }}\uloss{\base{t}}(\uls{\alpha}),\ftdelta{t}\rangle%
    \end{align}%
\end{subequations}%
\noindent where $\uls{\alpha}=\alpha \unexpp{0}+(1-\alpha)\unexpp{t}$ is a point on the 1-D slice parameterized by $\alpha$,  $\ftdelta{t}=\unexpp{0}-\unexpp{t}$ is the difference between the two fine-tuned patches, and the superscript (ft) refers to the fine-tuning update. This leads to the following Lemma~\ref{lim-gc}:
\begin{figure}{}{}%
    \centering
    \includegraphics[trim={0mm 48mm 0mm 78mm},clip,width=1\textwidth]{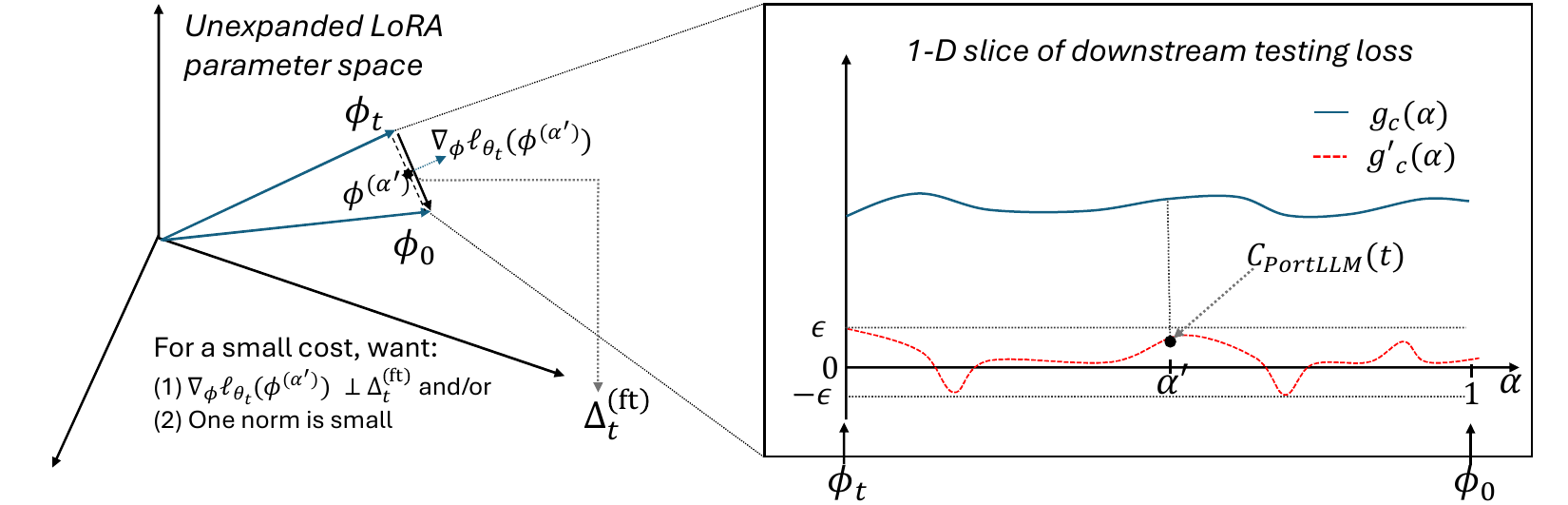}
    \vspace{-8mm}
    \caption{Geometric interpretation for analyzing the cost of employing PortLLM instead of stepwise fine-tuning, $\cpllm{}(t)$. The PortLLM patch $\unexpp{0}$ and the stepwise fine-tuned patch $\unexpp{t}$ are shown as blue arrows in unexpanded LoRA parameter space on the left. The downstream testing loss on a 1-D slice between the two patches is shown on the right. The loss, shown as a blue solid line, is approximately constant so that the derivative, shown as a red dashed line, has magnitude bounded by $\epsilon$. $\alpha'$ is the point on the 1-D slice at which the derivative exactly predicts the cost, $\cpllm(t)$. $\alpha'$ on the 1-D slice corresponds to $\uls{\alpha'}$ in parameter space. 
    }\label{cost_illustration}
    \vspace{-8mm}
\end{figure}%
\begin{lemma}\label{lim-gc}
    There exists an $\alpha'\in[0,1]$ such that $\cpllm(t)=\langle \nabla_{\unexpp{ }}\uloss{\base{t}}(\uls{\alpha'}),\ftdelta{t}\rangle$ where $\uls{\alpha}=\alpha \unexpp{0}+(1-\alpha)\unexpp{t}$ is a point on the 1-D slice between the two patches and $\ftdelta{t}=\unexpp{0}-\unexpp{t}$ is the difference between the two patches.
\end{lemma}
\begin{remark}\label{rmk:gc_inner}
    The inner product in~\ref{lim-gc} can be written as a displaced-gradient--displacement product as $\langle \nabla_{\unexpp{ }}\uloss{\base{t}}(\uls{\alpha}),\ftdelta{t}\rangle=\langle \nabla_{\unexpp{ }}\uloss{\base{t}}(\unexpp{t}+\alpha\ftdelta{t}),\ftdelta{t}\rangle$. 
    Using the two metrics derived from Lemma~\ref{lim-dgdp}, we find that the alignment and sharpness are both small, indicating that the displaced-gradient--displacement product is small. First, the gradient $\nabla_{\unexpp{ }}\uloss{\base{t}}(\unexpp{t})$ of the fine-tuning loss at the reference input $\unexpp{t}$ is largely unrelated to the displacement direction $\ftdelta{t}/\|\ftdelta{t}\|_2$ where $\ftdelta{t}=\unexpp{0}-\unexpp{t}$, indicating that the two are approximately orthogonal and that their alignment is small. Second, the results in Appendix~\ref{app:1d-slice-exp} indicate that the loss landscape is flat in a region around $\unexpp{t}$ with a radius at least $\|\ftdelta{}\|_2$ so that the sharpness $\sigma_\text{max}(\unexpp{t}+\beta'\alpha\ftdelta{t})$ is small. 
\end{remark}
From Lemma~\ref{lim-gc}, we arrive a Theorem~\ref{thm-gc}.
\begin{theorem}\label{thm-gc}
    Let $\epsilon=\max_{\alpha\in[0,1]}|\langle \nabla_{\unexpp{ }}\uloss{\base{t}}(\uls{\alpha}),\ftdelta{t}\rangle|$ be the maximum derivative magnitude on the 1-D slice between the PortLLM patch and the stepwise fine-tuned patch where $\uls{\alpha}=\alpha \unexpp{0}+(1-\alpha)\unexpp{t}$ is a point on the 1-D slice and $\ftdelta{t}=\unexpp{0}-\unexpp{t}$ is the difference between the two patches. Then the cost of using PortLLM instead of stepwise fine-tuning is bounded as $|\cpllm(t)|\leq \epsilon$. 
\end{theorem}
Proof: By Lemma~\ref{lim-gc}, there exists an $\alpha'\in[0,1]$ such that $\langle \nabla_{\unexpp{ }}\uloss{\base{t}}(\uls{\alpha'}),\ftdelta{t}\rangle$ exactly predicts the cost. Taking the maximum over $\alpha\in[0,1]$ upper bounds the cost.

\begin{figure}
    \subfloat{\includegraphics[width=0.12\textwidth,clip,trim={4mm 3mm 27mm 20mm}]{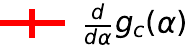}
    \includegraphics[width=0.2\textwidth,clip,trim={4mm 12mm 4mm 10mm}]{figures/data_dump/cost_legend.pdf}
    \includegraphics[width=0.11\textwidth,clip,trim={4mm 20mm 28mm 4mm}]{figures/data_dump/cost_legend.pdf}\hfill}
    \centering
    \addtocounter{subfigure}{-1}\\
    
    \subfloat[]{\includegraphics[width=0.25\textwidth]{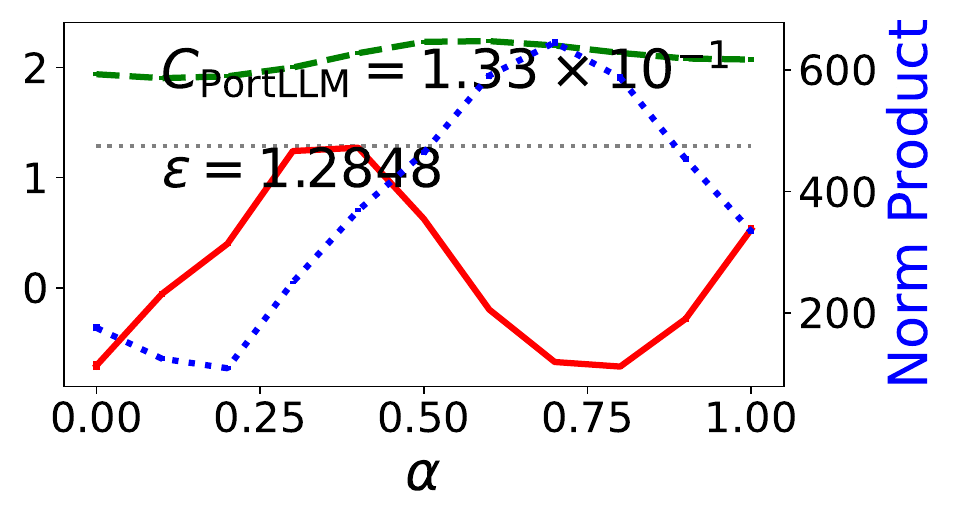}}
    \subfloat[]{\includegraphics[width=0.25\textwidth]{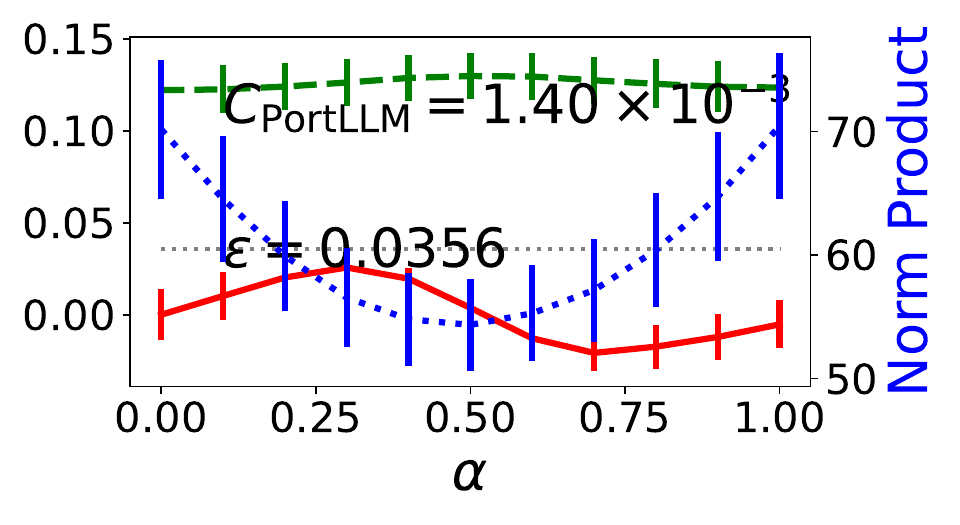}}
    \subfloat[]{\includegraphics[width=0.25\textwidth]{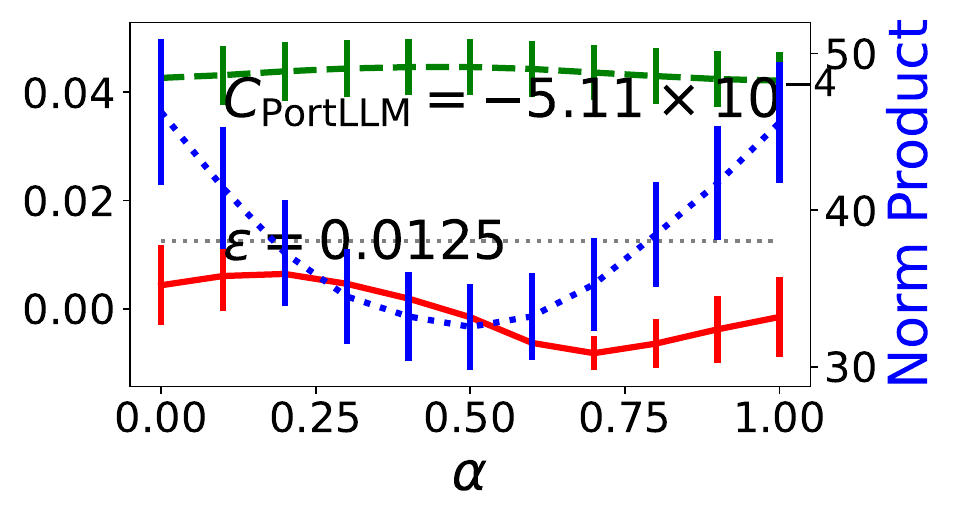}}
    \subfloat[]{\includegraphics[width=0.25\textwidth]{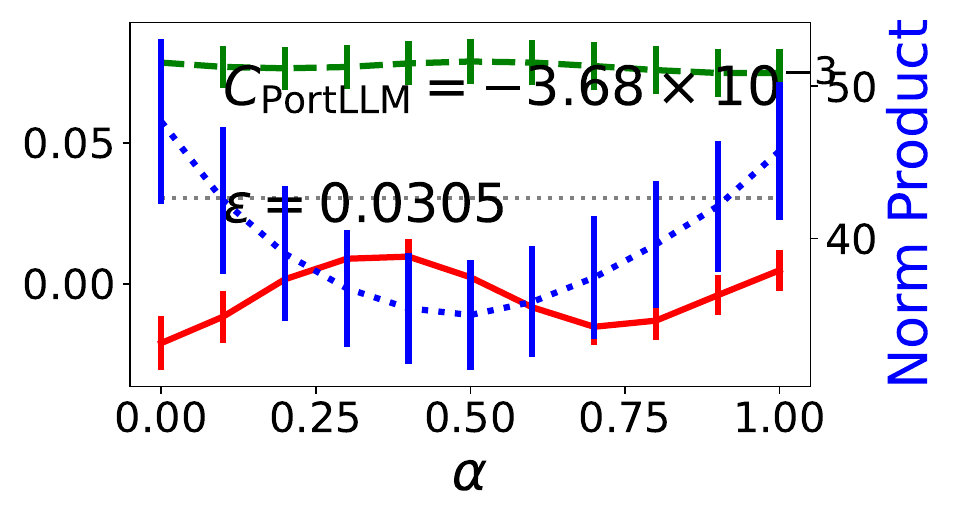}} \\

    \subfloat{\includegraphics[width=0.12\textwidth,clip,trim={4mm 3mm 36mm 20mm}]{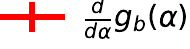}
    \includegraphics[width=0.22\textwidth,clip,trim={4mm 12mm 4mm 10mm}]{figures/data_dump/benefit_legend.pdf}
    \includegraphics[width=0.15\textwidth,clip,trim={4mm 20mm 28mm 4mm}]{figures/data_dump/benefit_legend.pdf}\hfill}
    \addtocounter{subfigure}{-1}\\

    \subfloat[]{\includegraphics[width=0.25\textwidth]{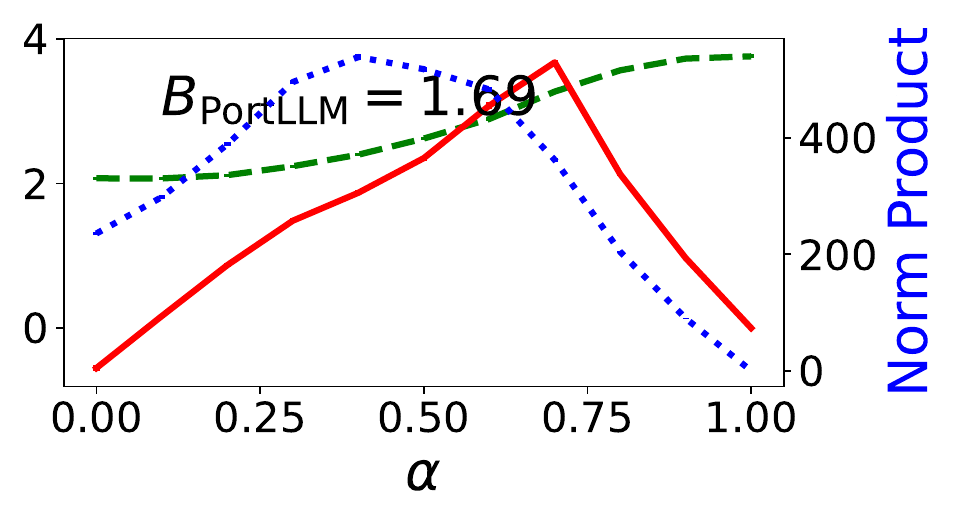}}
    \subfloat[]{\includegraphics[width=0.25\textwidth]{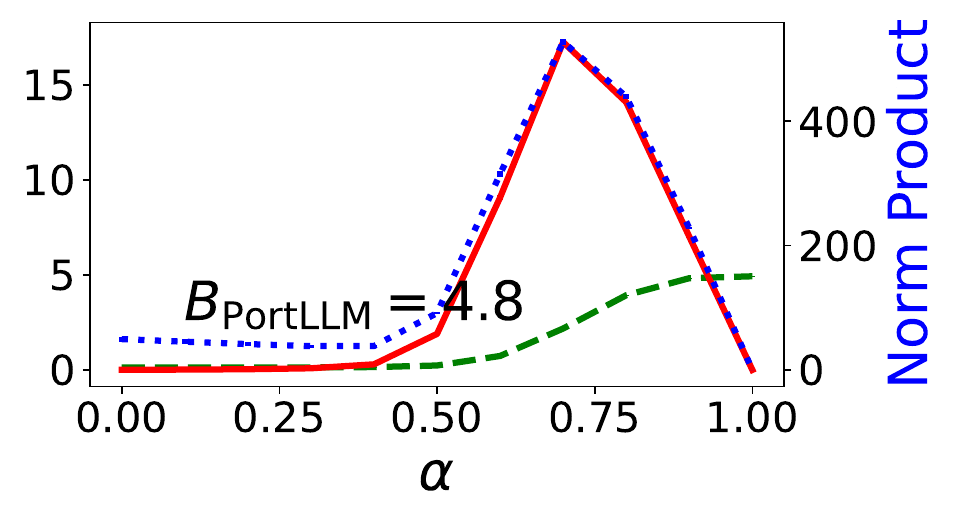}}
    \subfloat[]{\includegraphics[width=0.25\textwidth]{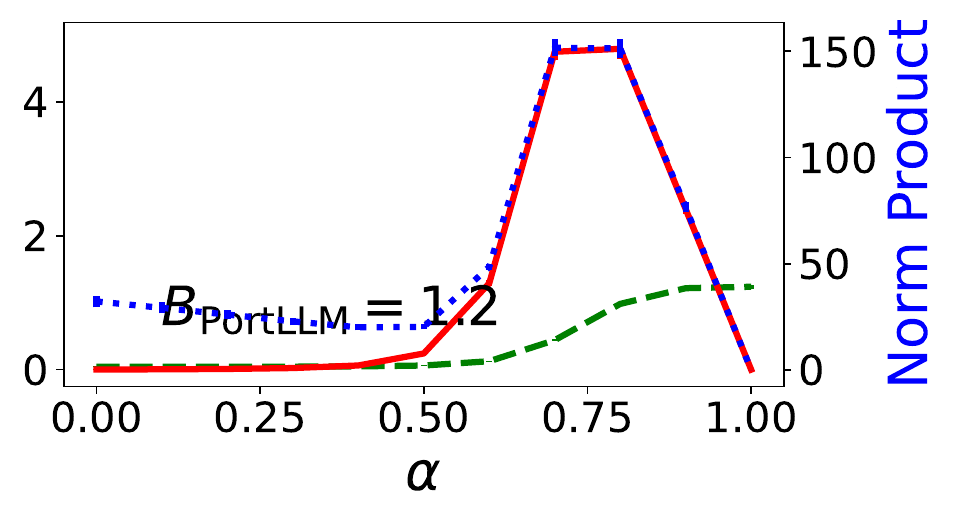}}
    \subfloat[]{\includegraphics[width=0.25\textwidth]{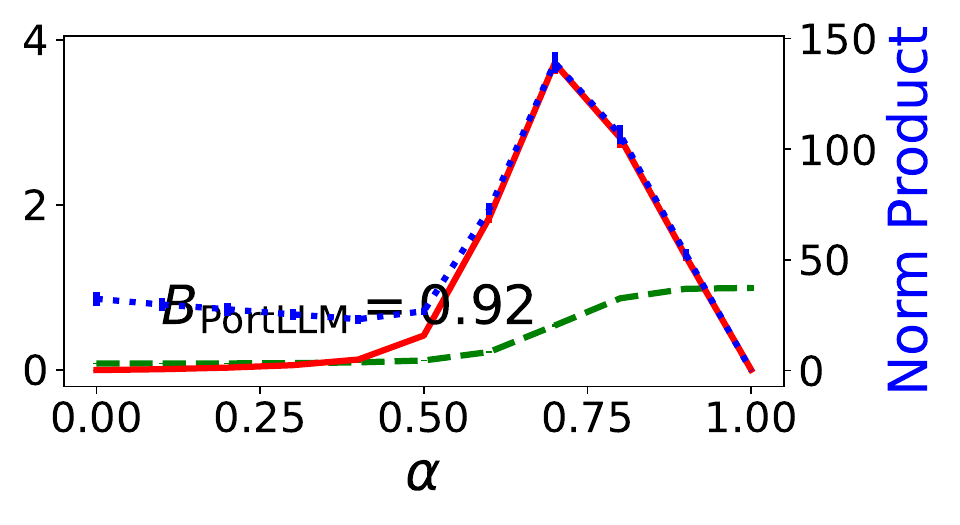}}
   
    \caption[]{Empirical results for cost (top row) and benefit (bottom row) metrics on 1-D slice for $t=2$. We shown (a) and (e) WinoGrande (b) and (f) BoolQ (c) and (g) ARC-Easy and (d) and (h) ARC-Challenge. The derivative $\gcp{\alpha}$ is small in most cases, indicating that the cost of using PortLLM instead of stepwise fine-tuning is small. The derivative $\gcp{\alpha}=\langle \nabla_{\unexpp{ }}\uloss{\base{t}}(\uls{\alpha}),\ftdelta{t}\rangle$ is much smaller than the norm product $\|\nabla_{\unexpp{ }}\uloss{\base{t}}(\uls{\alpha})\| \|\ftdelta{t}\|$, indicating that the two vectors are nearly orthogonal. The derivative $\gbp{\alpha}$ is large for $0.5<\alpha<1$, explaining why PortLLM substantially outperforms no patching.
    }
    \label{fig:mt_sample}
    \vspace{-8mm}
\end{figure}

The following limitation should be remembered when invoking Theorem~\ref{thm-gc}.
\begin{remark}\label{rmk:lim-gc}
    While the expression in Lemma~\ref{lim-gc} is exact, the bound in Theorem~\ref{thm-gc} may be loose. It is possible that $\exists \alpha''$ such that $|\langle \nabla_{\unexpp{ }}\uloss{\base{t}}(\uls{\alpha''}),\ftdelta{t}\rangle|$ is large, making $\epsilon$ large. However, if $\alpha'$ is such that $|\langle \nabla_{\unexpp{ }}\uloss{\base{t}}(\uls{\alpha'}),\ftdelta{t}\rangle|\ll |\langle \nabla_{\unexpp{ }}\uloss{\base{t}}(\uls{\alpha''}),\ftdelta{t}\rangle|$, the cost may still be small.
\end{remark}
We empirically find that $\epsilon$ is small for most benchmarks in Figure~\ref{fig:mt_sample}. WinoGrande is an example of Remark~\ref{rmk:lim-gc} where $\epsilon$ overestimates $\cpllm$. Remark~\ref{rmk:lim-gc} arises because $\alpha'\in[0,1]$ is unknown. We leave it for future work to estimate a smaller range for $\alpha'$ to mitigate this limitation. We also find that $|\langle \nabla_{\unexpp{ }}\uloss{\base{t}}(\uls{\alpha'}),\ftdelta{t}\rangle|\ll \|\nabla_{\unexpp{ }}\uloss{\base{t}}(\uls{\alpha'})\| \|\ftdelta{t}\|$, indicating that near-orthogonality, not just small norms, causes $\epsilon$ and $\cpllm$ to be small.

\vspace{2mm}
\noindent\textit{Results for $\bpllm$ (RQ2).}
Here, we are interested in comparing the PortLLM performance $\eloss{\base{t}+\expp{0}}$ to the base model performance $\eloss{\base{t}}$, as quantified in $\bpllm(t)$ defined in~\eqref{benefit-def}. The following Lemma is derived in Appendix~\ref{app:1d-slice-rq2}.
\begin{lemma}\label{lim-gg}
    There exists an $\alpha'\in[0,1]$ such that the benefit of using PortLLM patching instead of no patching is given by $\bpllm(t)=-\langle \nabla_{\unexpp{ }}\uloss{\base{t}}((1-\alpha')\unexpp{0}),\unexpp{0}\rangle$.
\end{lemma}
\begin{remark}\label{rmk:gg_inner}
    The inner product in Lemma~\ref{lim-gg} can be rewritten as $-\langle \nabla_{\unexpp{ }}\uloss{\base{t}}((1-\alpha)\unexpp{0}),\unexpp{0}\rangle=\langle \nabla_{\unexpp{ }}\uloss{\base{t}}(\unexpp{0}-\alpha\unexpp{0}),-\unexpp{0}\rangle$ in the form of a displaced-gradient--displacement product. 
    The key observation here, which explains why $\bpllm$ is large while $\cpllm$ and $\vpllm$ are small, is that $(1-\alpha)\unexpp{0}$ for $\alpha\leq0\leq1$ roughly traces optimization path from $\mathbf{0}$ to $\unexpp{0}$. If $\uloss{\base{0}}$ and $\uloss{\base{t}}$ have similar shapes, the loss landscape on the optimization path $\uloss{\base{t}}((1-\alpha)\unexpp{0})$ is steep, implying that the spectral norm $\sigma_{\text{max}}(\unexpp{0}-\beta'\alpha\unexpp{0})$ is large, except when $\beta'\alpha\approx0$. This steepness is empirically confirmed in the bottom row of Figure~\ref{fig:mt_sample}. Hence, the second metric from Lemma~\ref{lim-dgdp}, the sharpness, is large so that the displaced-gradient--displacement product may be large.
\end{remark}

\vspace{2mm}
\noindent\textit{Results for $\vpllm$ (RQ3).}
We compare PortLLM performance at time $t>0$ to performance at $t=0$. This difference is quantified by $\vpllm(t)$ defined in~\eqref{variation-def}. We present the following Lemma, which is derived in Appendix~\ref{app:1d-slice-rq3}.
\begin{lemma}\label{lim-gd}
    There exists an $\alpha'\in[0,1]$ such that the change in PortLLM performance at time step $t$ is given by $\vpllm(t)=\langle \nabla_{\base{}}\eloss{\els{\alpha'}},\ptdelta{t} \rangle$ where $\els{\alpha}=\alpha\base{0}+(1-\alpha)\base{t}+\expp{0}$ is a point on the 1-D slice between the continually pretrained base model $\base{t}$ and the initial base model $\base{0}$ and $\ptdelta{t}=\base{0}-\base{t}$ is the pretraining update. 
\end{lemma}
If the pretraining and fine-tuning distributions are sufficiently different so that the pretraining update is approximately orthogonal to the fine-tuning gradient, the change in PortLLM performance will be very small. The displaced-gradient--displacement product form can be used to further characterize the inner product.
\begin{remark}\label{rmk:gd_inner}
    The inner product in Lemma~\ref{lim-gd} can be written as a displaced-gradient--displacement product as $\langle \nabla_{\base{}}\lft(\els{\alpha}),\ptdelta{t} \rangle=\langle \nabla_{\base{}}\lft(\base{t}+\expp{0}+\alpha\ptdelta{t}),\ptdelta{t} \rangle$. We consider the two metrics derived from Lemma~\ref{lim-dgdp}. First, because the pretraining and fine-tuning tasks are not strongly related, the displacement direction $\ptdelta{t}/\|\ptdelta{t}\|_2$, or the direction of the pretraining update, is expected to be approximately orthogonal to the fine-tuning gradient so that the alignment is small. Second, the loss landscape is plausibly flat around the reference $\base{t}+\expp{0}$, which is a fine-tuned model, so that the sharpness is small. These observation explain why we observe small variation in PortLLM performance $\vpllm$ in Section~\ref{exp-results-main-text}. 
\end{remark}
We present a theorem similar to Theorem~\ref{thm-gc} as follows:
\begin{theorem}\label{thm-gd}
    Let $\epsilon=\max_{\alpha\in[0,1]}| \langle \nabla_{\base{}}\eloss{\els{\alpha}},\ptdelta{t} \rangle|$ be the maximum derivative magnitude on the 1-D slice between the continually pretrained base model $\base{t}$ and the initial base model $\base{0}$ where $\els{\alpha}=\alpha\base{0}+(1-\alpha)\base{t}+\expp{0}$ is a point on the 1-D slice and $\ptdelta{t}=\base{0}-\base{t}$ is the pretraining update. Then the change in performance is bounded as $|\vpllm(t)|\leq \epsilon$. 
\end{theorem}

While our analysis above focuses on the PortLLM scenario, it is not limited to this scenario. Our analysis method, namely, defining a scalar function to characterize a 1-D slice of the fine-tuning loss landscape, can be used to compare performance of two arbitrary LoRA patches on a given task with an associated loss function.

In Figure~\ref{fig:mt_sample} and Appendix~\ref{app:1d-slice-exp}, we empirically plot the 1-D slices of the loss landscape associated with the cost $\cpllm$ and the benefit $\bpllm$ based on~\eqref{gc-derivative} and~\eqref{gg-derivative-app} for Mistral pretrained on Fineweb and fine-tuned on WinoGrande, BoolQ, ARC-Easy, and ARC-Challenge.\footnote{Compute constraints prevent us from calculating the derivative of the 1-D slice for the variation metric $\vpllm$, which would require calculating gradients in base model parameter space. We offer additional theoretical analysis of $\vpllm$ in Section~\ref{pt-iter-analysis}.} In the top row of Figure~\ref{fig:mt_sample}, the loss curve $\gc{\alpha}$ is flat so that the derivative $\gc{\alpha}$ is small for most benchmarks. This observation of flatness in the loss landscape around fine-tuned models aligns with~\cite{chen2025}, who argue that the loss landscape of LLMs exhibits basin-like structures. The results for WinoGrande demonstrate Remark~\ref{rmk:lim-gc}, which notes that Theorem~\ref{thm-gc} can overestimate the cost $\cpllm$. While the derivative $\gcp{\alpha}$ is large at points so that $\epsilon$ is large, the change in loss is not dramatic so that $\cpllm$ is not too large. In the bottom row of Figure~\ref{fig:mt_sample}, $\gb{\alpha}$ defined in~\eqref{gg-def-app} of Appendix~\ref{app:1d-slice-rq2} is analogous to $\gc{\alpha}$ for the benefit metric $\bpllm(t)$. As predicted in Remark~\ref{rmk:gg_inner}, the derivative $\gbp{\alpha}$ is large for $0.5<\alpha<1$.

\subsection{Analysis Based on Continual Pretraining Optimization Steps}\label{pt-iter-analysis}
We further leverage information about the continual pretraining process to provide a more informative bound on the variation in fine-tuning performance $\vpllm(t)$ as the base model is continually pretrained. The goal is to compare LoRA-adapted performance for a fixed LoRA patch $\expp{}$ on the initial base model $\lft(\base{0}+\expp{})$ to LoRA-adapted performance on the continually pretrained base mode $\lft(\base{t}+\expp{})$. We achieve this comparison by starting at the initial base model and considering how each iteration of continual pretraining affects downstream fine-tuning loss. We apply this analysis to a PortLLM patch $\expp{0}$, but it could be applied in other scenarios, i.e., LoRA variants. 

To aid this analysis, we formalize the notion of approximate orthogonality. Quasi-orthogonality as a relaxation of strict orthogonality has been defined in various fields with respect to different mathematical objects, e.g., quasi-orthogonal random fields~\citep{Moricz1989}, quasi-orthogonal families of binary sequences~\citep{Yang2000}, and quasi-orthogonal polynomials~\citep{Chihara1957}. Following~\cite{svozil2026} and~\cite{kainen2020}, we defined $\epsilon\text{-}$quasi-orthogonality with respect to vectors as follows:
\begin{definition}
    Let $\mathbf{a},\mathbf{b}\in \real^n$ be two $n\text{-}$dimensional real vectors. $\mathbf{a}$ and $\mathbf{b}$ are $\epsilon\text{-}$quasi-orthogonal for $\epsilon \in (0,1)$ if the magnitude of their cosine similarity is upper-bounded by $\epsilon$, namely,\footnote{In~\cite{svozil2026} and~\cite{kainen2020}, the definition of quasi-orthogonality is restricted to unit norm vectors. We extend the definition to allow any nonzero real vectors $\mathbf{a},\mathbf{b}\in \real^n$.} $|\langle \mathbf{a},\mathbf{b}\rangle|/(\|\mathbf{a}\|_2 \|\mathbf{b}\|_2)\leq \epsilon.$
\end{definition}
We state our assumptions for this analysis below. 
\begin{assumption}\label{amt-orthogonal}
    There exists an $\epsorth \geq 0$ such that the pretraining gradient $\nabla \lpt(\base{})$ and the PortLLM-adapted fine-tuning gradient $\nabla\lft(\base{}+\expp{0})$ are $\epsorth\text{-}$quasi-orthogonal $\forall\base{}\in\real^N$.
\end{assumption}
While this assumption may appear strong, we argue that it holds for a small $\epsorth$ in most practical scenarios. The traits of high-dimensional space are such that independent random vectors in high-dimensional space are (approximately) orthogonal. See Appendix~\ref{app:quasiorthogonal} for more detailed discussion and Table~\ref{tab:cos_sim} for numeric examples. 
If the pretraining and fine-tuning gradients are conditionally independent conditioned on the initial base model $\base{0}$, then the gradients are likewise approximately orthogonal. 

An additional assumption describes how continual pretraining updates are obtained.
\begin{assumption}\label{amt-sgd}
    For each time step $t>0$, the model $\base{t}$ is obtained with $I$ iterations of standard gradient descent starting from the model $\base{t-1}$. Each iteration uses learning rate $\lr$. At each iteration $i$, when training $\base{t}$, let the model be denoted as $\iter{i}$.
\end{assumption}
While we assume standard gradient descent for simplicity, our proof would hold in more general cases. Our proof depends only on the norm of the update $\|\iter{i}-\iter{i-1} \|_2$ being small, but not on the exact value of model parameters at each iteration. Assumptions~\ref{amt-stable} and~\ref{amt-small-lr} ensure that this norm is small. 
\begin{assumption}\label{amt-stable}
    The continual pretraining process is stable such that the gradient norm at each iterate is bounded as $\|\nabla \lpt\big(\iter{i}\big)\|_2\leq \lippt$ for all $i\in\{0,1,\dots,I\}$.\footnote{This assumption explains why our analysis applies to continual pretraining but not to repeated random perturbations of base model parameters. Repeated random perturbations are likely to move the model to a sharp place in the loss landscape where gradients are large. } 
\end{assumption}
This assumption, essentially claiming that gradient explosion does not occur, is a more realistic alternative to assuming global Lipschitzness.

Intuitively, we expect that the models obtained through continual pretraining are reasonably conditioned for the downstream task such that initial performance before fine-tuning is not catastrophically bad.\footnote{The idea that pretraining conditions the model for fine-tuning is supported by~\cite{sun2025}, who show that, depending on the downstream task, pretraining may either (1) improve base model performance on the downstream task or (2) adjust the base model such that fine-tuning is more effective on the downstream task.} Based on this idea and an assumption that the fine-tuning loss function is well-behaved,\footnote{The claim that the fine-tuning loss function is well-behaved is supported by~\cite{das2026}, who show that cross-entropy loss on a transformer model with GELU activation is infinitely differentiable. Their larger proof depends on use of $L^2$ regularization, but their claim about infinite differentiability does not depend on this regularization.} we conjecture that the fine-tuning gradients are bounded in a region around the pretrained models. 
\begin{assumption}\label{amt-ft-grad}
    The pretrained models are reasonably conditioned for the downstream task such that the fine-tuning gradient norm is bounded in a neighborhood around the pretraining iterates. Namely,
    \begin{equation}
        \max_{\base{}\in R}\|\nabla \lft(\base{})\|_2= \lipft
    \end{equation}
    where $R=\{x\in\real^N:\exists i\in\{1,2,\dots,I\} \text{ such that } \|x-\iter{i}\|_2 \leq d\}$. Further, we assume that the PortLLM patch norm is bounded as $\|\expp{0}\|_2\leq d$. 
\end{assumption}
This, of course, is not to claim that the model performs well on the downstream task before fine-tuning. Rather, we claim that the pretrained models provide a well-conditioned starting point for fine-tuning. 

\begin{assumption}\label{amt-small-lr}
    The distance between iterates $\|\iter{i}-\iter{i-1}\|_2=\|-\lr\nabla\lpt(\iter{i-1})\|_2\leq \lr \lippt$ is sufficiently small so that a first order Taylor series approximation is sufficient to approximate the PortLLM-adapted fine-tuning loss at the next iteration with a small error $R_2$, specifically,
    \begin{equation}
        \lft\big(\iter{i}+\expp{0}\big) = \lft\big(\iter{i-1}+\expp{0}\big)+\langle \nabla \lft\big(\iter{i-1}+\expp{0}\big), -\lr\nabla \lpt\big( \iter{i-1}\big) \rangle+R_2, \hspace{0.5em} |R_2|\leq \epslr
    \end{equation}
    where $R_2$ is the error of neglecting higher-order terms and $\epslr$ is a bound on the error.
\end{assumption}
See Remark~\ref{rmk-epslr} in Appendix~\ref{app:train-iterate-proof} for a characterization of $\epslr$ where we find that it is on the order of $\frac{1}{2}\lr^2 e^2 \lippt^2\sum_{j=1}^N \sigma_j$, where $e\ll1$ is a quasi-orthogonality constant and $\{\sigma_j\}_{j=1}^N$ are singular values of the fine-tuning Hessian. We prove the following theorem in Appendix~\ref{app:train-iterate-proof}.
\begin{theorem}\label{thm-rq3}
    Let Assumptions~\ref{amt-orthogonal}--\ref{amt-small-lr} hold. Then the magnitude of the difference $|\vpllm(t)|$ between PortLLM performance at time stem $t$ and time step $0$ is bounded as
    \begin{equation}
        |\vpllm(t)|=|\lft\big( \base{t}+\expp{0}\big)-\lft\big( \base{0}+\expp{0}\big)| \leq It(\lr \epsorth \lippt \lipft+\epslr)
    \end{equation}
    where $\lr$ is the learning rate during continual pretraining, $I$ is the number of gradient descent iterations, $\epsorth$ is the orthogonality constant in Assumption~\ref{amt-orthogonal}, $\epslr$ defined in Assumption~\ref{amt-small-lr} is the error due to using a first order Taylor series approximation, $\lippt$ is the pretraining stability constant defined in Assumption~\ref{amt-stable}, and $\lipft$ is the bound on fine-tuning gradients from Assumption~\ref{amt-ft-grad}. 
\end{theorem}
Some remarks on Theorem~\ref{thm-rq3} are instructive.
\begin{remark}\label{rmk-applicability}
    Assumptions~\ref{amt-orthogonal}--\ref{amt-small-lr} do not specify the LoRA fine-tuning algorithm. Hence, our analysis is applicable to LoRA variants, e.g., CAR-LoRA~\citep{shahroz2026}. 
\end{remark}
\begin{remark}
    We assume standard gradient descent in Assumption~\ref{amt-sgd} for simplicity, but this could be generalized to include more practical training algorithms. If the updates at each iteration $i$ are perturbed by an error vector $\xi_{i-1}$ so that $\iter{i}-\iter{i-1}=-\lr \tilde{ \nabla \lpt}(\iter{i-1})=-\lr(\nabla\lpt(\iter{i-1})+\xi_{i-1})$, then both $\nabla \lpt(\iter{i-1})$ and $\xi_{i-1}$ are approximately orthogonal to $\nabla\lft(\iter{i-1}+\expp{0})$ due to the traits of high-dimensional spaces. Results similar to Theorem~\ref{thm-rq3} could be obtained using the same proof techniques.
\end{remark}
\begin{remark}
    Gradient clipping, which is common in practical scenarios, could replace Assumption~\ref{amt-stable} on pretraining stability to bound the distance between iterates. 
\end{remark}

\section{Discussion}
Near-zero inner products of high-dimensional vectors appear repeatedly in core steps of our theoretical analyses, revealing that traits of high-dimensional spaces improve temporal portability. In Section~\ref{1d-slice} that presents analysis based on a 1-D slice of the loss landscape, the cost $\cpllm$ of PortLLM compared to stepwise fine-tuning, the benefit $\bpllm$ of PortLLM compared to no patching, and the variation $\vpllm$ of PortLLM performance across time steps can each be written as a single inner product. For $\cpllm$ and $\vpllm$, these inner products are small because their corresponding vectors are approximately orthogonal. Furthermore, the first term $\gamma\langle\nabla f(x_0) ,\Delta x\rangle$ in the upper bound in Lemma~\ref{lim-dgdp} for displaced-gradient--displacement products is an inner product.
In Section~\ref{pt-iter-analysis} that leverages the relationships between pretraining optimization steps to characterize RQ3, our results depend on Assumption~\ref{amt-orthogonal} that the pretraining and fine-tuning gradients are approximately orthogonal to bound their inner product. These cases demonstrate that near-orthogonality is a key justification for temporal portability.
Future work may consider how the mathematical device of near-orthogonality may be generally applicable to other other scenarios, e.g., improving analysis of fine-tuning strategies or considering how different LoRA patches for different tasks can be jointly used.

\section{Conclusion}\label{conclusion}
We have shown that the temporal portability of LoRA patches observed in~\cite{Khan2025} extends across our observation span of $10$ time steps. The PortLLM method performs similarly to stepwise fine-tuned patches trained at the time step of interest (RQ1), is able to substantially outperform no patching (RQ2), and approximately maintains $t=0$ performance (RQ3). We have provided two theoretical analysis methods to explain this surprising results. These methods reveal that near-orthogonality in high-dimensional spaces explains temporal portability. In this scenario, the ``curse'' of dimensionality is actually a blessing. 
\appendix

\section{Analysis for 1-D Slice of Loss Landscape}\label{app:1d-slice-analysis}
\subsection{Formalized Definitions}\label{app:definitions}
We begin with more formal definitions for the expanded and unexpanded vectorized LoRA patch parameters and the associated loss functions, which were introduced in Section~\ref{definitions}. Let $\base{ }\in\real^N$ be the parameters for a model of size $N$. Assume the model contains $m$ weight matrices, given by the ordered tuple $(W_i)_{i=1}^m$. In the LoRA paradigm~\citep{hu2022}, associated with each $W_i\in\real^{d_i\times k_i}$ are $A\in \real^{r \times k_i}$ and $B\in\real^{d_i\times r}$ so that the weight $W_i$ is updated as $W_i+B_iA_i$, where $r$ is the LoRA rank. A LoRA patch can be described by the ordered tuple of ordered pairs $((A_i,B_i))_{i=1}^m$. Denoting the vectorizing operation as $\vfunc(\cdot)$, the following vectors can be defined as concatenations of vectorized LoRA matrices
\begin{subequations}\label{patch-defs}
    \begin{align}
        \unexpp{ }&=[\vfunc(B_1),\vfunc(A_1), \dots ,\vfunc(B_m),\vfunc(A_m)] \\
        \expp{ }&=[\vfunc(B_1A_1),\dots,\vfunc(B_mA_m)].
    \end{align}
\end{subequations}
Here, $\unexpp{ }\in \real^n$ is the unexpanded patch parameters, where $n=r\sum_{i=1}^m(d_i+k_i)$ is the number of LoRA trainable parameters. $\expp{ }\in \lrank{r}\subset \real^N$ is the expanded LoRA parameters, where $\lrank{r}=\{[\vfunc(B_1A_1),\dots,\vfunc(B_mA_m)]: B_i\in \real^{{d_i\times r}},A\in \real^{r \times k_i}\}$ is a subset of $\real^N$ subject to LoRA rank restrictions. Let base model parameters at continual pretraining time step $t$ be denoted as $\base{t}$. Similarly, patches trained on $\base{t}$ are denoted as $\unexpp{t}$ and $\expp{t}$ for unexpanded and expanded patch parameters, respectively.

The loss on the base model at time $t$ is given by $\lft(\base{t})$ and the loss on the stepwise fine-tuned model is given by $\lft(\base{t}+\expp{t})$, where $\lft$ is the downstream fine-tuning loss defined in Section~\ref{definitions}. Let $\uloss{\base{}}: \real^n\rightarrow \real$ be the downstream testing loss function applied to unexpanded patch parameters on base model $\base{}$. Specifically, for an ordered tuple of LoRA patch parameters $((A_i,B_i))_{i=1}^m$ with  $\unexpp{}=[\vfunc(B_1),\vfunc(A_1), \dots ,\vfunc(B_m),\vfunc(A_m)]$ as the unexpanded vectorized patch and $\expp{}=[\vfunc(B_1A_1),\dots,\vfunc(B_mA_m)]$ as the expanded patch, we define $\uloss{\base{}}(\unexpp{})=\lft(\base{}+\expp{})$. 

\subsection{Analysis for \texorpdfstring{$\bpllm$}{BPortLLM} (RQ2).}\label{app:1d-slice-rq2}
\begin{figure}{}{}
    \centering
    \includegraphics[trim={0mm 42mm 0mm 73mm},clip,width=1\textwidth]{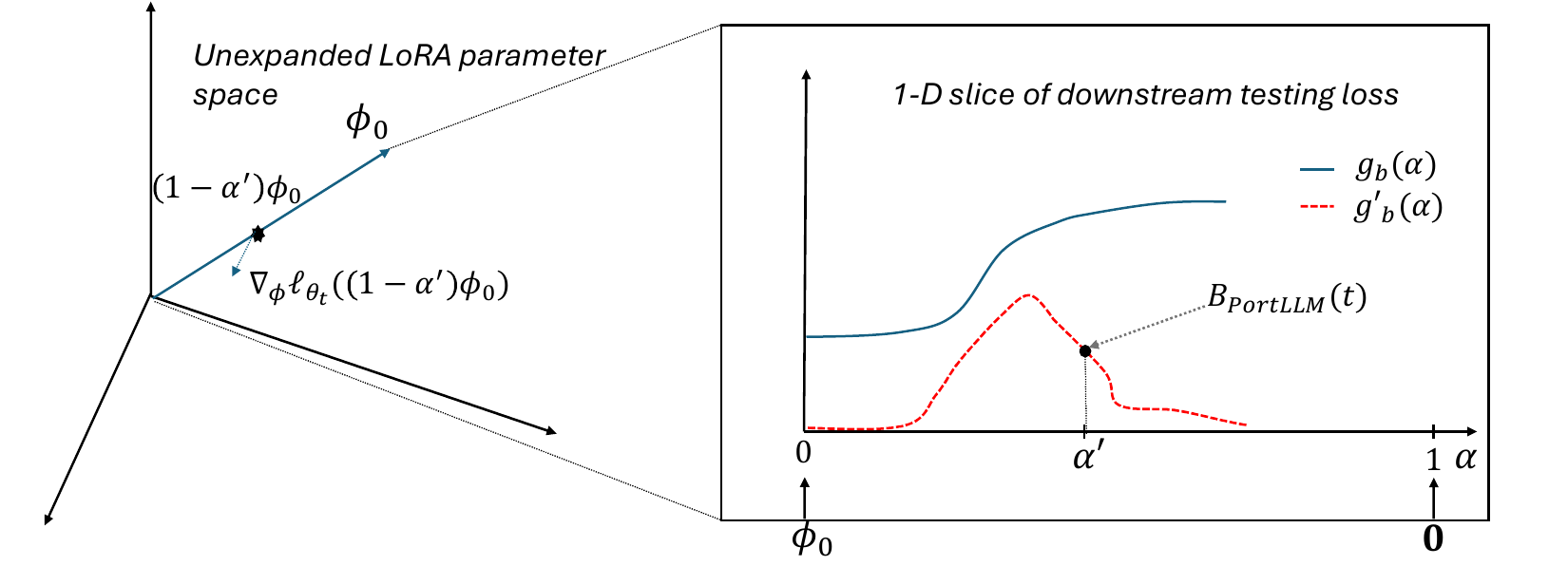}
    \caption{Illustration of analysis for the benefit of employing PortLLM instead of no patching, $\bpllm{}(t)$. The PortLLM patch $\unexpp{0}$ is shown as blue arrows in unexpanded LoRA parameter space on the left. The downstream testing loss on a 1-D slice between the PortLLM patch and no patching is shown on the right. The loss is shown as a blue solid line while the derivative of the loss is shown as a red dashed line. The derivative is small for $\alpha \approx 0$ and large for $\alpha \approx 0.5$.  $\alpha'$ is the point on the 1-D slice at which the derivative exactly predicts the benefit, $\bpllm(t)$. $\alpha'$ on the 1-D slice corresponds to $(1-\alpha')\unexpp{0}$ in parameter space. If, as depicted, the gradient points roughly opposite to $\unexpp{0}$, the benefit is positive.}\label{benefit_illustration}
\end{figure}

Here we derive the results presented in Section~\ref{1d-slice} for RQ2. We are interested in comparing the PortLLM performance $\lft(\base{t}+\expp{0})$ to the base model performance $\lft(\base{t})$ at an arbitrary fixed time step $t>0$. This scenario is depicted in Figure~\ref{benefit_illustration}. In contrast to analysis for RQ1 in Section~\ref{1d-slice}, where we compare two LoRA patches $\unexpp{0}$ and $\unexpp{t}$, here we are interested in comparing the PortLLM patch $\unexpp{0}$ to the base model, which can be represented by the all-zero patch $\mathbf{0}\in\real^n$. Define the loss on a 1-D slice between the unexpanded PortLLM patch parameters $\unexpp{0}$ and the all-zero patch $\mathbf{0}$ as
\begin{equation}\label{gg-def-app}
    \gb{\alpha}=\uloss{\base{t}}\big((1-\alpha)\unexpp{0}\big)
\end{equation}
for $\alpha\in\mathbb{R}$, but $\alpha\in[0,1]$ is most meaningful. In Figure~\ref{benefit_illustration}, $\gb{\alpha}$ is shown as a blue solid line and its derivative is shown as a red dashed line. Note that $\gb{1}=\uloss{\base{t}}(\mathbf{0})$ describes base model performance and $\gb{0}=\uloss{\base{t}}(\unexpp{0})$ describes PortLLM performance so that $\bpllm(t)=\gb{1}-\gb{0}$ from~\eqref{benefit-def}. Taylor's theorem gives that, for some $\alpha'\in[0,1]$
\begin{equation}\label{gg-taylor-app}
    \gb{1}=\gb{0}+(1-0)\gbp{\alpha'}
\end{equation}
so that $\gb{1}-\gb{0}=\bpllm(t)=\gbp{\alpha'}$. The derivative can be written as
\begin{subequations}\label{gg-derivative-app}
    \begin{align}
        \gbp{\alpha}&=\frac{d}{d\alpha}\uloss{\base{t}}\big((1-\alpha) \unexpp{0}\big) = \nabla_{\unexpp{ }}\uloss{\base{t}}\big((1-\alpha) \unexpp{0}\big)^\top \frac{d}{d\alpha} \big((1-\alpha) \unexpp{0}\big) \\
        &=-\langle \nabla_{\unexpp{ }}\uloss{\base{t}}((1-\alpha)\unexpp{0}),\unexpp{0}\rangle
    \end{align}
\end{subequations}
From this, the benefit is given by $\bpllm(t)=-\langle \nabla_{\unexpp{ }}\uloss{\base{t}}((1-\alpha)\unexpp{0}),\unexpp{0}\rangle$ for a fixed but unknown $\alpha'\in[0,1]$. This proves Lemma~\ref{lim-gg}. 

\subsection{Analysis for \texorpdfstring{$\vpllm$}{DPortLLM} (RQ3).}\label{app:1d-slice-rq3}
\begin{figure}{}{}
    \centering
    \includegraphics[trim={0mm 52mm 0mm 67mm},clip,width=1\textwidth]{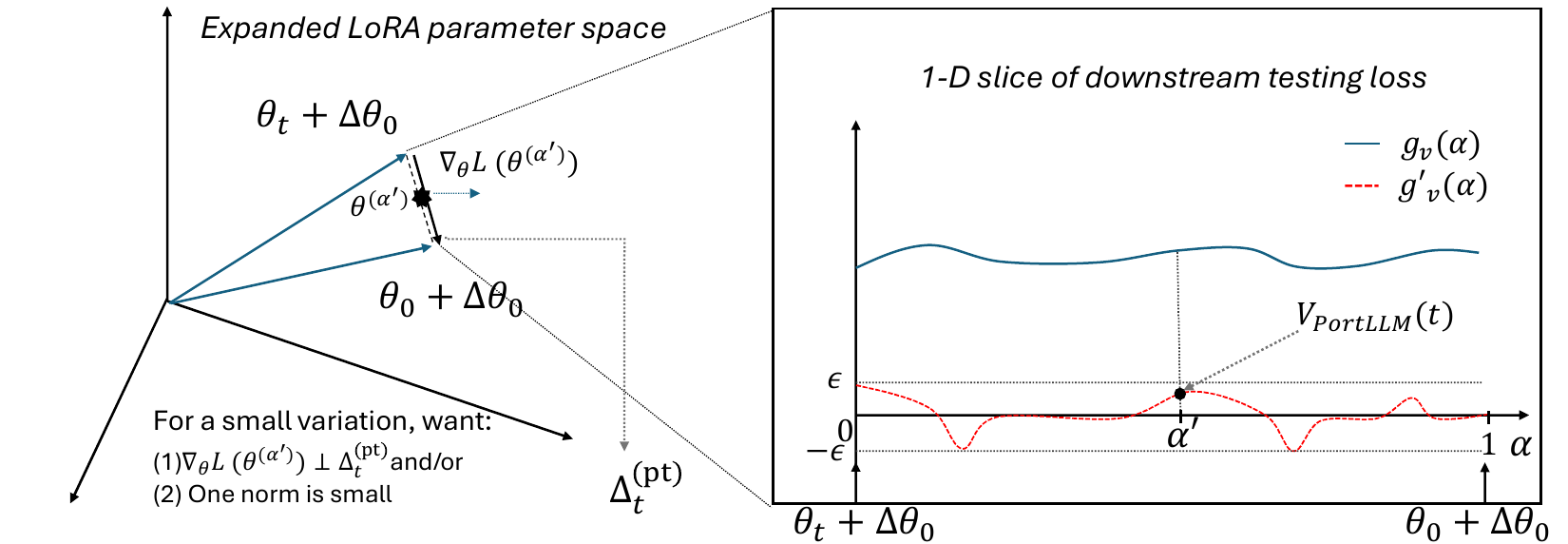}
    \caption{Geometric interpretation for the variation of PortLLM performance, $\vpllm{}(t)$. The initial PortLLM parameters $\base{0}+\expp{0}$ and the PortLLM parameters at time $t>0$, $\base{t}+\expp{0}$ are shown as blue arrows in expanded LoRA parameter space on the left. The downstream testing loss on a 1-D slice between the two parameters is shown on the right. The loss, shown as a blue solid line, is approximately constant so that the derivative, shown as a red dashed line, has magnitude bounded by $\epsilon$. $\alpha'$ is the point on the 1-D slice at which the derivative exactly predicts the variation, $\vpllm(t)$. $\alpha'$ on the 1-D slice corresponds to $\els{\alpha'}$ in parameter space. If, as depicted, the gradient $\nabla_{\base{}} \eloss{\els{\alpha'}}$, is approximately orthogonal to the error between the two patches, the variation $\vpllm(t)$, is small.}\label{variation_illustration}
    \vspace{-5mm}
\end{figure}

Now we derive the results presented in Section~\ref{1d-slice} related to RQ3. We are interested in comparing PortLLM performance at time $t>0$ to performance at $t=0$. In this case, pretraining updates must be considered, so the 1-D slice has to be defined in $N$-dimensional space in terms of expanded parameters. This case is shown in Figure~\ref{variation_illustration}, where expanded parameter space is depicted on the left. Define the function

\begin{equation}\label{gd-def-app}
    \gv{\alpha}=\lft(\alpha\base{0}+(1-\alpha)\base{t}+\expp{0}),
\end{equation}
where $\alpha\in\real$ but $\alpha\in[0,1]$ is most meaningful so that $\gv{0}=\lft(\base{t}+\expp{0})$ is the PortLLM performance at time step $t$ and $\gv{1}=\lft(\base{0}+\expp{0})$ is the PortLLM performance at time step $0$. Then~\eqref{variation-def} can be rewritten as $\vpllm(t)=\gv{1}-\gv{0}$. By Taylor's theorem,
\begin{equation}\label{gd-taylor-app}
    \gv{1}=\gv{0}+(1-0)\gvp{\alpha'}
\end{equation}
for some $\alpha'\in[0,1]$. This gives that $\vpllm(t)=\gv{1}-\gv{0}=\gvp{\alpha'}$ for some $\alpha'\in[0,1]$. This equality for an example $\alpha'$ is shown in the left side of Figure~\ref{variation_illustration}. To simplify notation, define $\els{\alpha}=\alpha\base{0}+(1-\alpha)\base{t}+\expp{0}$ as a point on the line between the models. Using the chain rule, the derivative can be written as

\begin{subequations}\label{gd-derivative-app}
    \begin{align}
        \frac{d}{d\alpha}\gv{\alpha}&=\frac{d}{d\alpha}\lft(\els{\alpha}) = \nabla_{\base{}}\lft(\els{\alpha})^\top \frac{d}{d\alpha} (\alpha\base{0}+(1-\alpha)\base{t}+\expp{0}) \\
        &=\langle \nabla_{\base{}}\lft(\els{\alpha}),\ptdelta{t} \rangle
    \end{align}
\end{subequations}
where $\ptdelta{t}=\base{0}-\base{t}$ represents the pretraining update and the superscript (pt) abbreviates ``pretraining.'' Now the variation in PortLLM performance across time steps is given by $\vpllm(t)=\langle \nabla_{\base{}}\lft(\els{\alpha'}),\ptdelta{t} \rangle$, which proves Lemma~\ref{lim-gd}. 

\section{Analysis Based on Pretraining Optimization Steps}\label{app:train-iterate-proof}
\subsection{Characterization of \texorpdfstring{$\epslr$}{epsilon\_eta}}
While $\epslr$ could be bounded by the largest singular value of the fine-tuning Hessian matrix, we argue that the largest singular value will not be activated by the pretraining gradient in most cases. Instead of this potentially loose bound, we offer the following tighter characterization. 
\begin{remark}\label{rmk-epslr}
    By Taylor's theorem, $\exists \alpha' \in[0,1]$ such that
    \begin{equation}
        R_2=\frac{1}{2}\lr^2 \nabla \lpt(\iter{i-1})^\top \nabla^2 \lft\big(\iter{i-1}-\alpha'\lr\nabla \lpt(\iter{i-1})+\expp{0}\big)\nabla \lpt(\iter{i-1})
    \end{equation}
    where $\nabla^2 \lft(\cdot)$ denotes the Hessian of the fine-tuning loss with respect to base model parameters $\base{}$. We can further characterize this error by considering singular value decomposition of the Hessian. Specifically, consider the singular value decomposition for a fixed but arbitrary iteration $i\in\{1,2,\dots,I\}$:
    \begin{equation}
        \begin{split}
            \nabla^2 \lft\big(\iter{i-1}-\alpha'\lr\nabla \lpt(\iter{i-1})+\expp{0}\big)&=U\Sigma V^\top = \sum_{j=1}^N \sigma_ju_jv_j^\top.
        \end{split}
    \end{equation}
    The second order error can be analyzed as
    \begin{equation}
        \begin{split}
            R_2&=\frac{1}{2}\lr^2 \nabla \lpt(\iter{i-1})^\top \Big[\sum_{j=1}^N \sigma_ju_jv_j^\top\Big]\nabla \lpt(\iter{i-1}) \\
            &=\frac{1}{2}\lr^2 \sum_{j=1}^N \big[\sigma_j \nabla \lpt(\iter{i-1})^\top u_jv_j^\top\nabla \lpt(\iter{i-1})\big] \\
            &=\frac{1}{2}\lr^2 \sum_{j=1}^N \big[\sigma_j \langle\nabla \lpt(\iter{i-1}), u_j\rangle  \langle v_j,\nabla \lpt(\iter{i-1})\rangle\big] \\
        \end{split}
    \end{equation}
    If the pretraining and fine-tuning tasks are largely unrelated, then, in high dimensional space, we expect that the pretraining gradient and the singular values of the fine-tuning Hessian are approximately orthogonal. If this expectation holds so that the pretraining gradient $\nabla \lpt(\iter{i-1})$ is $e\text{-}$quasi-orthogonal to each singular value in $\{u_1,u_2,\dots,u_N,v_1,v_2,\dots,v_N\}$ then the $R_2$ error can be bounded as
    \begin{equation}
        \begin{split}
            R_2&=\frac{1}{2}\lr^2 \sum_{j=1}^N \big[\sigma_j \langle\nabla \lpt(\iter{i-1}), u_j\rangle  \langle v_j,\nabla \lpt(\iter{i-1})\rangle\big] \leq \frac{1}{2}\lr^2 e^2 \lippt^2\sum_{j=1}^N \sigma_j.
        \end{split}
    \end{equation}
    Given the definition of $\epslr$ in Assumption~\ref{amt-small-lr}, the quantity $\frac{1}{2}\lr^2 e^2 \lippt^2\sum_{j=1}^N \sigma_j$ characterizes the order of magnitude of $\epslr$. While the sum of singular values may be quite large, both $\lr^2$ and $e^2$ are expected to be very small. If $e=O\big([\sum_{j=1}^N \sigma_j]^{-1/2}\big)$, then $\epslr=O(\lr^2\lippt^2)$ so that the error decreases quadratically with the learning rate. 
\end{remark}

\subsection{Proof of Theorem~\ref{thm-rq3}}
We prove Theorem~\ref{thm-rq3} presented in Section~\ref{pt-iter-analysis}. 

Let Assumptions~\ref{amt-orthogonal}--\ref{amt-small-lr} hold. As in Assumption~\ref{amt-sgd}, let the base model $\base{t}$ be continually pretrained from $\base{t-1}$ as follows. Set $\iter{0}=\base{t-1}$. For each iteration $i\in\{1,2,\dots,I\}$, let the model be updated as
\begin{equation}
    \iter{i}=\iter{i-1}-\lr \nabla \lpt(\iter{i-1})
\end{equation}
where $\lr$ is the learning rate. For each iteration $i\in\{1,2,\dots,I\}$,
\begin{subequations}
    \begin{align}
        &\lft\big(\iter{i} +\expp{0}\big)=\lft\big(\iter{i-1}-\lr \nabla \lpt(\iter{i-1}) +\expp{0} \big) \\
        &= \lft\big( \iter{i-1}+\expp{0})\big)+\langle \nabla \lft \big( \iter{i-1}+\expp{0}\big) ,-\lr \nabla \lpt(\iter{i-1} \rangle+R_2 \\
        & \leq \lft\big( \iter{i-1}+\expp{0})\big)+\epsorth \lr\|\nabla \lft \big( \iter{i-1}+\expp{0}\big)\|_2 \|\nabla \lpt(\iter{i-1}\|_2 +\epslr,
    \end{align}
\end{subequations}
where the equality is a first order Taylor series approximation with error $R_2$ and the inequality follows from Assumption~\ref{amt-small-lr} which bounds $R_2$ and  Assumption~\ref{amt-orthogonal} which bounds the cosine similarity of the gradients. Further, we use Assumptions~\ref{amt-stable} and~\ref{amt-ft-grad} to bound the gradients as follows:
\begin{equation}
    \lft\big(\iter{i} +\expp{0}\big) \leq \lft\big( \iter{i-1}+\expp{0})\big)+\lr\epsorth \lippt \lipft +\epslr
\end{equation}
Repeating $I$ times,
\begin{subequations}
    \begin{align}
        \lft\big(\iter{I} +\expp{0}\big)&=\lft\big(\base{t} +\expp{0}\big) \\
        \leq  \lft\big(\iter{0} +\expp{0}\big)+I(\lr \epsorth \lippt \lipft+\epslr)&=\lft\big(\base{t-1} +\expp{0}\big)+I(\lr \epsorth \lippt \lipft+\epslr)
    \end{align}
\end{subequations}
so that the difference in PortLLM loss $|\lft\big( \base{t}+\expp{0}\big)-\lft\big( \base{t-1}+\expp{0}\big)|$ is bounded as
\begin{equation}
    |\lft\big( \base{t}+\expp{0}\big)-\lft\big( \base{t-1}+\expp{0}\big)| \leq I(\lr \epsorth \lippt \lipft+\epslr).
\end{equation}
Repeating $t$ times and applying the triangular inequality, we find that
\begin{equation}
    |\lft\big( \base{t}+\expp{0}\big)-\lft\big( \base{0}+\expp{0}\big)| \leq It(\lr \epsorth \lippt \lipft+\epslr).
\end{equation}
This completes the proof of Theorem~\ref{thm-rq3}.

\section{Analysis of Displacement-Gradient-Displacement Products}\label{app:dis_grad_inner}
We derive Lemma~\ref{lim-dgdp} from Section~\ref{1d-slice}. By Taylor's theorem, there exists a $\beta'\in[0,1]$ such that $\nabla f(x_0+\gamma\Delta x) = \nabla f(x_0) + \gamma \nabla^2 f(x_0+\beta'\gamma\Delta x) \Delta x$ where $\nabla^2f$ denotes the Hessian matrix so that 
\begin{subequations}
    \begin{align}
        F(\gamma \Delta x) &= \langle\nabla f(x_0) + \gamma \nabla^2 f(x_0+\beta'\gamma\Delta x) \Delta x,\gamma\Delta x\rangle \\
        & = \langle\nabla f(x_0) ,\gamma\Delta x\rangle+\langle\gamma \nabla^2 f(x_0+\beta'\gamma\Delta x) \Delta x,\gamma\Delta x\rangle \\
        & = \underbrace{\gamma\langle\nabla f(x_0) ,\Delta x\rangle}_{\text{T1}}+\underbrace{\gamma^2 \Delta x^\top\nabla^2 f(x_0+\beta'\gamma\Delta x)\Delta x}_{\text{T2}}.
    \end{align}
\end{subequations}
Note that T2 can be bounded by the spectral norm of the Hessian as
\begin{equation}
    F(\gamma \Delta x)\leq \gamma\langle\nabla f(x_0) ,\Delta x\rangle+\gamma^2\sigma_{\text{max}}(x_0+\beta'\gamma\Delta x),
\end{equation}
where $\sigma_\text{max}(x)=\|\nabla^2f(x)\|_2$ is the spectral norm. This result is Lemma~\ref{lim-dgdp} in Section~\ref{1d-slice}.

\section{Quasi-orthogonality in High Dimensional Spaces}\label{app:quasiorthogonal}
We discuss a trait of high-dimensional spaces, specifically, that independent random vectors in high-dimensional spaces are approximately orthogonal. This idea is important because it explains why $\epsorth$ is small in Assumption~\ref{amt-orthogonal} and $\epslr$ is small in Remark~\ref{rmk-epslr}. Table~\ref{tab:cos_sim} gives simulation results for random vectors independently drawn from a standard normal distribution. For each vector dimension $M$, $10^4$ pairs of white Gaussian vectors are used to calculate inner product and cosine similarity values. We report the standard deviation around the zero mean. The standard deviation of the cosine similarity decreases with increasing vector dimension so that each cosine similarity is close to zero with high probability. In contrast, the inner product standard deviations increase with increasing dimension. These trends motivate our concern to bound the norms of the vectors of interest in Assumption~\ref{amt-stable} and~\ref{amt-ft-grad} as well as their cosine similarity.
\begin{table}[t]
    \centering
    \begin{tabular}{p{2cm} p{2.2cm} p{2cm}}
    \hline
    $M$ & $\text{std}\Big(\frac{\langle \mathbf{a},\mathbf{b} \rangle}{\|\mathbf{a}\|_2\|\mathbf{b}\|_2}\Big)$ & $\text{std}(\langle \mathbf{a},\mathbf{b} \rangle)$ \\
    \hline
    $10^{3}$ & $3.2\times 10^{-2}$ & $3.2\times 10^{1}$ \\
    $10^{4}$ & $1.0\times 10^{-2}$ & $4.8\times 10^{-1}$ \\
    $10^{5}$ & $3.2\times 10^{-3}$ & $1.7\times 10^{0}$ \\
    $10^{6}$ & $1.0\times 10^{-3}$ & $1.0\times 10^{3}$ \\
    $10^{7}$ & $3.2\times 10^{-4}$ & $3.2\times 10^{3}$ \\
    $10^{8}$ & $9.9\times 10^{-5}$ & $9.9\times 10^{3}$ \\
    $10^{9}$ & $3.2\times 10^{-5}$ & $3.2\times 10^{4}$ \\
    \hline
    \end{tabular}
    \caption{Inner product and cosine similarity for random vectors $\mathbf{a},\mathbf{b}\in\real^M$ whose coordinates are independently drawn from a standard normal distribution.}
    \label{tab:cos_sim}
\end{table}

\section{Training Details}\label{app:hyperparams}
\begin{table}[t]
    \centering
    \begin{tabular}{p{3cm} p{2cm} p{2cm} p{2cm} p{1cm} p{1.5cm} p{1.2cm}}
    \hline
    Model & Data set & Batch size & LR & $L2$ & Dropout & Epochs \\
    \hline
    Mistral-7B & Fineweb & $90$ & $3\times 10^{-5}$ & $0.01$ & $0.1$ & $2$ \\
    Mistral-7B & Cosmopedia & $90$ & $3\times 10^{-5}$ & $0.01$ & $0.1$ & $2$ \\
    Gemma-12B & Fineweb & $130$ & $3\times 10^{-5}$ & $0.01$ & $0.1$ & $2$ \\
    Qwen2-0.5B & Fineweb & $64$ & $1\times 10^{-5}$ & $0.1$ & $0$ & $1$ \\
    \hline
    \end{tabular}
    \caption{Hyperparameter selections for pretraining. LR refers to learning rate and $L2$ refers to $L2$ weight decay regularization.}
    \label{tab:hyperparams_pt}
\end{table}
\begin{table}[t]
    \centering
    \begin{tabular}{p{3cm} p{2cm} p{2cm} p{2cm} p{2cm} p{2cm} p{2cm}}
    \hline
    Data set & Batch size & Max length & LR  & Epochs \\
    \hline
    WinoGrande & 64 & 128 & $5\times 10^{-5}$ & 2 \\
    BoolQ & 32 & 256 & $5\times 10^{-5}$ & 2 \\
    ARC-Easy & 32 & 256 & $5\times 10^{-5}$ & 2 \\
    ARC-Challenge & 32 & 256 & $5\times 10^{-5}$ & 2 \\
    MetaMath & 50 & 700 & $5\times 10^{-5}$ & 1 \\
    MNLI & 120 & 256 & $5\times 10^{-5}$ & 2 \\
    SST2 & 120 & 256 & $5\times 10^{-5}$ & 2 \\
    \hline
    \end{tabular}
    \caption{Hyperparameter selections for fine-tuning. Max length refers to the maximum sequence length (in tokens) allowed per example. LR refers to learning rate. }
    \label{tab:hyperparams_ft}
\end{table}
We use LoRA rank $r=64$ to approximate Mistral-7B pretraining and rank $r=128$ to approximate Gemma3-12B pretraining. We run full pretraining on Qwen2-0.5B. All fine-tuning uses LoRA $r=8$. In each case, we follow standard practice and use LoRA $\alpha$ as $\alpha=2r$. For pretraining, we use a maximum token length per example as $2048$ tokens. While pretraining, we use gradient clipping with threshold $1.0$. We run pretraining on two NVIDIA H200 cards and fine-tuning on one NVIDIA H200 card.\footnote{Pretraining took approximately 2-3 hours per time step. Fine-tuning each patch took approximately 2 hours per patch for MetaMath and MNLI and less than 1 hour per patch for other benchmarks. Cumulatively, our experiments required $3 \text{ base models} \times 3\text{ repetitions} \times 10\text{ time steps}=90 \text{ continual pretraining updates}$ for Fineweb and $1 \text{ base models} \times 3\text{ repetitions} \times 8\text{ time steps}=24 \text{ continual pretraining updates}$ for Cosmopedia. } Hyperparameters for pretraining and fine-tuning are shown in Tables~\ref{tab:hyperparams_pt} and~\ref{tab:hyperparams_ft}, respectively. All training used the Axolotl~\citep{axolotl} framework. We run evaluations using the Language Model Evaluation Harness~\citep{eval-harness}.  
\begin{table}[t]\label{tab:fw-chunks}
    \centering
    \begin{tabular}{p{2cm} p{3.5cm} p{5cm} }
    \hline
         $t$ & Dump & Time period \\
         \hline
         $1$ & CC-MAIN-2023-50 & November/December 2023 \\
         $2$ & CC-MAIN-2024-10 & February/March 2024 \\
         $3$ & CC-MAIN-2024-18 & April 2024 \\
         $4$ & CC-MAIN-2024-26 & June 2024 \\
         $5$ & CC-MAIN-2024-33 & August 2024 \\
         $6$ & CC-MAIN-2024-42 & October 2024 \\
         $7$ & CC-MAIN-2024-51 & December 2024 \\
         $8$ & CC-MAIN-2025-08 & February 2025 \\
         $9$ & CC-MAIN-2025-18 & April 2025 \\
         $10$ & CC-MAIN-2025-26 & June 2025 \\
         \hline
    \end{tabular}
    \caption{Fineweb division into temporal chunks for continual pretraining time steps. }
    \label{tab:fineweb_chunks}
\end{table}

For fine-tuning, we select the best checkpoint based on a validation split. We used the publicly available train, validation, and test splits for ARC-Easy and ARC-Challenge. For WinoGrande, BoolQ, MNLI and SST2, we use 90\% of the training split for training, 10\% of the training split for validation, and the validation split for testing. The MetaMath data set (which we used to train patches for GSM8k evaluations) is very large. We used 2.5\% for validation and restricted training to $3000$ steps, which is about 39\% of the data set. 

For pretraining, we do not use a validation split and always select the last checkpoint. We report the temporal division of Fineweb pretraining data sets for each time step in Table~\ref{tab:fineweb_chunks}. We use the ``dump'' column for temporal information about data samples. For Cosmopedia, we used each data subset for a different pretraining time step in the following order: \texttt{auto\_math\_text}, \texttt{wikihow}, \texttt{openstax}, \texttt{stanford}, \texttt{stories}, \texttt{web\_samples\_v1}, \texttt{web\_samples\_v2}, \texttt{khanacademy}. We used 50M tokens from each split except khanacademy. For that split, we used all of the available 21M tokens.

A potential problem during continual pretraining is the stability gap~\citep{guo2024}, or the observation that base model performance may drop sharply at the beginning of continual pretraining and then slowly recover. To mitigate this problem, we follow two suggestions from~\cite{guo2024}, namely (1) train on a smaller corpus for multiple epochs and (2) select the smaller corpus as a high-quality subset of the available data. Specifically, we train for two epochs and filter data quality utilizing the ``language\_score'' column of the Fineweb data set. We select only training examples with a language score $\geq 0.9$. 

\section{Additional Experimental Results}\label{app:exp}
We present additional results on temporal portability. Overall, we find that, when base model performance and stepwise fine-tuning performance are approximately constant across time steps, PortLLM performance is also approximately constant. If base model performance or stepwise fine-tuning performance exhibit gradual degradation, PortLLM may mirror that degradation. These results highlight the importance of robust continual pretraining to maintain downstream performance. 
\subsection{Additional Benchmarks for Mistral Pretrained on Fineweb}\label{app:add_bm}
\begin{figure}
    \subfloat{\includegraphics[width=0.165\textwidth,clip,trim={5mm 35mm 55mm 9mm}]{figures/experiments/legend.pdf}
    \includegraphics[width=0.22\textwidth,clip,trim={5mm 19mm 25mm 25mm}]{figures/experiments/legend.pdf}
    \includegraphics[width=0.25\textwidth,clip,trim={4mm 4mm 9mm 40mm}]{figures/experiments/legend.pdf}\hfill}
    \centering
    \addtocounter{subfigure}{-1}\\
    \subfloat[]{
    \includegraphics[width=0.3\textwidth]{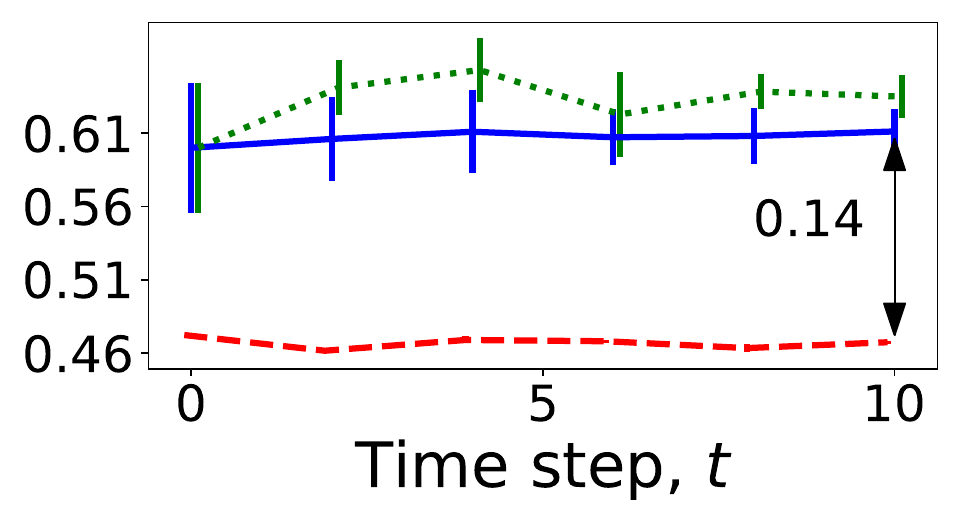}
    }
    \subfloat[]{
    \includegraphics[width=0.3\textwidth]{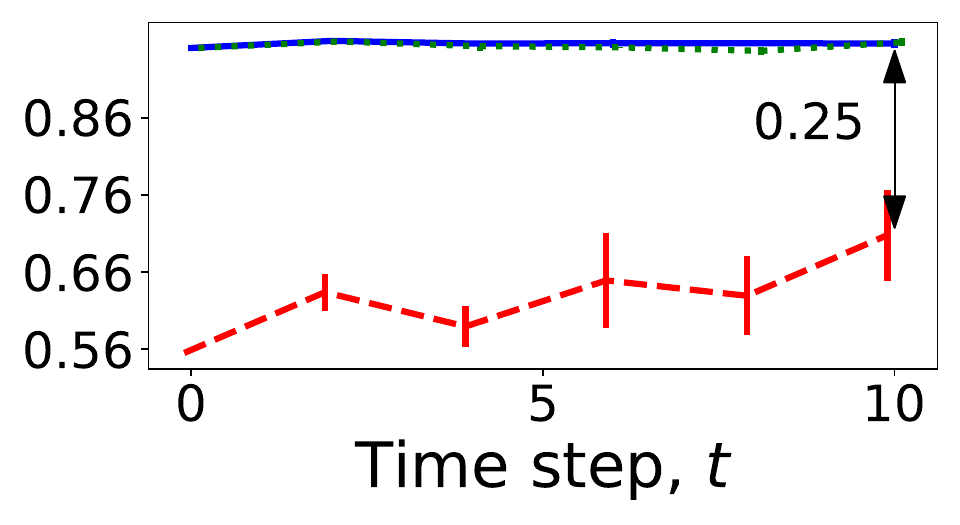}
    }
    \subfloat[]{
    \includegraphics[width=0.3\textwidth]{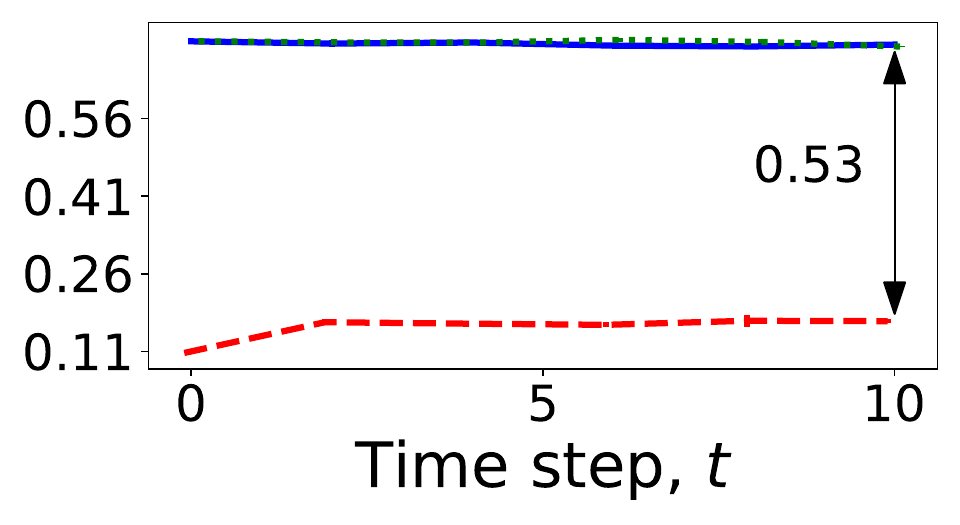}
    }

    \caption{Temporal portability results for Mistral-7B-v0.1 pretrained on Fineweb~\citep{penedo2024} and downstream fine-tuned on additional benchmarks. Testing results are shown for (a) MNLI (b) SST2 and (c) GSM8k. 
    For each benchmark, reported results are a sample average across $3$ repetitions. The error bars shown $95\%$ confidence interval. 
    }
    \label{fig:add_bm_mistral_fw}
\end{figure}

Figure~\ref{fig:add_bm_mistral_fw} shows temporal portability results for additional benchmarks after pretraining Mistral-7B on Fineweb data. Specifically, we test GSM8k~\citep{cobbe2021} for mathematical reasoning, MNLI~\citep{williams2018} for recognizing textual entailment, and SST2~\citep{socher2013} for sentiment analysis. Patches for GSM8k are trained using MetaMath~\citep{yu2023} data, while MNLI and SST2 use the publicly available training splits. These results largely mirror those presented in Section~\ref{exp-results-main-text}. Namely, PortLLM performance is comparable to stepwise fine-tuning (RQ1), better than no patching (RQ2), and roughly constant across time steps (RQ3). 

\subsection{Gemma3-12B Pretrained on Fineweb}\label{app:gemma-exp}
\begin{figure}
    \subfloat{\includegraphics[width=0.165\textwidth,clip,trim={5mm 35mm 55mm 9mm}]{figures/experiments/legend.pdf}
    \includegraphics[width=0.22\textwidth,clip,trim={5mm 19mm 25mm 25mm}]{figures/experiments/legend.pdf}
    \includegraphics[width=0.25\textwidth,clip,trim={4mm 4mm 9mm 40mm}]{figures/experiments/legend.pdf}\hfill}
    \centering
    \addtocounter{subfigure}{-1}\\
    \subfloat[]{\includegraphics[width=0.25\textwidth]{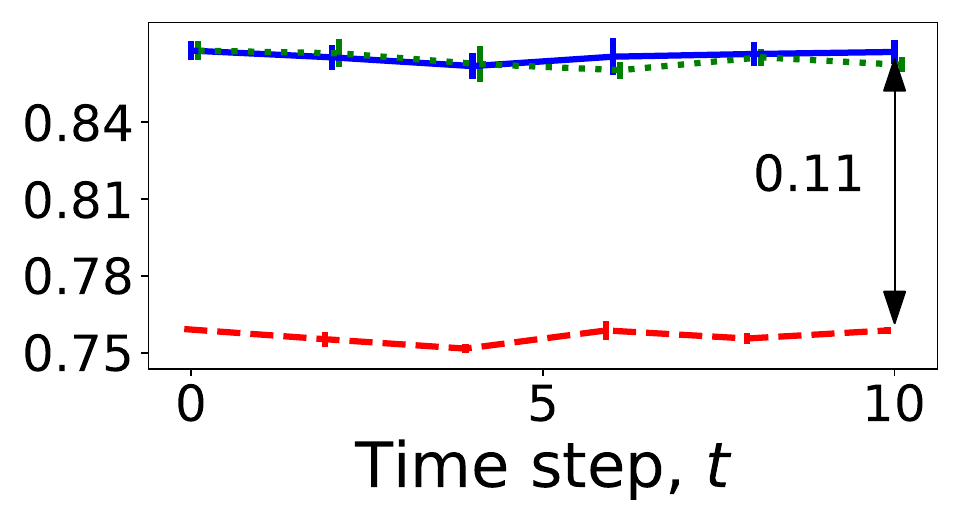}}\subfloat[]{\includegraphics[width=0.25\textwidth]{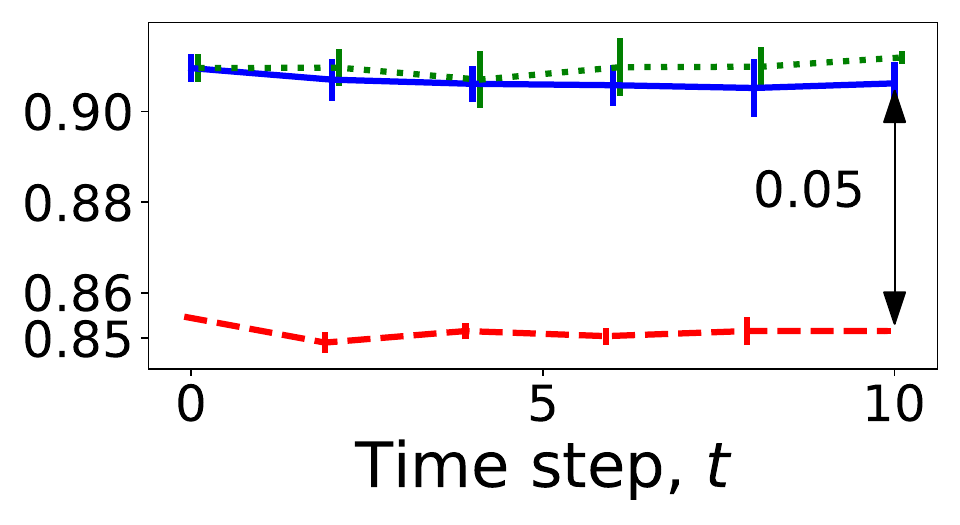}}
    \subfloat[]{\includegraphics[width=0.25\textwidth]{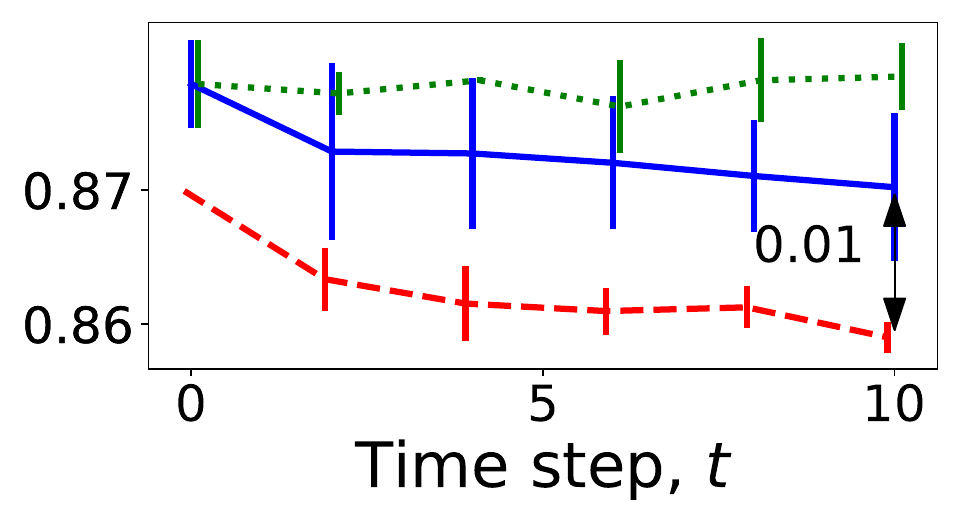}}\subfloat[]{\includegraphics[width=0.25\textwidth]{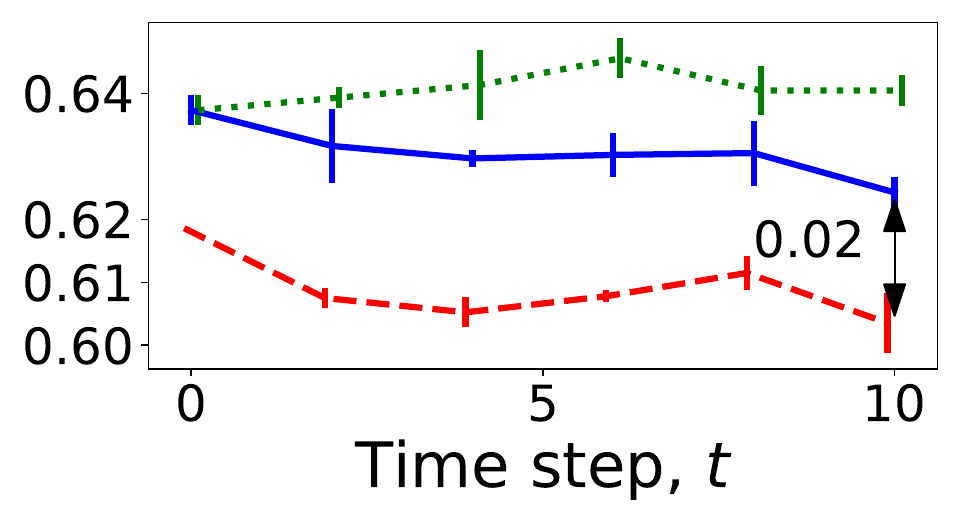}}
    \caption{Temporal portability results for continually pretraining Gemma3-12B on Fineweb data and downstream fine-tuning on (a) WinoGrande, (b) BoolQ, (c) ARC-Easy, and (d) ARC-Challenge. Average performance across $3$ repetitions for PortLLM patching, stepwise fine-tuning, and no patching performance are shown. The error bars indicate the 95\% confidence interval. 
    }
    \label{fig:gemma_exp}
\end{figure}
Temporal portability results for pretraining Gemma3-12B on Fineweb data are shown in Figure~\ref{fig:gemma_exp}. Temporal trends on WinoGrande and BoolQ are similar to those of Mistral-7B in Section~\ref{exp-results-main-text}. On ARC-Easy and ARC-Challenge, PortLLM performance degrades mildly across time steps but continues to outperform no patching performance. This mild degradation mirrors a degradation in base model performance, indicating that PortLLM performance may be affected by changes in base model performance. 

\subsection{Qwen2-0.5B Pretrained on Fineweb}\label{app:qwen-exp}
\begin{figure}
    \subfloat{\includegraphics[width=0.165\textwidth,clip,trim={5mm 35mm 55mm 9mm}]{figures/experiments/legend.pdf}
    \includegraphics[width=0.22\textwidth,clip,trim={5mm 19mm 25mm 25mm}]{figures/experiments/legend.pdf}
    \includegraphics[width=0.25\textwidth,clip,trim={4mm 4mm 9mm 40mm}]{figures/experiments/legend.pdf}\hfill}
    \centering
    \addtocounter{subfigure}{-1}\\
    \subfloat[]{\includegraphics[width=0.25\textwidth]{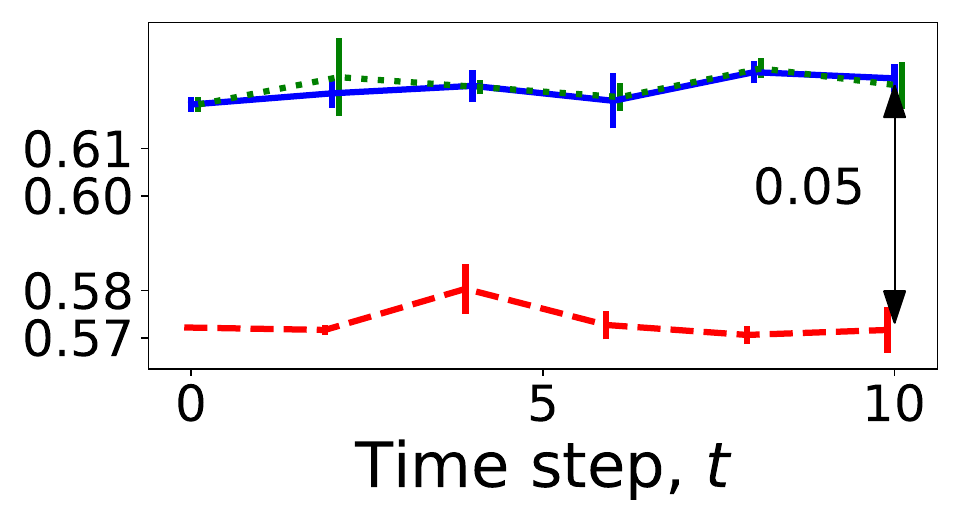}}\subfloat[]{\includegraphics[width=0.25\textwidth]{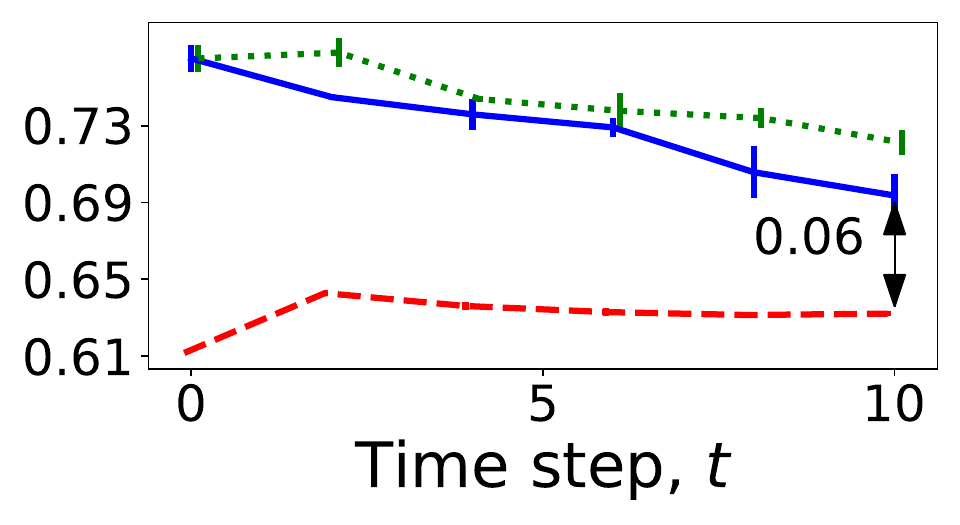}}
    \subfloat[]{\includegraphics[width=0.25\textwidth]{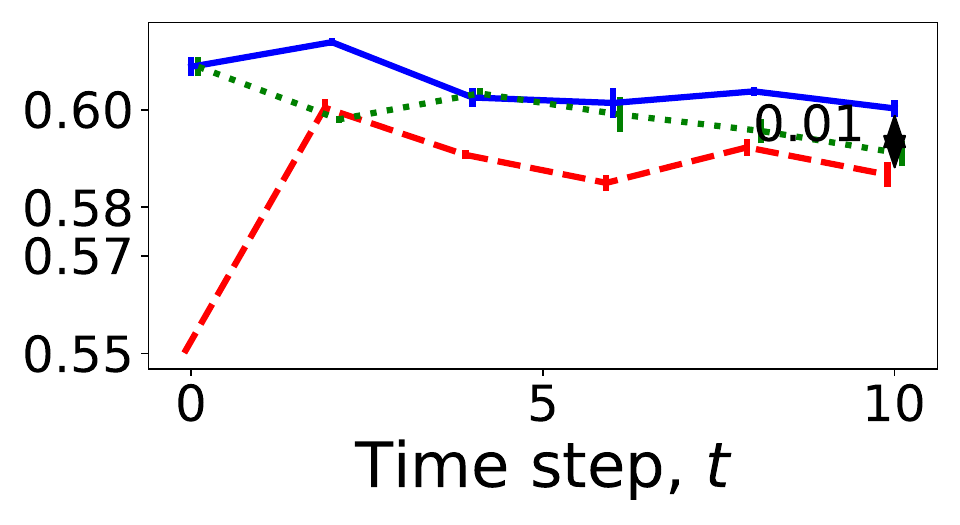}}\subfloat[]{\includegraphics[width=0.25\textwidth]{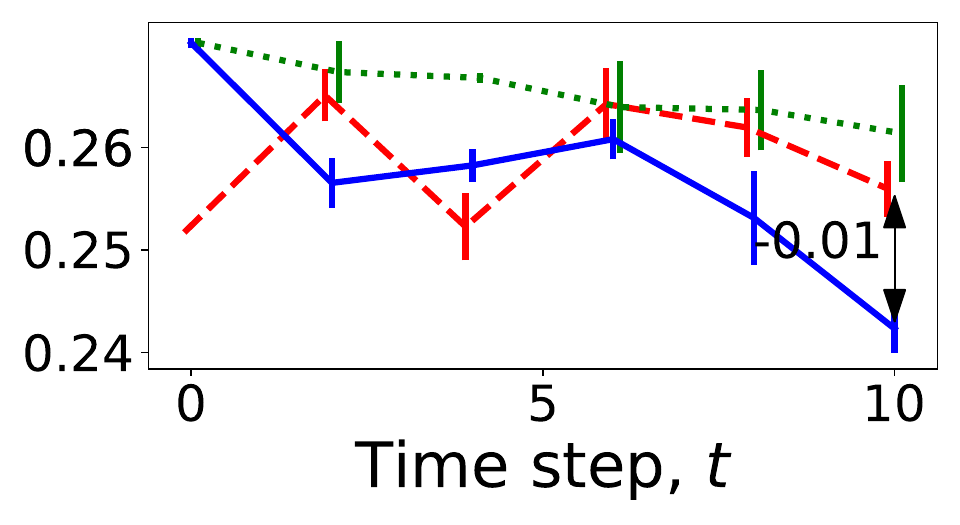}}
  
    \caption{Temporal portability results for continually pretraining Qwen2-0.5B on Fineweb data and downstream fine-tuning on (a) WinoGrande, (b) BoolQ, (c) ARC-Easy, and (d) ARC-Challenge. Average performance across $3$ repetitions for PortLLM patching, stepwise fine-tuning, and no patching performance are shown. The error bars indicate the 95\% confidence interval. 
    }
    \label{fig:qwen_exp}
\end{figure}
Results for pretraining Qwen2-0.5B on Fineweb are shown in Figure~\ref{fig:qwen_exp}. The near random guessing performance on ARC-Challenge indicates that $0.5$B scale models are not well-equipped for the ARC-Challenge benchmark. The other three benchmarks are more informative. PortLLM is comparable to stepwise fine-tuning on WinoGrande, slightly underperforms stepwise fine-tuning on BoolQ, and slightly outperforms stepwise fine-tuning on ARC-Easy. PortLLM outperforms the base model for all three benchmarks. PortLLM and stepwise fine-tuning exhibit similar trends across time steps: performance remains approximately constant on WinoGrande, and degrades slightly on BoolQ and ARC-Easy. This indicates that, if continual pretraining moves the model to a place where stepwise fine-tuning is less effective, PortLLM performance may also be affected.

\subsection{Mistral-7B Pretrained on Cosmopedia}\label{app:cos-exp}
\begin{figure}
    \subfloat{\includegraphics[width=0.165\textwidth,clip,trim={5mm 35mm 55mm 9mm}]{figures/experiments/legend.pdf}
    \includegraphics[width=0.22\textwidth,clip,trim={5mm 19mm 25mm 25mm}]{figures/experiments/legend.pdf}
    \includegraphics[width=0.25\textwidth,clip,trim={4mm 4mm 9mm 40mm}]{figures/experiments/legend.pdf}\hfill}
    \centering
    \addtocounter{subfigure}{-1}\\
    \subfloat[]{\includegraphics[width=0.25\textwidth]{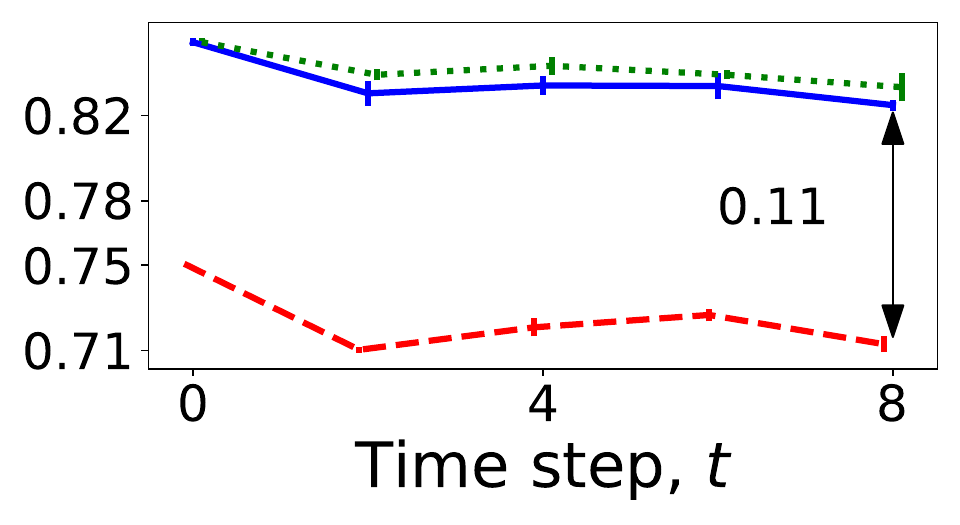}}\subfloat[]{\includegraphics[width=0.25\textwidth]{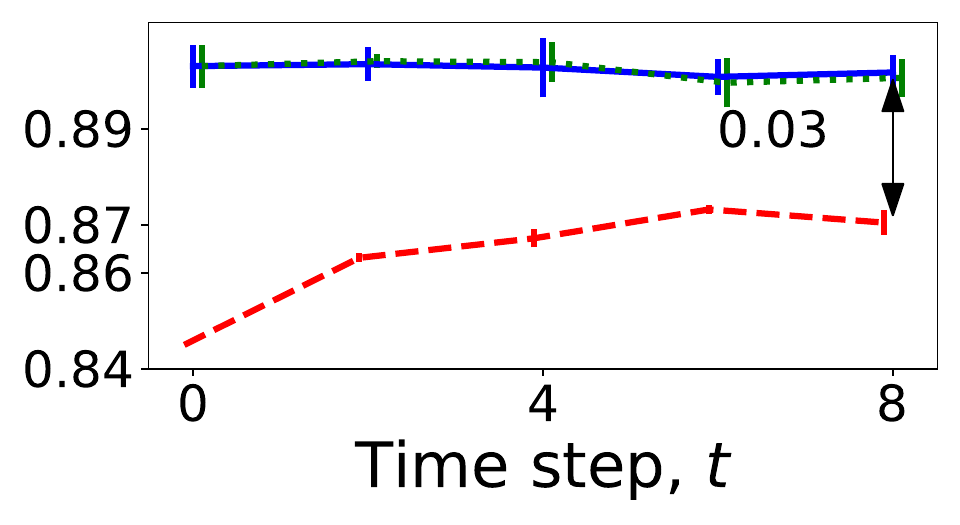}}\subfloat[]{\includegraphics[width=0.25\textwidth]{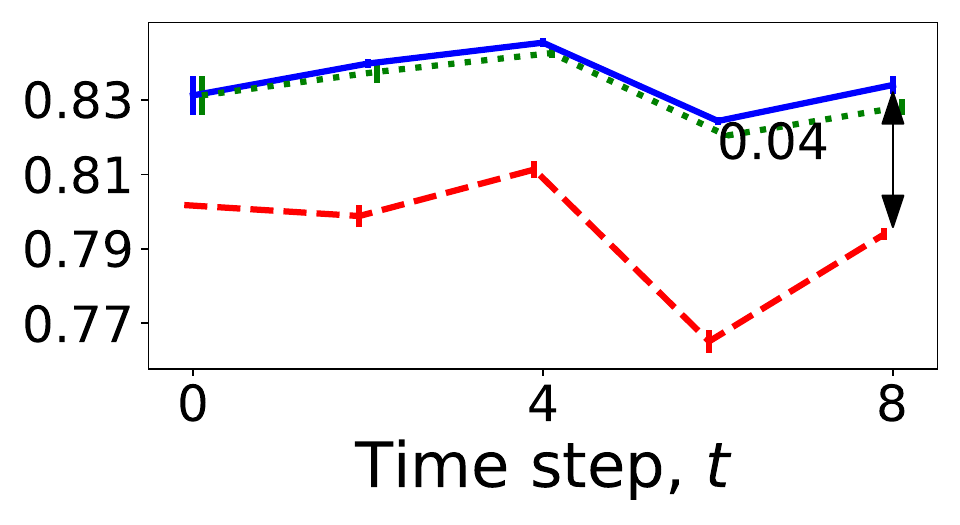}}\subfloat[]{\includegraphics[width=0.25\textwidth]{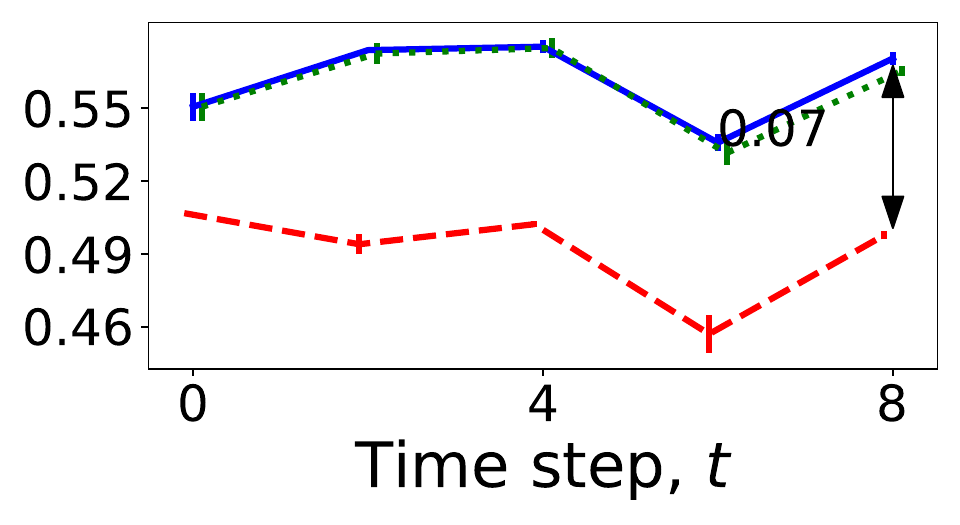}}
    \caption{Temporal portability results for continually pretraining Mistral-7B-v0.1 on Cosmopedia data and downstream fine-tuning on (a) WinoGrande, (b) BoolQ, (c) ARC-Easy, and (d) ARC-Challenge. Average performance across $3$ repetitions is plotted. The error bars indicate the 95\% confidence interval. 
    }
    \label{fig:cos_mistral_exp}
\end{figure}

Temporal portability results for continually pretraining Mistral-7B on Cosmopedia data are shown in Figure~\ref{fig:cos_mistral_exp}. On WinoGrande and BoolQ, PortLLM and stepwise fine-tuning are comparable to each other and maintain approximately constant performance across time steps. On ARC-Easy and ARC-Challenge, they exhibit non-monotonic variation in performance across time steps. This variation mirrors variation in base model performance. We hypothesize that this occurs because of the heterogeneity of Cosmopedia data across time steps. This indicates that, when continual pretraining changes base model performance, that variation may be reflected in PortLLM performance. 

\section{Empirical Results for 1-D Slice of Loss Landscape}\label{app:1d-slice-exp}
\begin{figure}
    \centering
    \subfloat[]{\includegraphics[width=0.3\textwidth]{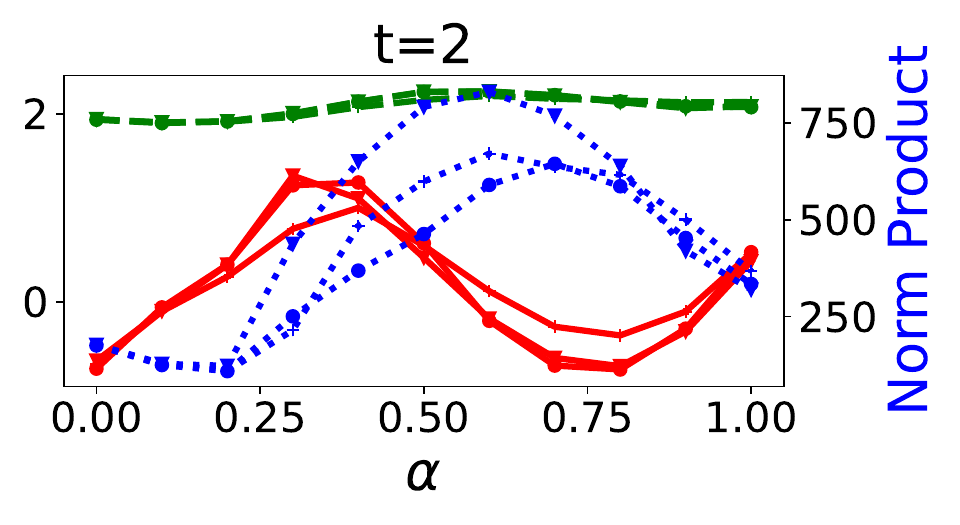}}
    \subfloat[]{\includegraphics[width=0.3\textwidth]{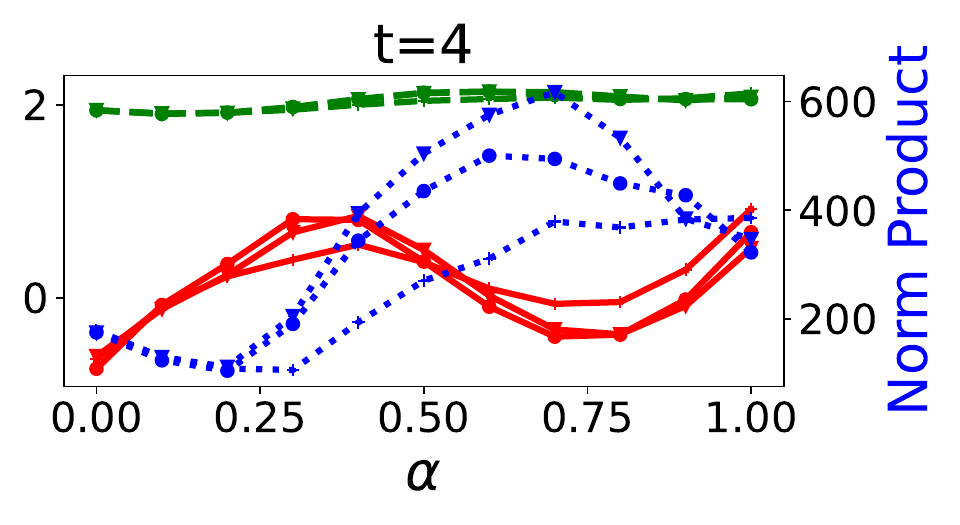}}
    \subfloat{\vspace{15pt}\includegraphics[valign=t,width=0.2\textwidth]{figures/data_dump/cost_legend.pdf}} 
    \addtocounter{subfigure}{-1}\\
    \subfloat[]{\includegraphics[width=0.3\textwidth]{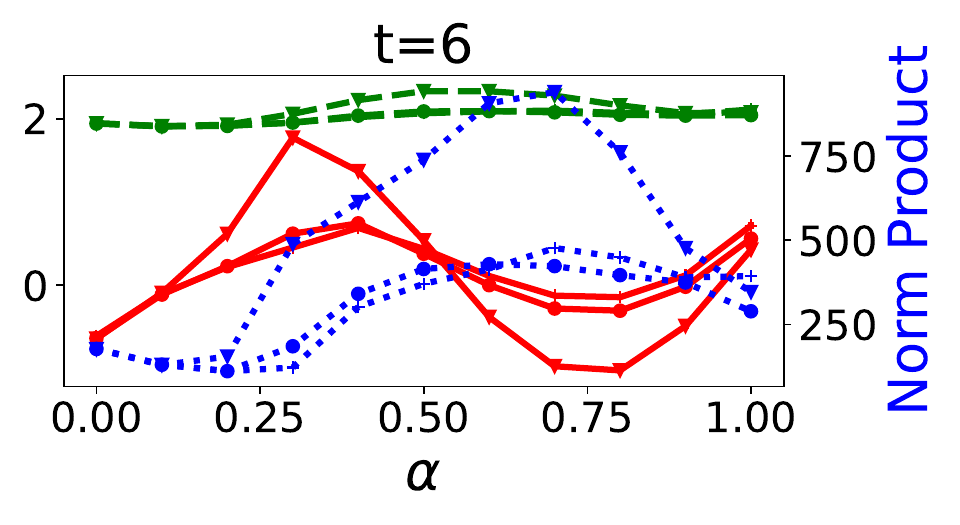}}
    \subfloat[]{\includegraphics[width=0.3\textwidth]{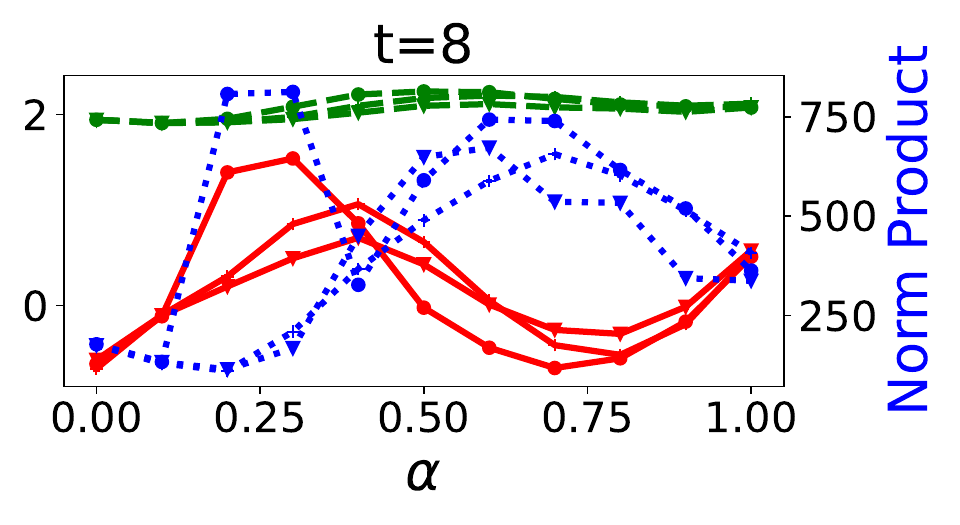}}
    \subfloat[]{\includegraphics[width=0.3\textwidth]{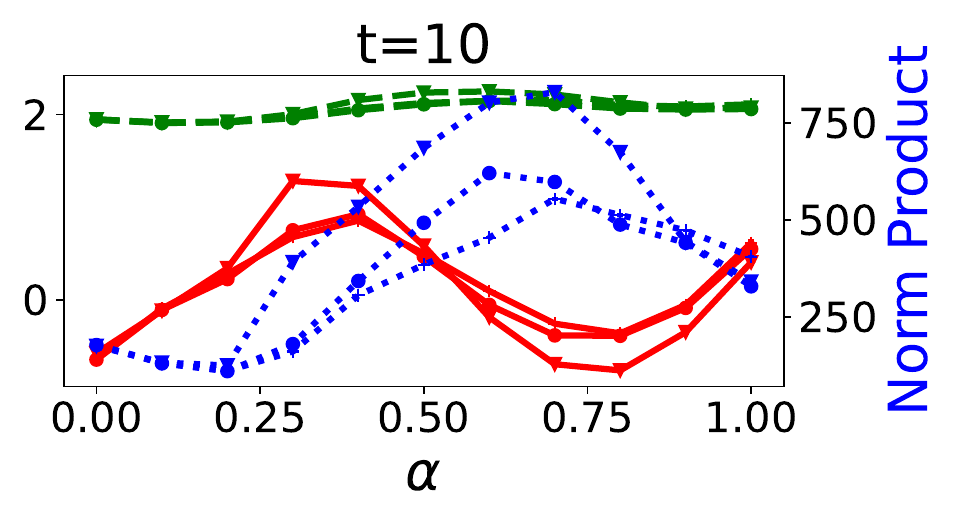}}
    \caption[]{WinoGrande empirical results for RQ1. The plots include the loss for downstream testing data $\gc{\alpha}$ on a 1-D slice between the PortLLM patch and the stepwise fine-tuning patch (dashed green line), the derivative of the 1-D slice (solid red line), and the norm product  $\|\nabla_{\unexpp{ }}\uloss{\base{t}}(\uls{\alpha})\|\|\ftdelta{t}\|$ (dotted blue line). We show $t\in\{2,4,6,8,10\}$ and three repetitions on each plot.
    }
    \label{fig:wg_cost}
\end{figure}

\begin{figure}
    \centering
    \subfloat[]{\includegraphics[width=0.3\textwidth]{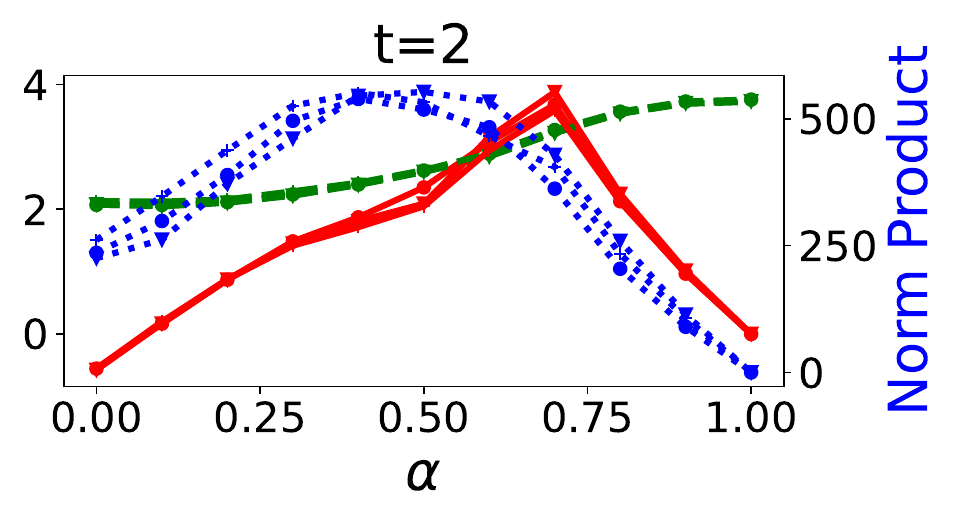}}
    \subfloat[]{\includegraphics[width=0.3\textwidth]{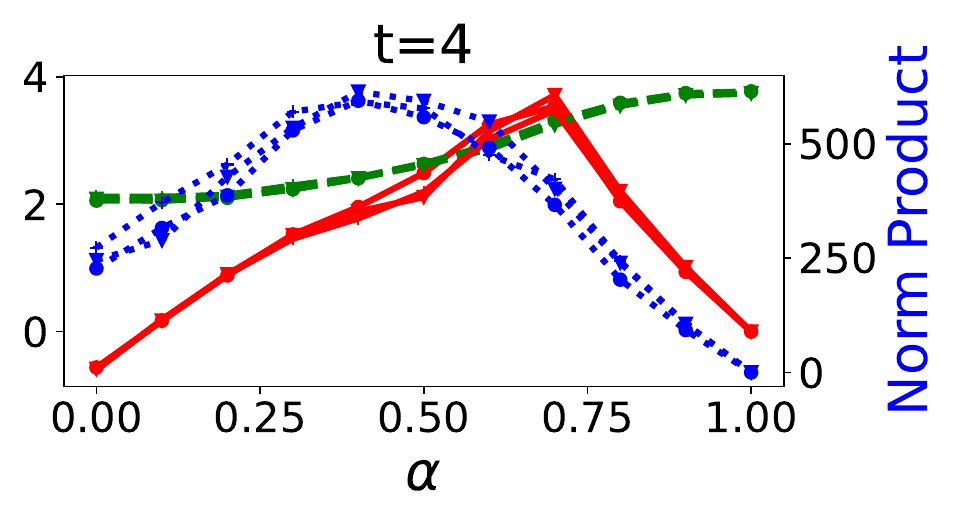}}
    \subfloat{\vspace{15pt}\includegraphics[valign=t,width=0.2\textwidth]{figures/data_dump/benefit_legend.pdf}} 
    \addtocounter{subfigure}{-1}\\
    \subfloat[]{\includegraphics[width=0.3\textwidth]{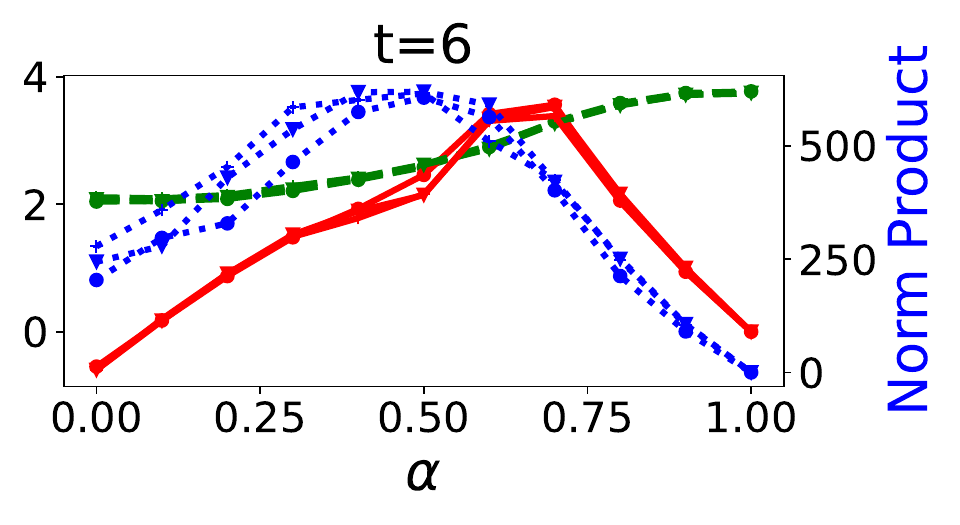}}
    \subfloat[]{\includegraphics[width=0.3\textwidth]{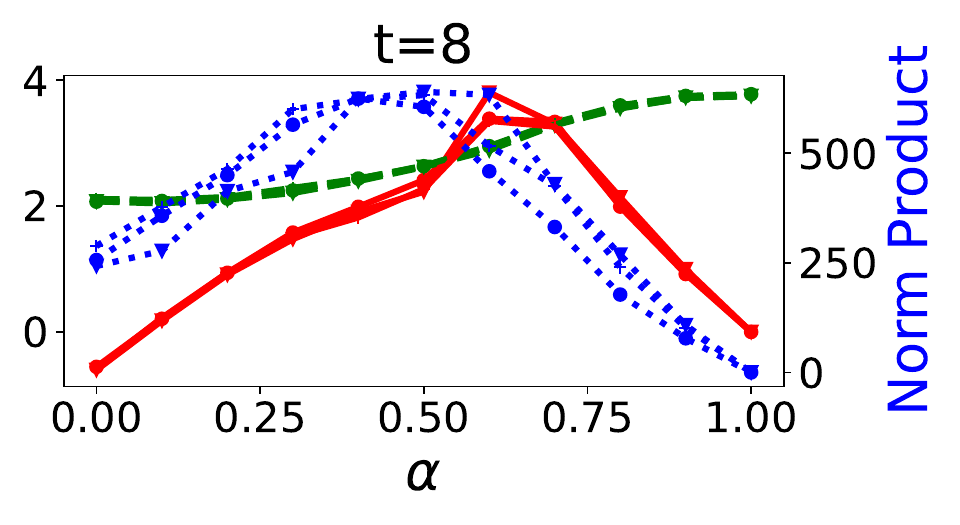}}
    \subfloat[]{\includegraphics[width=0.3\textwidth]{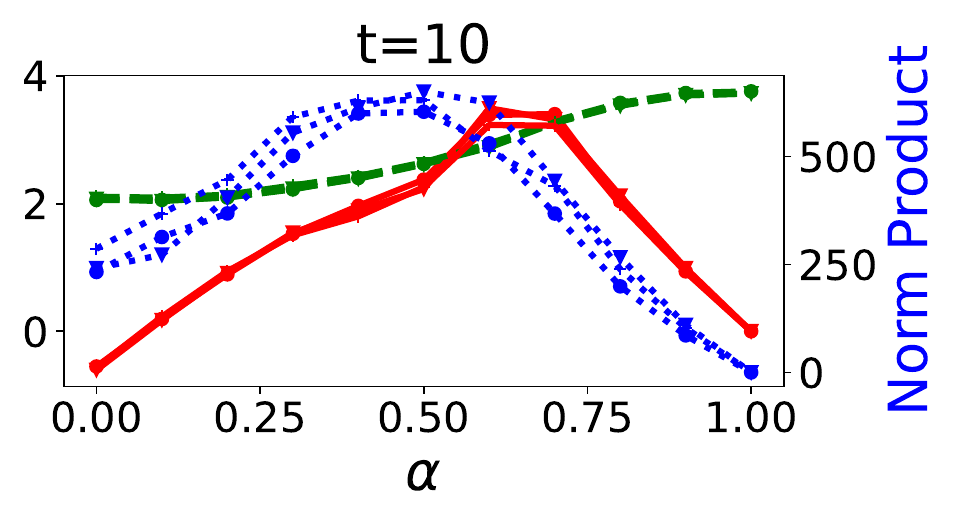}}
    \caption[]{WinoGrande empirical results for RQ2. The plots include the loss for downstream testing data $\gb{\alpha}$ on a 1-D slice between the PortLLM patch and the stepwise fine-tuning patch (dashed green line), the derivative of the 1-D slice (solid red line), and the norm product  $\|\nabla_{\unexpp{ }}\uloss{\base{t}}((1-\alpha)\unexpp{0})\|\|\unexpp{0}\|$ (dotted blue line). We show $t\in\{2,4,6,8,10\}$ and three repetitions on each plot.
    }
    \label{fig:wg_benefit}
\end{figure}

\begin{figure}
    \centering
    \subfloat[]{\includegraphics[width=0.3\textwidth]{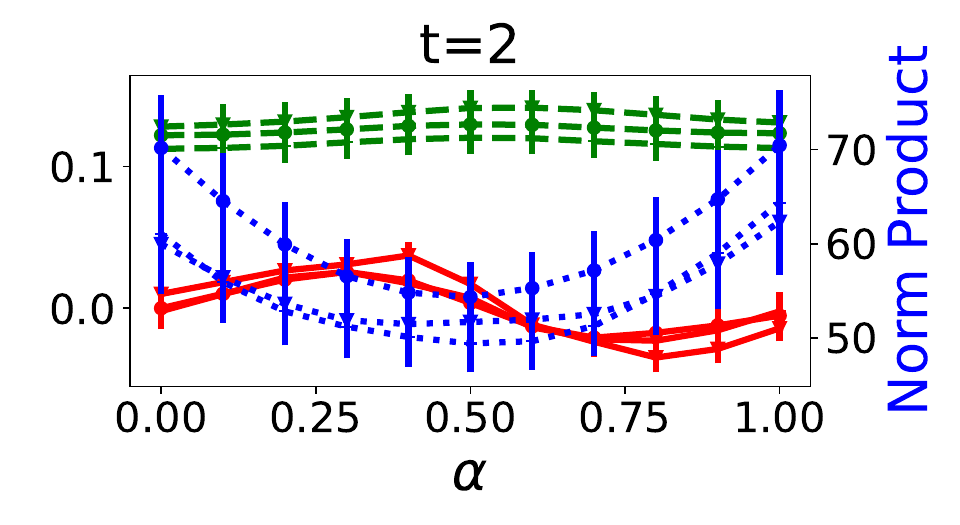}}
    \subfloat[]{\includegraphics[width=0.3\textwidth]{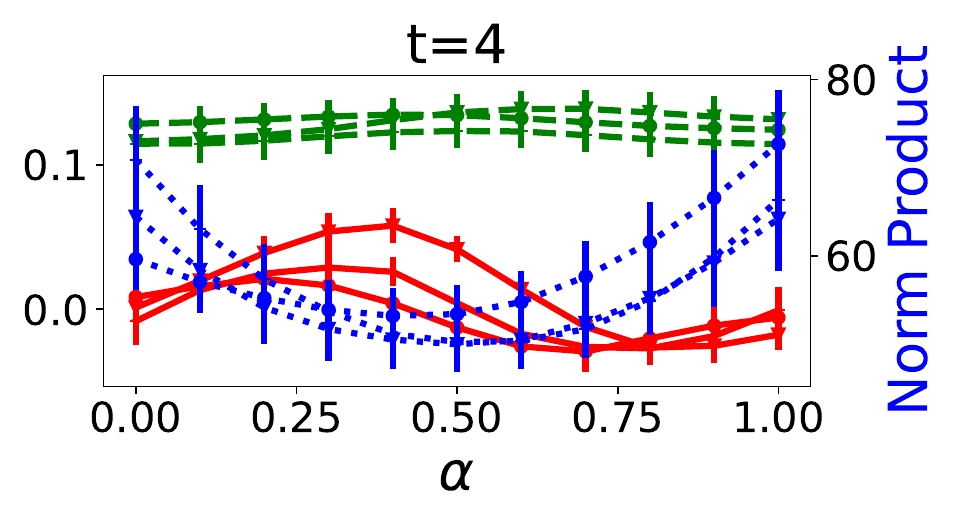}}
    \subfloat{\vspace{15pt}\includegraphics[valign=t,width=0.2\textwidth]{figures/data_dump/cost_legend.pdf}} 
    \addtocounter{subfigure}{-1}\\
    \subfloat[]{\includegraphics[width=0.3\textwidth]{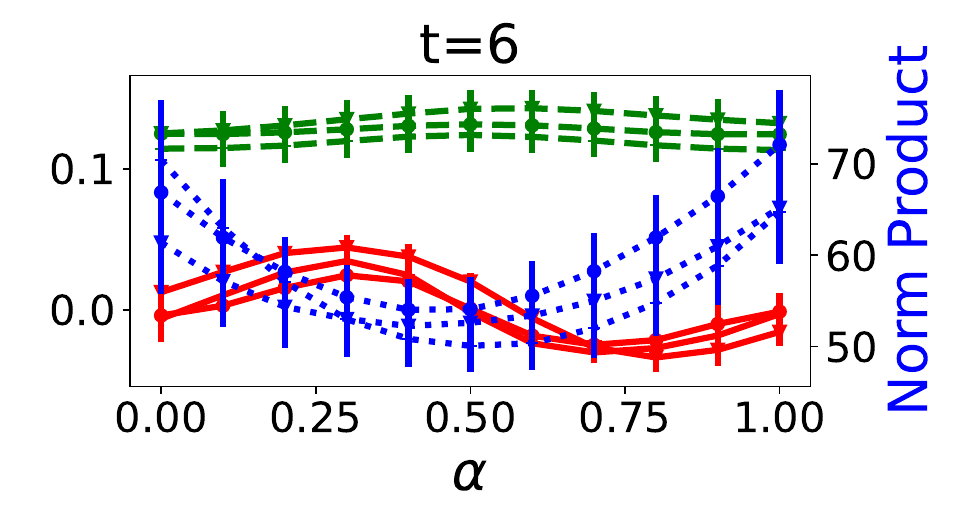}}
    \subfloat[]{\includegraphics[width=0.3\textwidth]{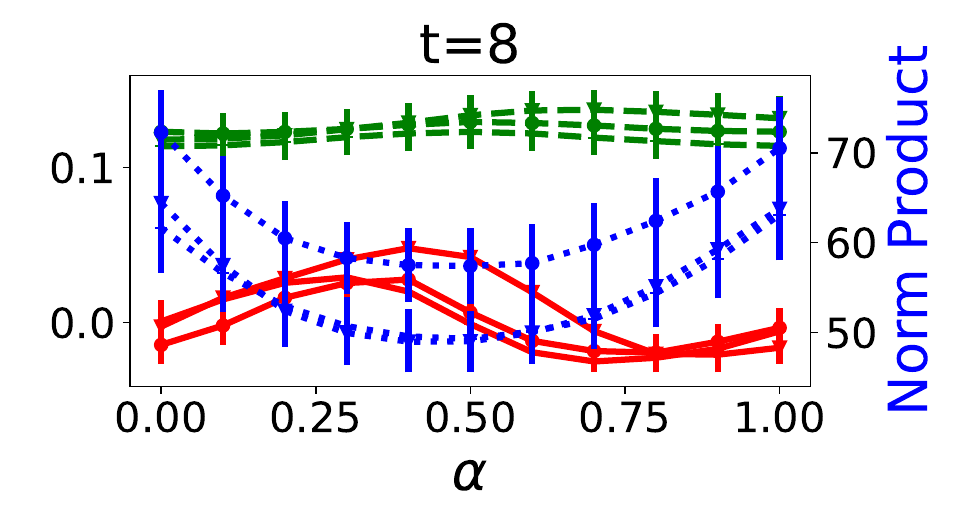}}
    \subfloat[]{\includegraphics[width=0.3\textwidth]{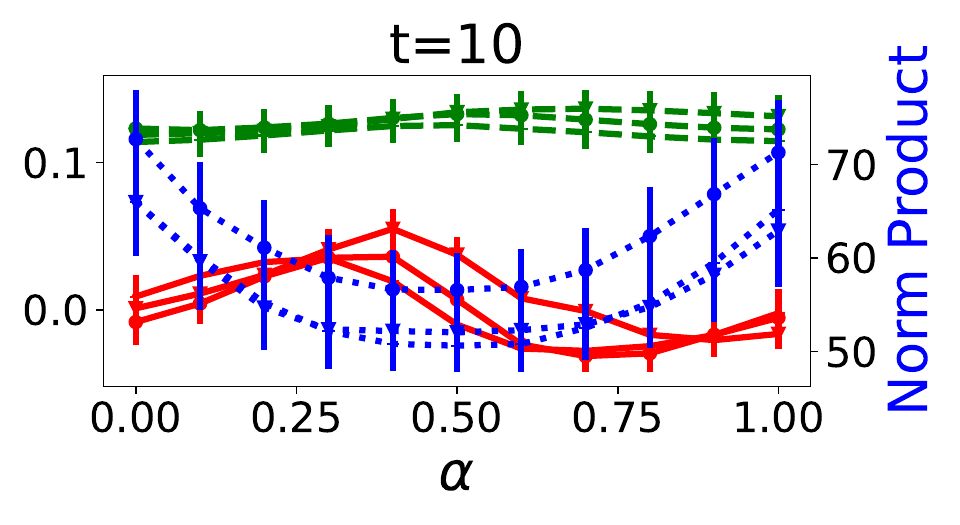}}
    \caption[]{BoolQ empirical results for RQ1. The plots include the loss for downstream testing data $\gc{\alpha}$ on a 1-D slice between the PortLLM patch and the stepwise fine-tuning patch (dashed green line), the derivative of the 1-D slice (solid red line), and the norm product  $\|\nabla_{\unexpp{ }}\uloss{\base{t}}(\uls{\alpha})\|\|\ftdelta{t}\|$ (dotted blue line). We show $t\in\{2,4,6,8,10\}$ and three repetitions on each plot.
    }
    \label{fig:bq_cost}
\end{figure}

\begin{figure}
    \centering
    \subfloat[]{\includegraphics[width=0.3\textwidth]{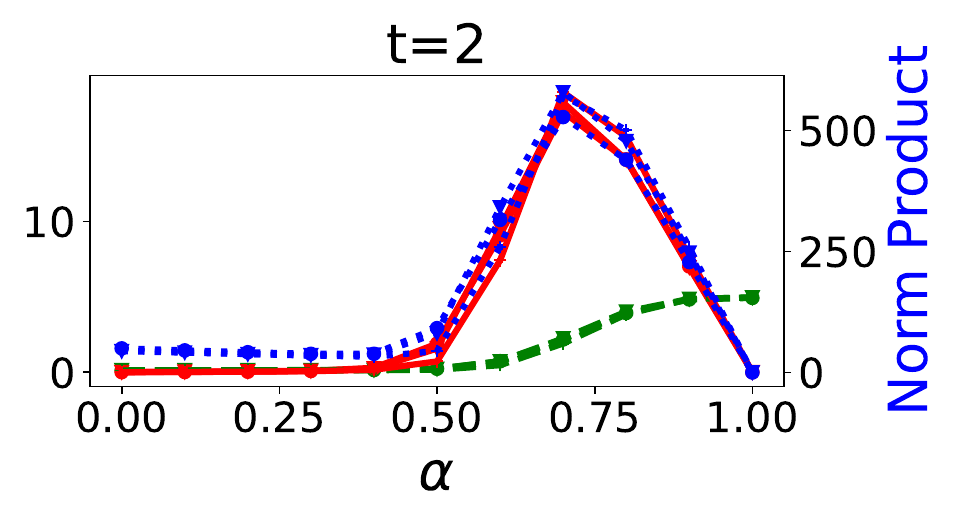}}
    \subfloat[]{\includegraphics[width=0.3\textwidth]{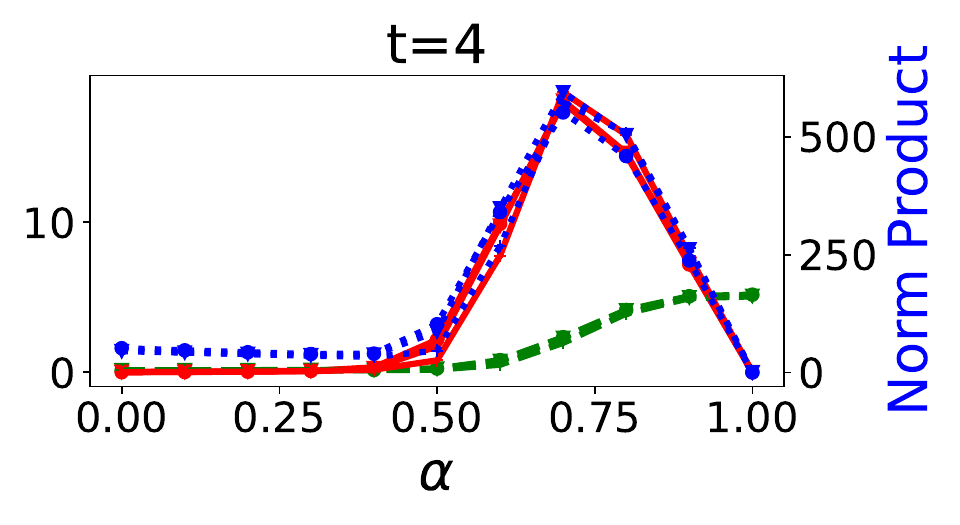}}
    \subfloat{\vspace{15pt}\includegraphics[valign=t,width=0.2\textwidth]{figures/data_dump/benefit_legend.pdf}}
    \addtocounter{subfigure}{-1}\\
    \subfloat[]{\includegraphics[width=0.3\textwidth]{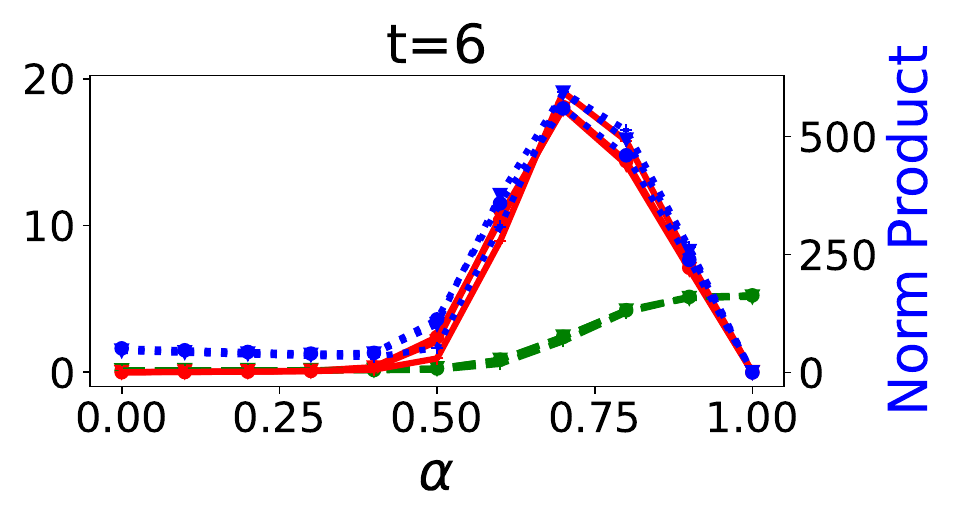}}
    \subfloat[]{\includegraphics[width=0.3\textwidth]{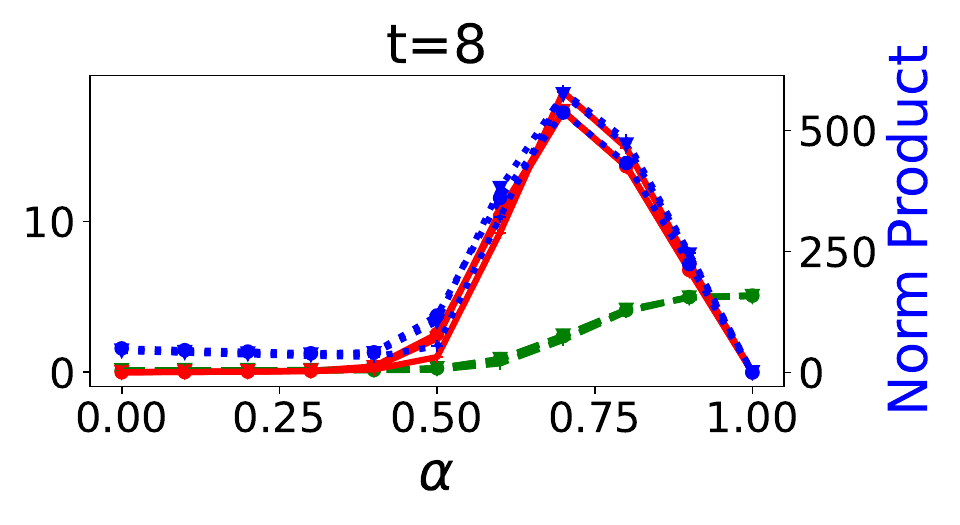}}
    \subfloat[]{\includegraphics[width=0.3\textwidth]{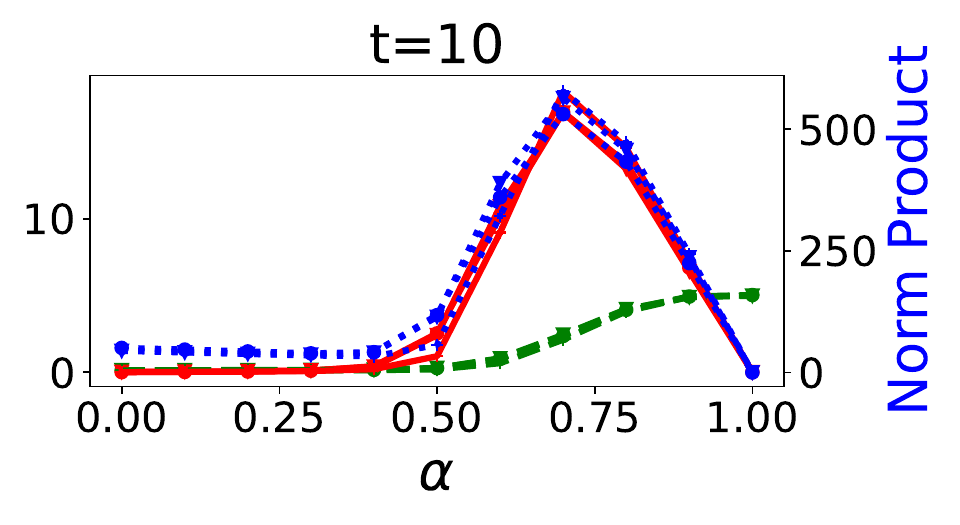}}
    \caption[]{BoolQ empirical results for RQ2. The plots include the loss for downstream testing data $\gb{\alpha}$ on a 1-D slice between the PortLLM patch and the stepwise fine-tuning patch (dashed green line), the derivative of the 1-D slice (solid red line), and the norm product  $\|\nabla_{\unexpp{ }}\uloss{\base{t}}((1-\alpha)\unexpp{0})\|\|\unexpp{0}\|$ (dotted blue line). We show $t\in\{2,4,6,8,10\}$ and three repetitions on each plot.
    }
    \label{fig:bq_benefit}
\end{figure}

\begin{figure}
    \centering
    \subfloat[]{\includegraphics[width=0.3\textwidth]{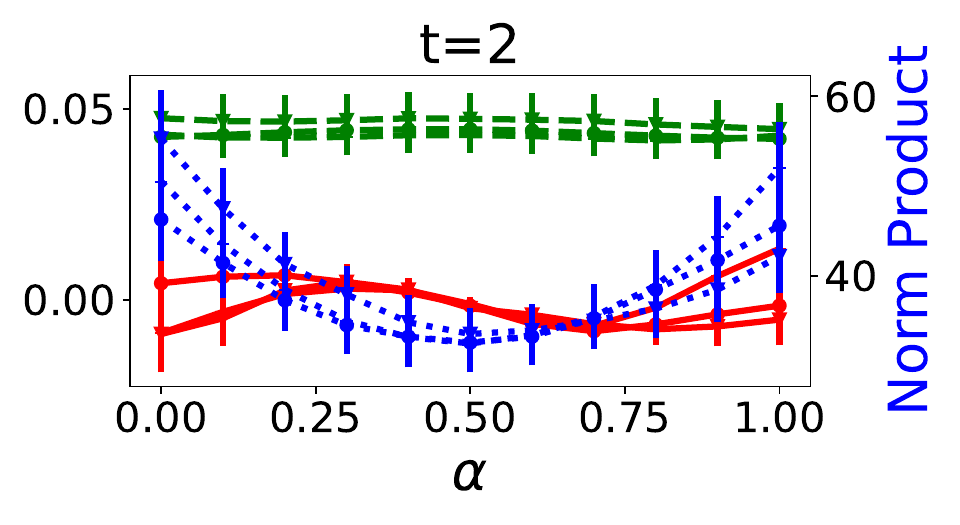}}
    \subfloat[]{\includegraphics[width=0.3\textwidth]{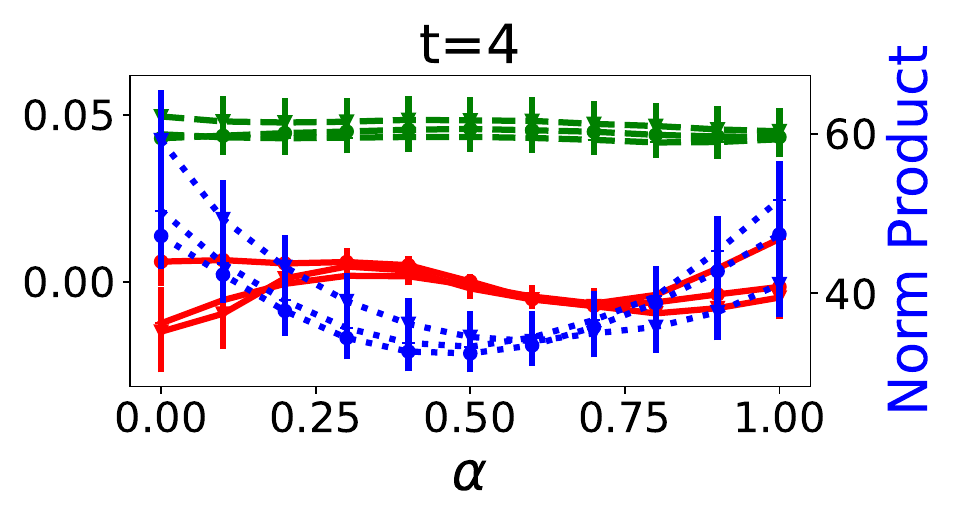}}
    \subfloat{\vspace{15pt}\includegraphics[valign=t,width=0.2\textwidth]{figures/data_dump/cost_legend.pdf}} 
    \addtocounter{subfigure}{-1}\\
    \subfloat[]{\includegraphics[width=0.3\textwidth]{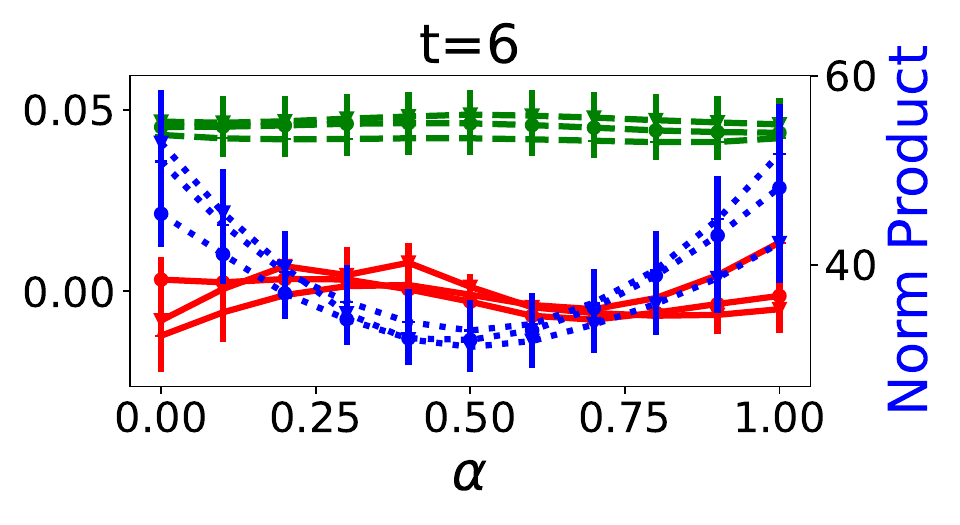}}
    \subfloat[]{\includegraphics[width=0.3\textwidth]{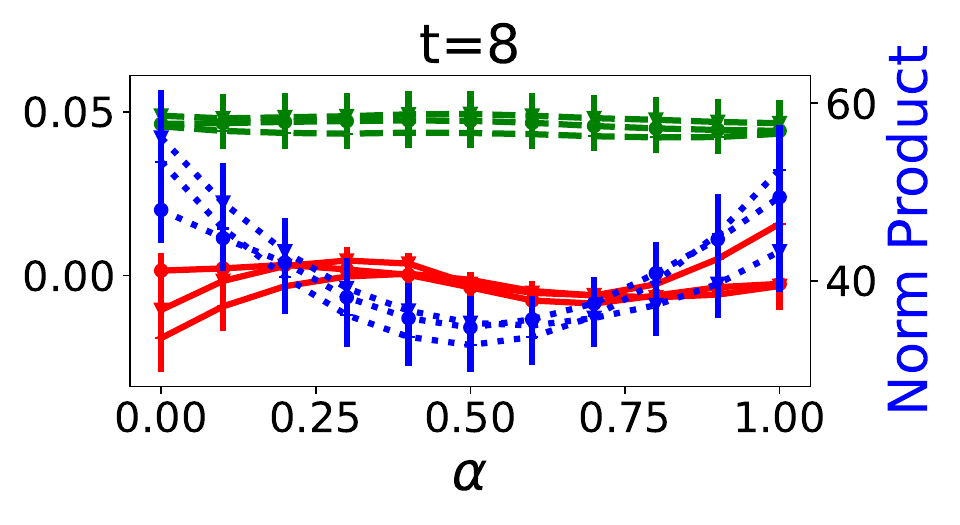}}
    \subfloat[]{\includegraphics[width=0.3\textwidth]{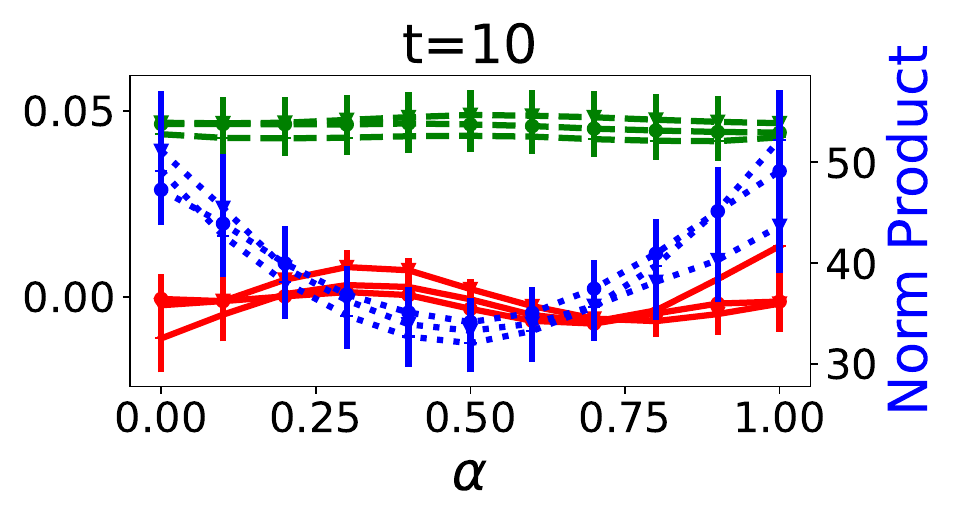}}
    \caption[]{ARC-Easy empirical results for RQ1. The plots include the loss for downstream testing data $\gc{\alpha}$ on a 1-D slice between the PortLLM patch and the stepwise fine-tuning patch (dashed green line), the derivative of the 1-D slice (solid red line), and the norm product  $\|\nabla_{\unexpp{ }}\uloss{\base{t}}(\uls{\alpha})\|\|\ftdelta{t}\|$ (dotted blue line). We show $t\in\{2,4,6,8,10\}$ and three repetitions on each plot.
    }
    \label{fig:ae_cost}
\end{figure}

\begin{figure}
    \centering
    \subfloat[]{\includegraphics[width=0.3\textwidth]{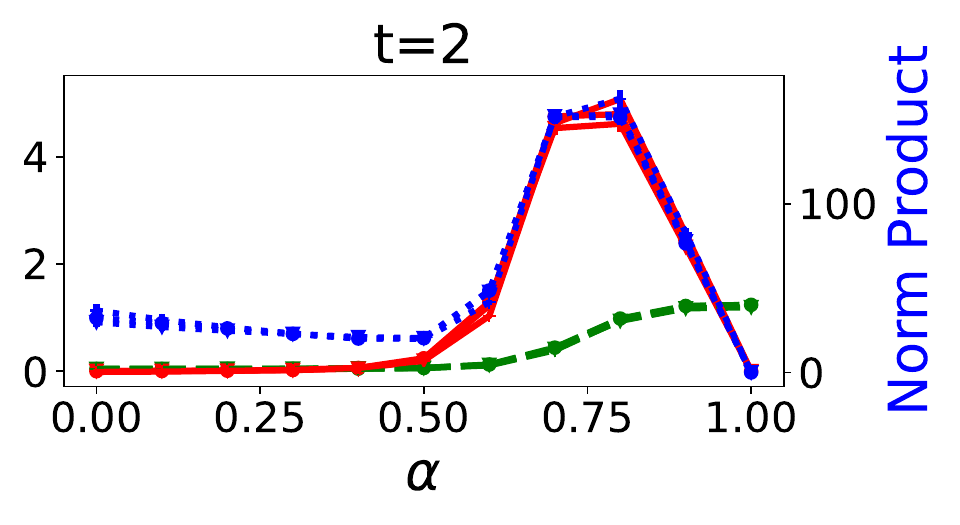}}
    \subfloat[]{\includegraphics[width=0.3\textwidth]{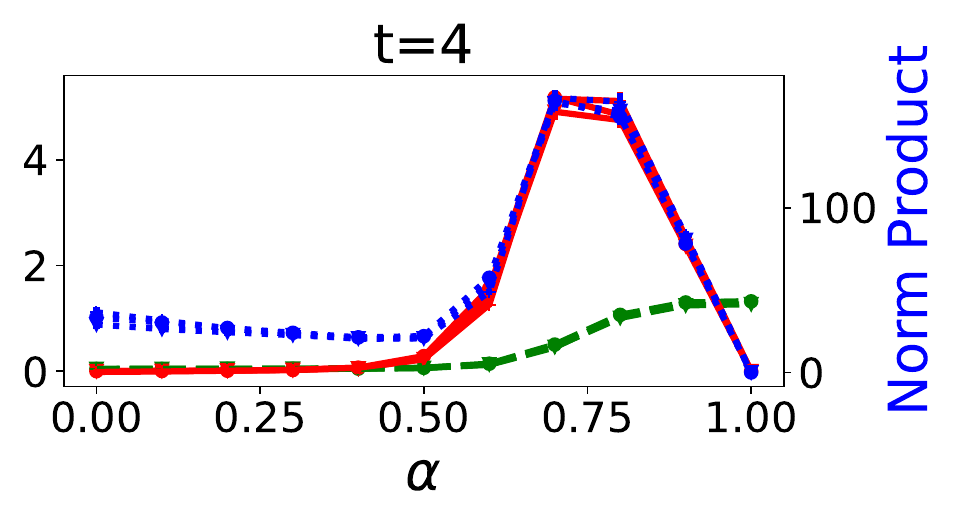}}
    \subfloat{\vspace{15pt}\includegraphics[valign=t,width=0.2\textwidth]{figures/data_dump/benefit_legend.pdf}} 
    \addtocounter{subfigure}{-1}\\
    \subfloat[]{\includegraphics[width=0.3\textwidth]{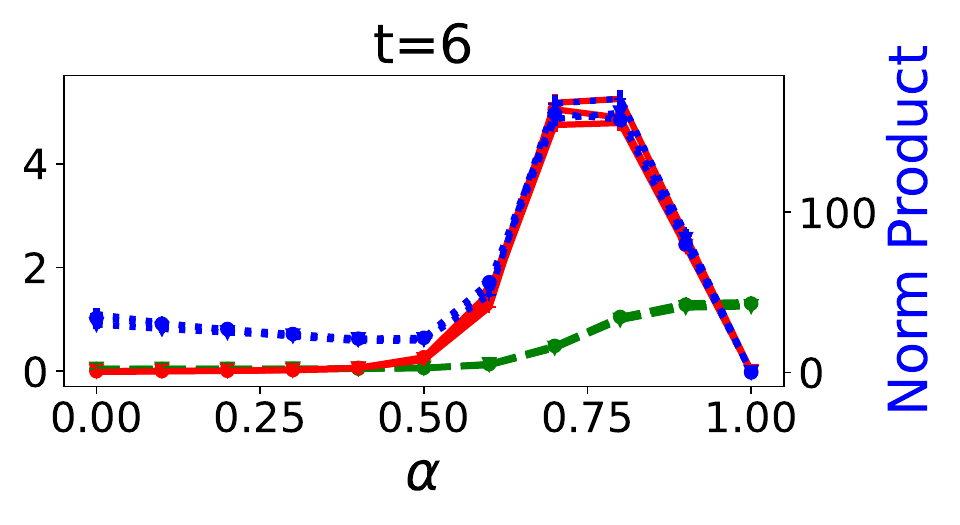}}
    \subfloat[]{\includegraphics[width=0.3\textwidth]{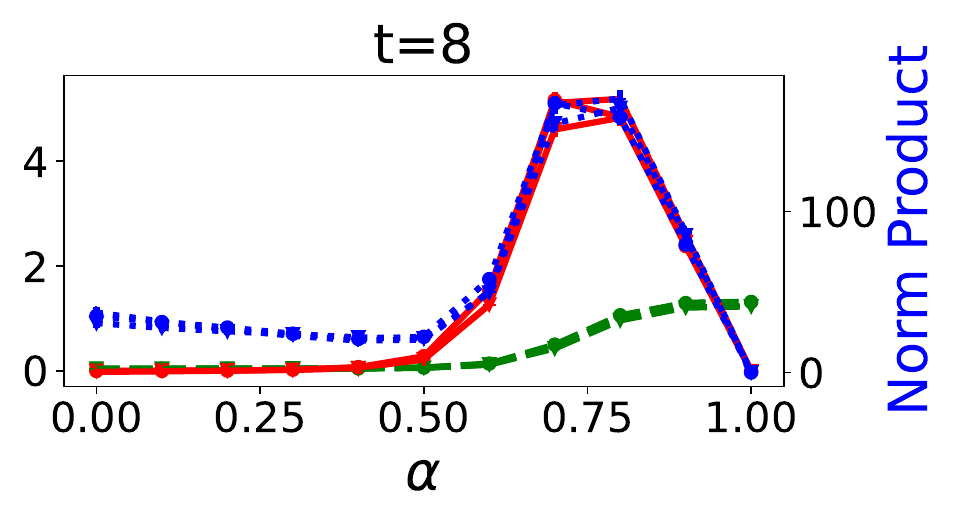}}
    \subfloat[]{\includegraphics[width=0.3\textwidth]{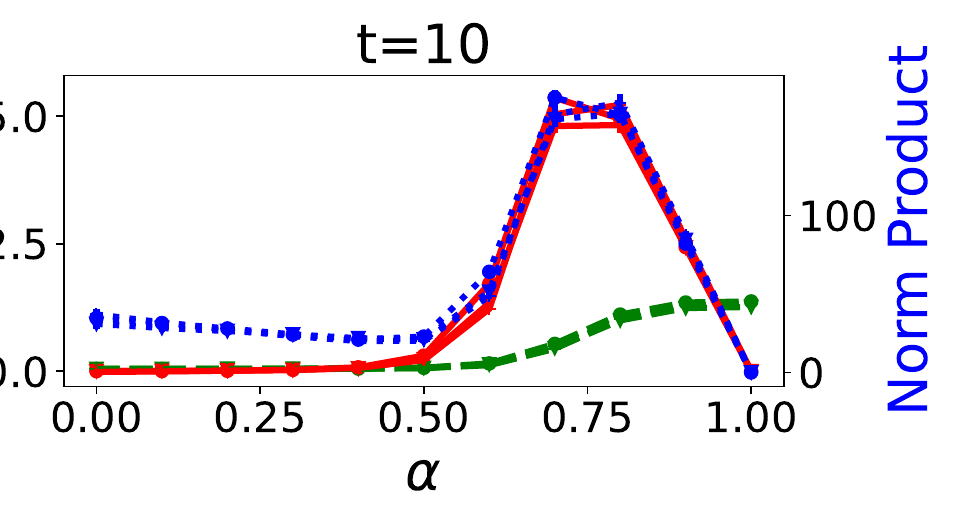}}
    \caption[]{ARC-Easy empirical results for RQ2. The plots include the loss for downstream testing data $\gb{\alpha}$ on a 1-D slice between the PortLLM patch and the stepwise fine-tuning patch (dashed green line), the derivative of the 1-D slice (solid red line), and the norm product  $\|\nabla_{\unexpp{ }}\uloss{\base{t}}((1-\alpha)\unexpp{0})\|\|\unexpp{0}\|$ (dotted blue line). We show $t\in\{2,4,6,8,10\}$ and three repetitions on each plot.
    }
    \label{fig:ae_benefit}
\end{figure}

\begin{figure}
    \centering
    \subfloat[]{\includegraphics[width=0.3\textwidth]{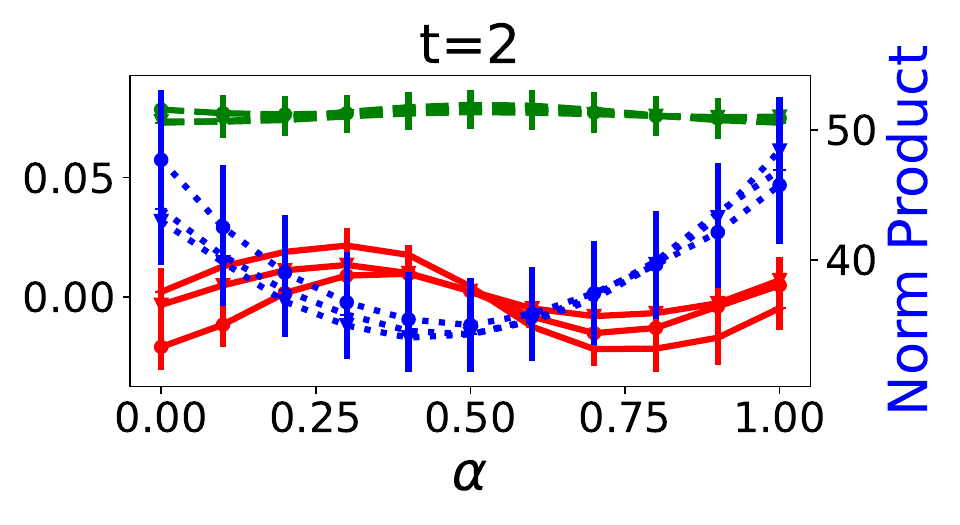}}
    \subfloat[]{\includegraphics[width=0.3\textwidth]{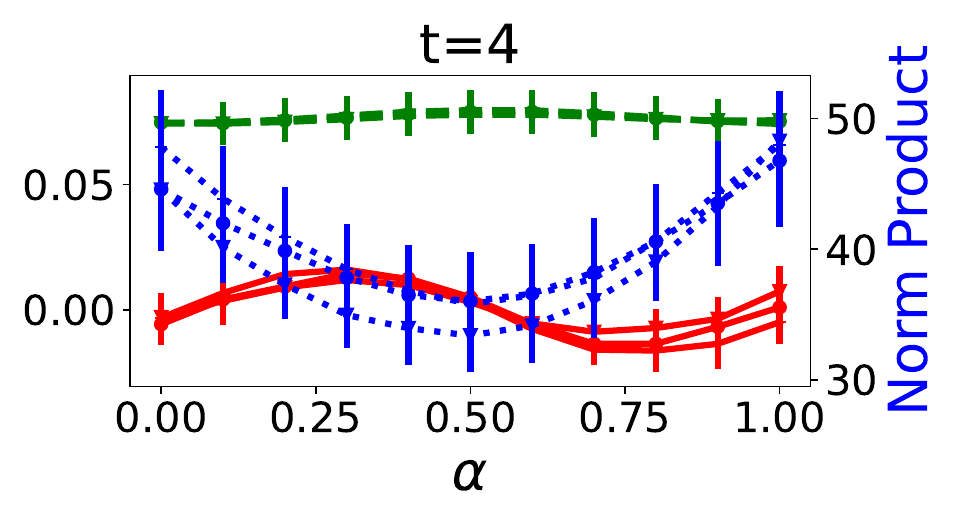}}
    \subfloat{\vspace{15pt}\includegraphics[valign=t,width=0.2\textwidth]{figures/data_dump/cost_legend.pdf}}
    \addtocounter{subfigure}{-1}\\
    \subfloat[]{\includegraphics[width=0.3\textwidth]{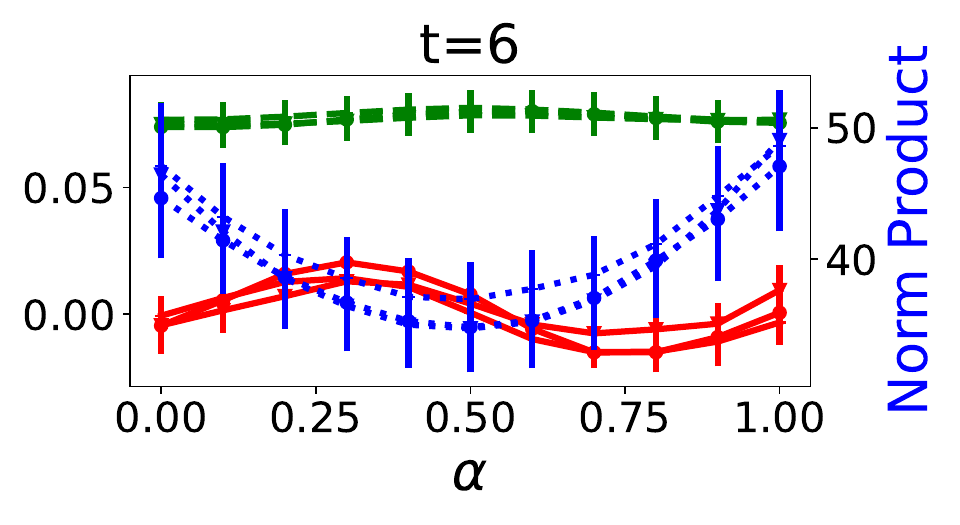}}
    \subfloat[]{\includegraphics[width=0.3\textwidth]{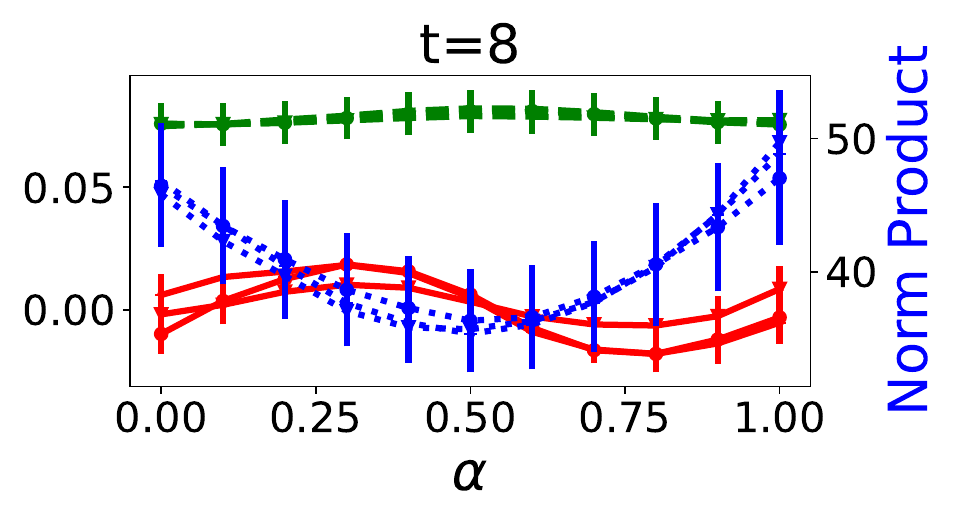}}
    \subfloat[]{\includegraphics[width=0.3\textwidth]{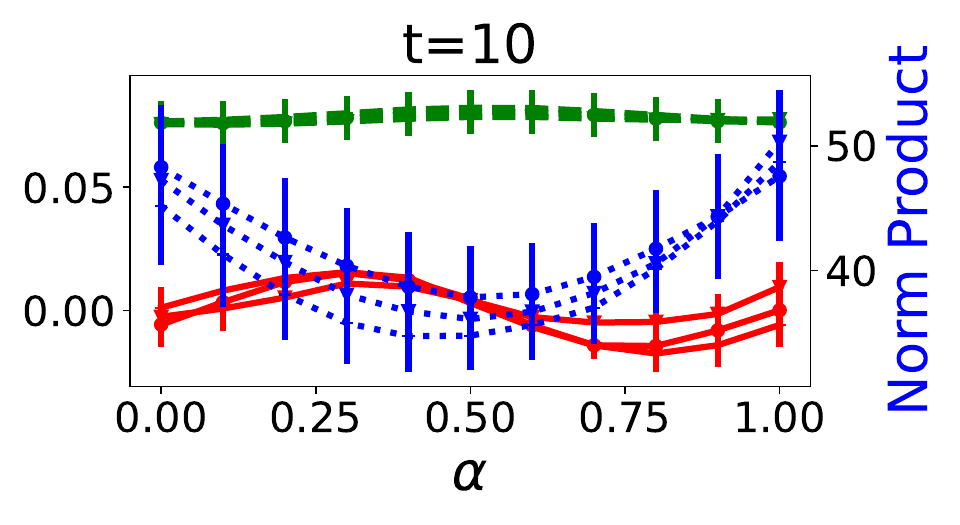}}
    \caption[]{ARC-Challenge empirical results for RQ1. The plots include the loss for downstream testing data $\gc{\alpha}$ on a 1-D slice between the PortLLM patch and the stepwise fine-tuning patch (dashed green line), the derivative of the 1-D slice (solid red line), and the norm product  $\|\nabla_{\unexpp{ }}\uloss{\base{t}}(\uls{\alpha})\|\|\ftdelta{t}\|$ (dotted blue line). We show $t\in\{2,4,6,8,10\}$ and three repetitions on each plot.
    }
    \label{fig:ac_cost}
\end{figure}

\begin{figure}
    \centering
    \subfloat[]{\includegraphics[width=0.3\textwidth]{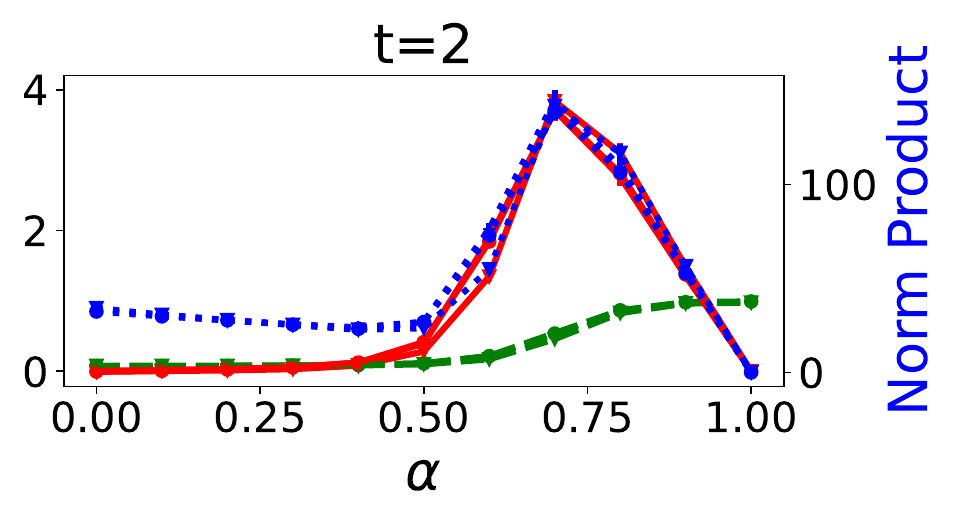}}
    \subfloat[]{\includegraphics[width=0.3\textwidth]{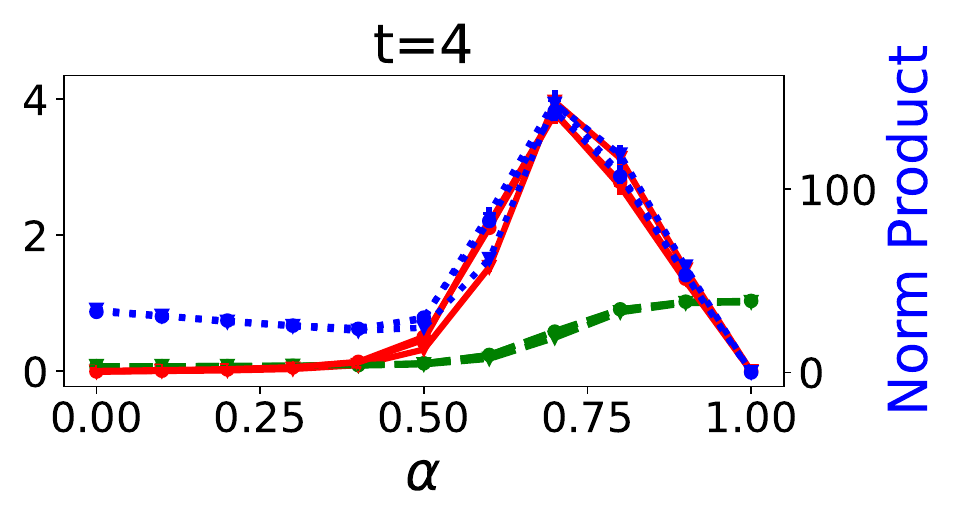}}
    \subfloat{\vspace{15pt}\includegraphics[valign=t,width=0.2\textwidth]{figures/data_dump/benefit_legend.pdf}} 
    \addtocounter{subfigure}{-1}\\
    \subfloat[]{\includegraphics[width=0.3\textwidth]{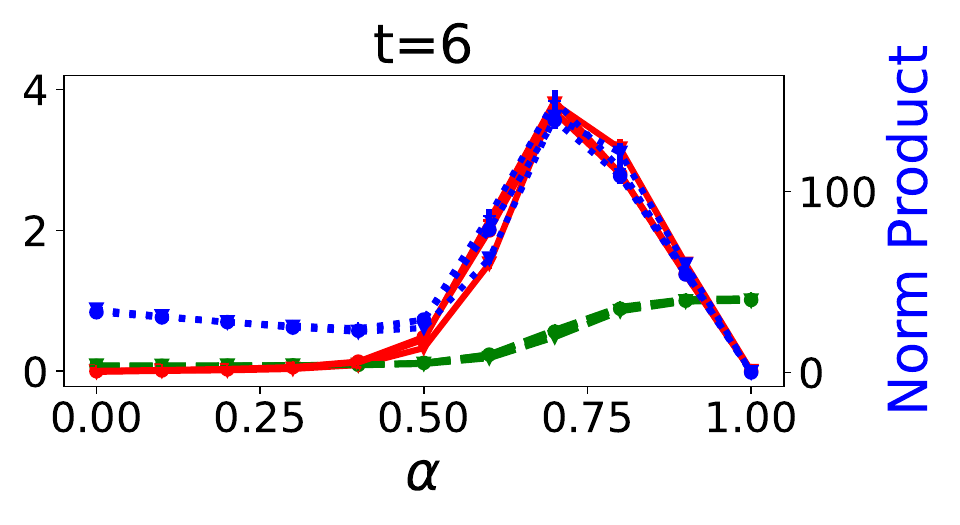}}
    \subfloat[]{\includegraphics[width=0.3\textwidth]{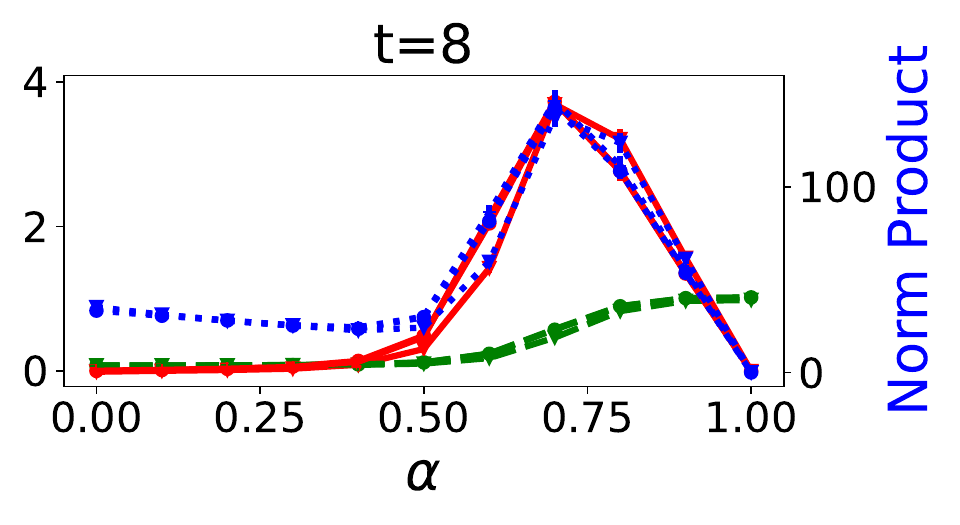}}
    \subfloat[]{\includegraphics[width=0.3\textwidth]{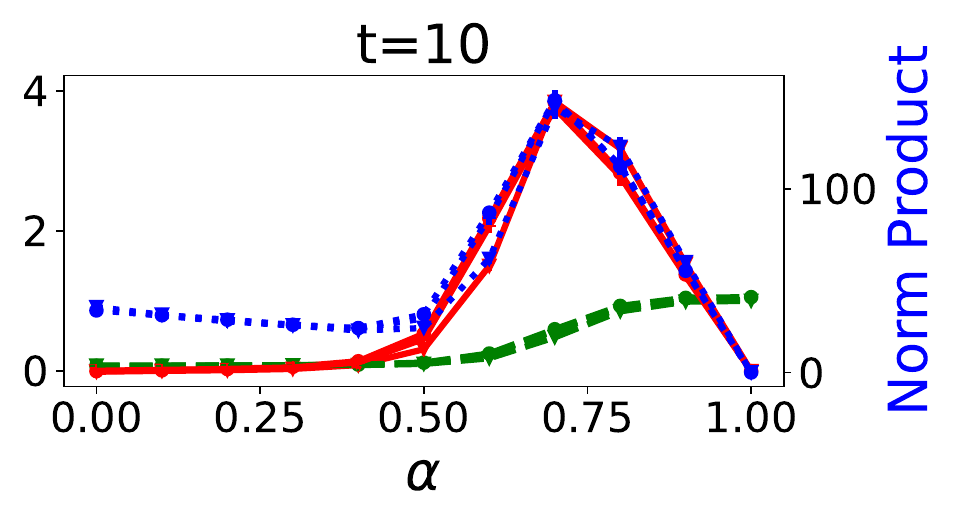}}
    \caption[]{ARC-Challenge empirical results for RQ2. The plots include the loss for downstream testing data $\gb{\alpha}$ on a 1-D slice between the PortLLM patch and the stepwise fine-tuning patch (dashed green line), the derivative of the 1-D slice (solid red line), and the norm product  $\|\nabla_{\unexpp{ }}\uloss{\base{t}}((1-\alpha)\unexpp{0})\|\|\unexpp{0}\|$ (dotted blue line). We show $t\in\{2,4,6,8,10\}$ and three repetitions on each plot.
    }
    \label{fig:ac_benefit}
\end{figure}

Here we provide all results for empirically plotting the 1-D slice of the loss landscape for cost (RQ1) and benefit (RQ2) for Mistral-7B pretrained on Fineweb data in Figures~\ref{fig:wg_cost}--\ref{fig:ac_benefit}. When available, we estimate the true loss using $2000$ testing examples (BoolQ and ARC-Easy). Otherwise, we use the entire available testing split (approximately $1270$ examples for WinoGrande and approximately $1170$ examples for ARC-Challenge). In each case, we show error bars for the $95\%$ confidence interval. Note that the norm product, $\|\nabla_{\unexpp{ }}\uloss{\base{t}}(\uls{\alpha})\|\|\ftdelta{t}\|$, is much larger than the inner product $\gcp{\alpha}=\langle \nabla_{\unexpp{ }}\uloss{\base{t}}(\uls{\alpha}),\ftdelta{t}\rangle$, indicating that it is approximate orthogonality, not small vector norms, that explains the small magnitude of $\gcp{\alpha}$.

\begin{figure}
    \centering
    \subfloat[]{\includegraphics[width=0.3\textwidth]{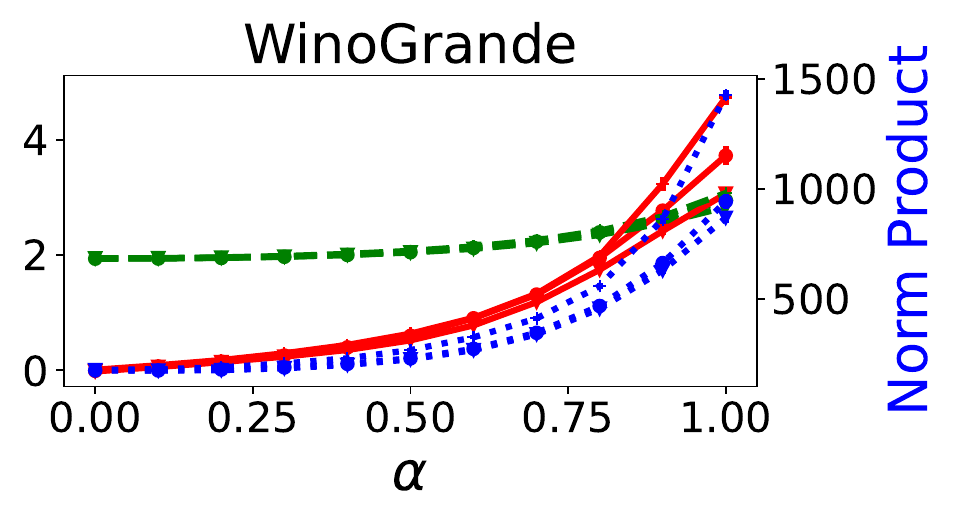}}
    \subfloat[]{\includegraphics[width=0.3\textwidth]{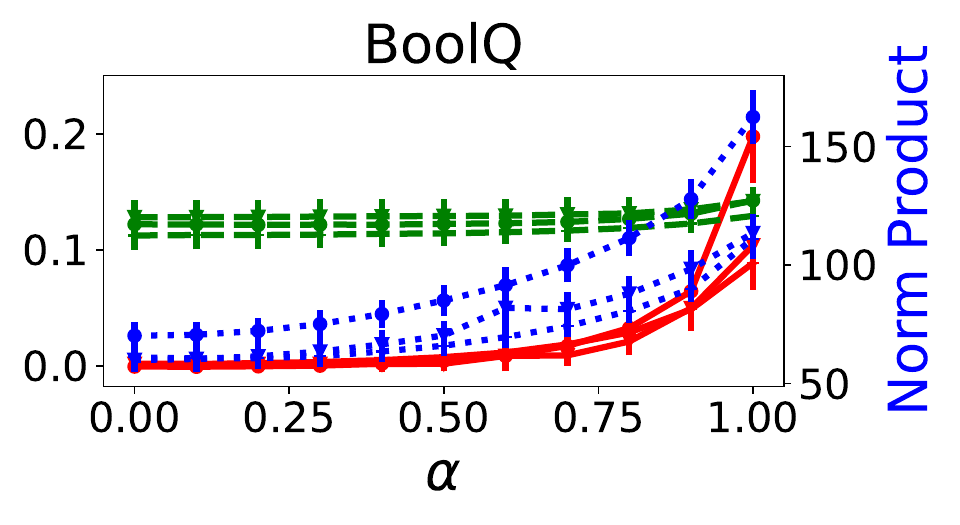}}
    \subfloat{\vspace{15pt}\includegraphics[valign=t,width=0.2\textwidth]{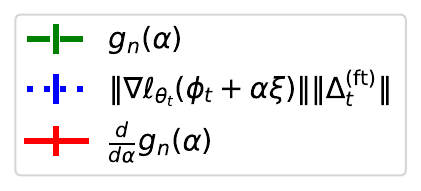}} 
    \addtocounter{subfigure}{-1}\\
    \subfloat[]{\includegraphics[width=0.3\textwidth]{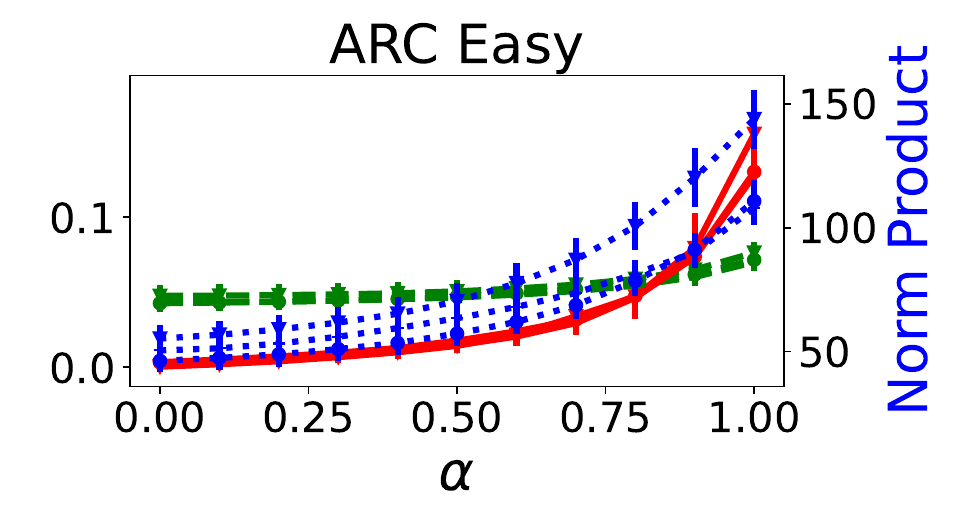}}
    \subfloat[]{\includegraphics[width=0.3\textwidth]{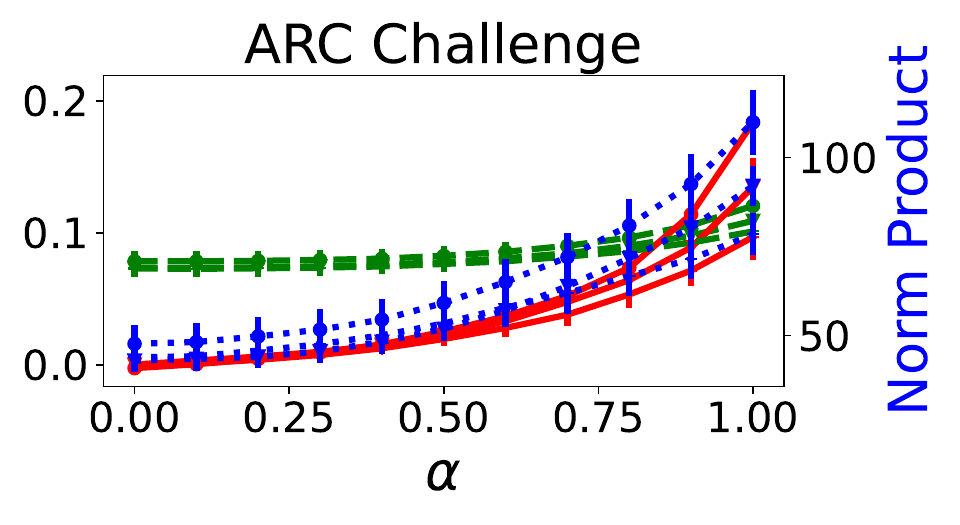}}
    \subfloat[]{\includegraphics[width=0.3\textwidth]{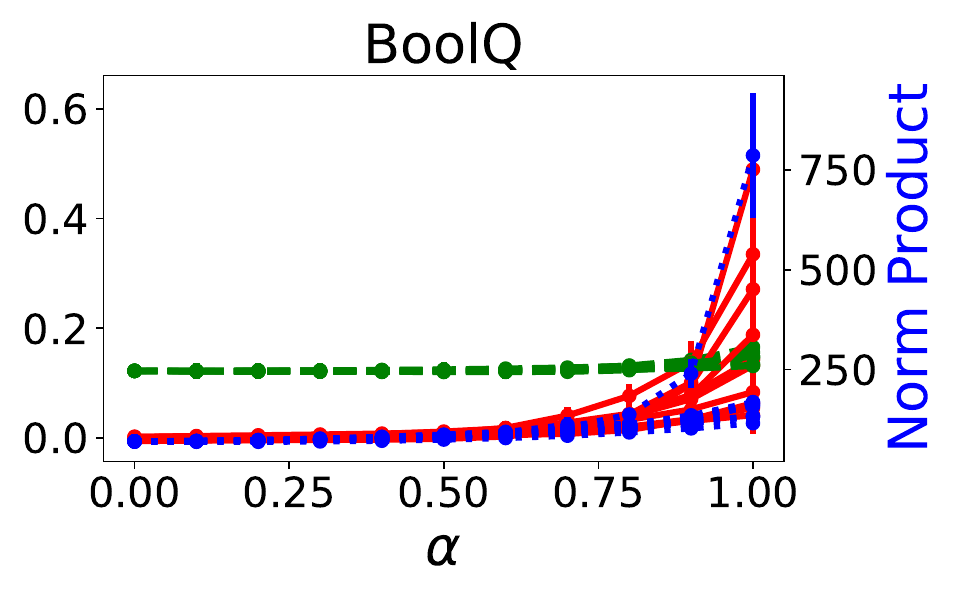}}
    \caption[]{1-D slice of the loss landscape in a random direction. We select the direction as a Gaussian random vector and rescale it to have a norm equal to the distance between the PortLLM and stepwise fine-tuning tasks. Results are for $t=2$ and (a) WinoGrande, (b) BoolQ, (c) ARC-Easy, and (d) ARC-Challenge. Three repetitions are shown for each plot. We also show $10$ random direction for BoolQ for a fixed training repetition in (e). 
    }
    \label{fig:rand_t2}
\end{figure}

To better understand these results about the loss landscape, we also plot the loss in a random direction starting at the stepwise fine-tuning patch for $t=2$.\footnote{We select one time step noting that results follow the same pattern for each $t\in\{2,4,6,8,10\}$ in Figures~\ref{fig:wg_cost}--\ref{fig:ac_benefit}.} The random direction $\xi'\sim\text{Norm}(0,I)$ is rescaled as $\xi=\xi'\cdot \|\ftdelta{}\|/\|\xi'\|$ so that is norm equals the distance between the stepwise fine-tuning patch and the PortLLM patch. We plot $g_n(\alpha)=\uloss{\base{t}}(\unexpp{t}+\alpha\xi)$ as a 1-D slice in the loss landscape in a random direction starting at $\unexpp{t}$. We find that the loss is approximately constant for $\alpha\leq0.9$ increases slightly for $\alpha>0.9$. This indicates that the stepwise fine-tuning patch $\unexpp{t}$ is in a flat region of the loss landscape. To increase confidence that the plotted directions are characteristic, we plot $10$ random direction for BoolQ for a fixed training repetition in Figure~\ref{fig:rand_t2}(e). 

\newpage
\bibliography{refs}

\end{document}